\documentclass[jair,twoside,11pt,theapa]{article}
\usepackage{jair, theapa, rawfonts}
\jairheading{7}{2023}{}{08/22}{06/23}
\ShortHeadings{Hierarchical Decomposition and Analysis for Generalized Planning}
{Srivastava}
\firstpageno{1}

\usepackage{siddmath}
\usepackage{multirow}
\usepackage[ruled, vlined]{algorithm2e}
\usepackage{color}
\usepackage{enumitem}
\usepackage{url}
\usepackage{subcaption}
\usepackage{cleveref}
\usepackage{rotating}
\usepackage{afterpage}

\newcommand{\vars}{\mc{V}}
\newcommand{\axns}{\mc{A}}
\newcommand{\dom}{\mc{D}}
\newcommand{\states}{\mc{S}}

\renewcommand{\cite}{\shortcite}
\renewcommand{\citeauthor}{\shortciteauthor}

\newcommand{\citet}[1]{\citeauthor{#1}~\citeyear{#1}}

\begin{document}

\title{Hierarchical Decompositions and Termination Analysis \\for  Generalized Planning}

\author{\name  Siddharth Srivastava \email siddharths@asu.edu \\
       \addr School of Computing and Augmented Intelligence \\
       Arizona State University\\
       Tempe, AZ 85281 USA}


\maketitle

\begin{abstract}
This paper presents new methods for analyzing and evaluating generalized plans
that can solve broad classes of related planning problems.  Although synthesis
and learning of generalized plans has been a longstanding goal in AI, it remains
challenging due to fundamental gaps  in methods for analyzing the scope and
utility of a given generalized plan. This paper addresses these gaps by
developing a new conceptual framework along with proof techniques and
algorithmic processes for assessing termination and goal-reachability related
properties of generalized plans. We build upon classic results from graph theory
to decompose generalized plans into smaller components that are then used to
derive hierarchical termination arguments.
These methods can be used to determine the utility of a given generalized plan,
as well as to guide the synthesis and learning processes for generalized plans.
We present theoretical as well as empirical results illustrating the scope of
this new approach. Our analysis shows that this approach significantly extends
the class of generalized plans that can be assessed automatically, thereby
reducing barriers in the synthesis and learning of reliable generalized
plans.
\end{abstract}

\section{Introduction}
\label{Introduction}

Enabling autonomous agents to learn and represent generalized knowledge that can be
used to solve multiple related planning problems is a longstanding goal in AI.
The ability to create transferrable, generalized plans that can solve large
classes of sequential decision making problems has long been recognized as
essential in achieving this goal. However, research on the problem has been
limited due to technical challenges in analyzing generalized plans. The problem
of determining whether an arbitrary generalized plan is useful -- i.e., whether
it can be ``transferred'' to solve a desired class of problem instances is
equivalent to determining whether a generalized plan terminates, which is
undecidable in general. This significantly limits approaches for computing
generalized plans: the absence of an effective evaluation function precludes
computationally popular approaches for learning from past data as well as
methods for synthesizing and improving candidate generalized plans. This reduces
not only the robustness of algorithms for computing generalized plans but also
the reliability of the computed generalized plans.

This paper presents a new, hierarchical approach for assessing the utility of a
generalized plan. Since reachability of a desired goal state while executing a
generalized plan can be reduced to determining termination, we focus on methods for
determining termination. Our main contribution is a new conceptual framework and
proof technique  that can be used to design and validate algorithms for
analyzing the reachability and termination properties of generalized plans. We
draw upon insights from classic results in graph theory to create a hierarchical
decomposition of a given generalized plan. This decomposition facilitates more
general termination analysis than is possible using prior work. Although the
problem of termination for generalized plans is undecidable in general, we show
that in practice, the methods developed this paper can determine termination for
broad classes of generalized plans that are beyond the scope of existing
methods. This approach can also be used in popular generate-and-test driven
paradigms for generalized planning where candidate generalized plans can be
synthesized or learned from examples and pruned or refined on the basis of
reachability analysis facilitated using the presented methods.

Advances in determining termination for limited classes of generalized
plans~\cite{srivastava10_precons,srivastava11_qnp,srivastava12_aij} have led  to
immense progress in the field   (e.g.,
\citet{segovia18_jair}; \citet{bonet18_ce}; \citet{illanes19_gp}; a more complete survey is
presented in the next section).  We expect that methods for more general methods
for analyzing generalized plans will further enable continued breakthroughs in
the field.

Prior work on theoretical aspects of generalized planning relies upon strong
assumptions that restrict the structure of generalized plans, or limit the
permitted actions to ``qualitative'' actions that cannot capture general forms
of behavior. The framework developed in this paper goes beyond these
limitations. It offers a sound but non-complete test of termination for
generalized plans with arbitrary structures and actions that can increment or
decrement variables by specific amounts in non-deterministic or deterministic
control structures. Sound and complete methods for termination assessment of
generalized plans are not possible, due to equivalence with the halting problem
for Turing machines. Our framework supports counter-based actions that can
express the full range of structured behaviors including arbitrary Turing
machines. They can express changes in Boolean state variables as well as changes
in higher-order state properties or features (e.g., the number of packages that
still need to be delivered in logistics planning problems). Prior work on
generalized planning has established that such counter-based representations
capture the essence of generalized plans as needed for analysis of their
utility~\cite{srivastava10_precons,srivastava12_aij} as well as for synthesis of
generalized plans and policies~\cite{srivastava08_genplan,bonet21_general}.

The main new insight in this paper  is that classic results in graph theory can
be used to decompose an arbitrary generalized plan, expressed as a finite-state
machine, into a finite hierarchy of generalized plans with a desirable property:
the structure of generalized plans necessarily decreases in complexity as one
descends in the hierarchy. This allows us to inductively build a new form of
termination argument for a parent generalized plan in the hierarchy by composing
termination arguments for its children generalized plans. Theoretical analysis
and empirical results  show that this method effectively addresses a broad class
of generalized plans that were not amenable to existing approaches.

The rest of this paper is organized as follows. Sec.\,\ref{sec:related} presents
a survey of related work on the topic followed by our formal framework and the
problem setting (Sec.\,\ref{sec:formulation}). Sec.\,\ref{sec:termination}
presents our new algorithmic framework for the analysis of generalized plans.
This is followed by  theoretical analysis illustrating the new proof techniques
that this approach enables along with our key theoretical results on the
soundness of the presented algorithm (Sec.\,\ref{sec:formal}) as well as our
empirical analysis (Sec.\,\ref{sec:empirical}). Sec.\,\ref{sec:conclusions}
presents our conclusions and  directions for future work.

\section{Related Work}
\label{sec:related}
Early work by \citet{levesque05_loop} articulated the value  of planning with
iterative constructs and the challenges of proving that such constructs would
terminate (or equivalently, achieve the desired goals) upon execution. This work focused on
settings where problem classes are  defined by varying a single numeric
\emph{planning parameter}. Levesque showed that this parameter could be used to
build iterative constructs. He argued that rather than attempting to assess or
prove the termination of such plans, asserting weaker guarantees could lead to
computationally pragmatic approaches. Levesque proposed validation of computed
iterative plans up to a certain upper bound on the single numeric planning
parameter. To the best of our knowledge, this work constitutes the first clear
articulation of the value and challenges of computing plans with loops using AI
planning  methods. 

 \subsection{Counter-Based Models for Generalized Planning}
 \citet{srivastava08_genplan} showed that counters based on logic-based
 properties could be used to create numeric features, e.g., the number of cells
 that need to be visited in a grid exploration task. Such features can  help
 identify useful iterative structures and compute ``generalized plans''.
 Furthermore, the team showed that  counter-based models for generalized
 planning could be used to determine whether a computed generalized would
 terminate and reach the goal. These methods were used to compute generalized
 plans with simple loops. The plan synthesis process ensured provable
 correctness and for a broad class of problems, the computed plans were
 guaranteed to solve infinitely many problem instances involving arbitrary
 numbers of counters. \citet{hu10_1d}  showed that determining termination for
 plans featuring iteration over a single numeric planning parameter is
 decidable. In subsequent work, \citet{srivastava10_precons} showed that the
 problem of identifying useful cyclic control structures in generalized plans
 could be studied using foundational models of computation such as abacus
 programs by transforming the planning-domain actions into equivalent operations
 that changed counters corresponding to logic-based state properties. The team
 developed algorithms for determining termination and graph-theoretic
 characterizations of generalized plans that could be assessed for termination
 despite the general incomputability of the problem~\cite{srivastava12_aij}.
 These methods were also used to develop directed search and learning techniques
 for generalized planning~\cite{srivastava11_hybrid}. 

In practice, the main technical problems in determining whether a generalized
plan is ``useful'' is that it is difficult to construct termination arguments
for complex terminating generalized plans: the set of generalized plans that can
be proved to be terminating is a small subset of the set of terminating
generalized plans.  Although the strict subset relationship must hold  due to
equivalence of this problem with the halting problem, it is essential to develop
new approaches that identify and push the boundary of decidability further. 

The framework of qualitative numeric planning (QNP, \citet{srivastava11_qnp})
extended the counter-based model for generalized planning further to address
these limitations. This framework introduces action semantics under
which the set of terminating generalized plans reduces to 
generalized plans for which the ``Sieve'' algorithm can assert
termination.  The QNP framework is broadly applicable in practice and yet
insufficient to express Turing machines. A more formal comparison of termination
under qualitative and deterministic semantics is presented in
Sec.\,\ref{sec:comparison}. QNPs were later extended to more general settings
along with analysis of their expressiveness and a more general version of the
Sieve algorithm~\cite{srivastava15_qnp}. \citet{bonet20_qnp} showed that the
analysis conducted by the Sieve algorithm for QNP problems can also be viewed as
a fully observable non-determinstic  (FOND) planning process. These foundational
formulations of generalized planning have been extended in several directions.
\citet{bonet19_ltl} develop connections between generalized planning and LTL
synthesis; \citet{belle22_analyzing} analyzes the relationships between various
correctness criteria in stochastic and non-deterministic settings.

\subsection{Meta-Level Planning for Computing Generalized Plans} The theoretical
advances discussed above have been accompanied with numerous advances in
approaches for computing generalized plans. \citet{bonet09_controller} showed
that the computation of finite-state controllers for some classes of planning
problems could be reduced to planning by creating meta-level planning domains
whose actions involved the addition or deletion of edges in a controller. These
finite-state controllers were observed to have good generalization capabilities.
Several threads of research have developed this approach of creating meta-level
planning problems that synthesize generalized plans in the form of
controllers~\cite{bonet09_controller,hu11_gp,hu13_fsc}. 
\citet{segovia18_jair}
present algorithms for computing hierarchical generalized plans that include
subroutines and are guaranteed to solve an input set of finitely many planning
problems. The planning domains used in this reduction include actions that
construct components of hierarchical finite-state controllers as well as validate
the resulting controllers on the input problem set. This paradigm of evaluation
using a finite validation set has been developed along multiple directions to
utilize finite sets of positive examples as well as negative examples indicating
undesired outcomes of plan execution~\cite{segovia20_gp_examples} and with
finite validation sets for use in a general heuristic search process for
computing generalized plans~\cite{segovia21_gphs,aguas22_socs}. 

\subsection{Broader Applications}  Foundations of generalized planning
discussed above have also enabled broader advances in sequential decision
making. As noted above, state representations based on counters capturing the
numbers of objects that satisfy various properties were developed originally for
identifying and analyzing iterative structures for generalized plans. They have
since been found to be useful also for computing generalized, domain-wide
planning knowledge.  Such methods have been utilized for learning and synthesis
of generalized knowledge for sequential decision making in the form of general
sketches and policies for
planning~\cite{bonet18_ce,frances21_learning,bonet21_general,drexler22_sketches},
for  learning neuro-symbolic generalized heuristics for
planning~\cite{ks21_aaai} and neuro-symbolic generalized Q-functions for
reinforcement learning in stochastic settings~\cite{ks22_ijcai}, as well as for
few-shot learning of generalized policy automata for stochastic shortest path
problems~\cite{kns22_neurips}. In the terminology of metrics for generalized
planning presented by \citet{srivastava11_aij}, these methods compute
generalized  plans that have a relatively higher cost of
instantiation\footnote{the computational cost of computing a plan for a specific
problem instance using the computed auxiliary data structure.} that is still
much lesser lower than that of planning from scratch. On the other hand, these
methods provide a significantly  higher domain coverage than that of an
optimized generalized plan. These directions of research bridge the gap between
generalized planning and lifted sequential decision
making~\cite{boutilier01_sdp,sanner09_fomdp,cui19_stochastic}. Other approaches
for learning generalized knowledge include approaches for learning general
policies in a relational
language~\cite{khardon99_strategies,winner03_Distill,yoon08_control}, learning
heuristics for solving multiple planning
problems~\cite{shen20_learning,toyer20_asnets,rivlin20_gdrl,ferber22_neural,stahlberg22_learning}
as well as generalized neural policies for relational
MDPs~\cite{garg20_symbolic}.

\section{Problem Formulation}
\label{sec:formulation}
We use a foundational but powerful representation
where all variables are numeric variables with $\mathbb{N}$ 
 as their domains. Let $\vars$ be the set of such
variables.
A {\em concrete state} is an assignment that maps each variable in
$\vars$ to a value in that variable's domain. We denote the set of all possible
concrete states as $\states_\vars$. The value of a variable $x$ in a state $s\in
\states_\vars$ is denoted as $s(x)$. In this paper we use the unqualified term
``state'' to refer to concrete states. 

\define{An \textbf{action} consists of a
  precondition, which maps each variable in $\vars$ to a union of intervals for that variable, and a set of action effects,
  $\emph{effects}(a)$. Each member of $\emph{effects}(a)$ is of the
  form $\oplus x$ or $\ominus x$, where $x\in vars$;
  $\emph{effects}(a)$ must include at most one occurrence of each
  variable in $\vars$.}

Prior work in generalized planning considers three types of interpretations of
  $\oplus, \ominus$ that correspond to popular frameworks in the
  literature~\cite{srivastava15_qnp}. We focus on deterministic and qualitative
  semantics in this work:

\paragraph{Deterministic semantics} Under deterministic semantics, each
occurrence of $\oplus$ ($\ominus$) in an action effect is interpreted as an
increment (decrement) by a fixed discrete quantity. E.g., an action's effects
may include $x_1 + 1, x_2-5, x_3+1$ for variables $x_1, x_2,$ and $x_3$.
$a(s_1)$ is defined as the unique state $s_2$ such that for every $x\in\vars$ if
$x+i$ is an effect of $a$, $s_2(x)=s_1(x)+i$ and if $x-j$ is an effect of $a$
then $s_2(x)=s_1(x)-j$. $s_2(x)=s_1(x)$ for all $x$ that don't occur in effects
of $a$. The special case of deterministic semantics where for every variable
$x$, $\oplus x = x +1$ and $\ominus x = x-1 $ yields formulations close to
abacus programs~\cite{lambek61_abacus}. 

Although we will focus on deterministic semantics in this paper, we introduce
qualitative semantics below in order to compare and contrast our approach with
prior work.

  \paragraph{Qualitative semantics}  Under qualitative semantics, each
  occurrence of $\oplus$ ($\ominus$) is interpreted as an increase  (decrease)
  by a non-deterministic amount.  The amount of change caused due to each
  $\ominus$ is such that in any  execution, a finite sequence of consecutive
  $\ominus x$ effects is sufficient to reduce $x$ to zero for any finite value
  of $x$. This is defined formally as follows~\cite{srivastava11_qnp}. Let
  $\epsilon>0$ be an unknown constant that is fixed for an entire execution. Formally, the
  effect of $\oplus x$ is to non-deterministically increase $x$ by
  $\delta_{\oplus}$, where $\delta_{\oplus}\in[\epsilon, \infty)$. The effect of
  $\ominus x$ is to non-deterministically decrease $x$ by $\delta_{\ominus}$ such
  that for every execution, $\delta_{\ominus}\in [\emph{min}\set{\epsilon, x},
  x]$.

To accommodate a unified discussion of these semantics, we will use the notation $a(s)$ to refer to the set of states that can result upon execution of $a$ in $s$ and clarify the semantics being used as required.

\define{A \textbf{planning domain} $\tuple{\vars,
    \axns}$ consists of a finite set of variables $\vars$, and a finite set of actions
  $\axns$ over  $\vars$. }

\define{A \textbf{planning problem} $\tuple{\dom,
    s_o, g}$ consists of a planning domain $\dom=\tuple{\vars,
    \axns}$, a concrete state $s_0$ and a goal mapping $g$ from
  $V\!\!\subseteq\!\vars$ to intervals in $\mathbb{N}$.
 }

\subsection{Solutions to Planning Problems}
\vspace{-2pt}
 At every step, the action to be performed for solving a given planning problem
 may depend on finite representations of the history of executed actions and
 received observations. Such solutions can be represented as graph-based
 generalized plans~\cite{srivastava11_aij}, or as finite-state
 controllers~\cite{bonet09_controller,srivastava10_precons}. Formally, we use the following
 notion of a finite-state controller based policy. 

\begin{definition}
Let $\dom=\langle \vars, \axns \rangle$ be a planning domain and let $\Psi$ be
the set of formulas over propositions of the form $x_i\gtreqless l_i $ where
$x_i\in \vars$ and $l_i\in \mathbb{N}$. A \emph{finite-memory policy (FMP)} for
$\dom$ is defined as $\langle Q, q_0\in Q, \delta,\kappa \rangle$ where $Q$ is a
finite set of control states, $q_0$ is the starting state, $\delta\subseteq
Q\times Q$ is the transition relation and  $\kappa: \delta \rightarrow \Psi
\times \mathcal{P}(\axns)$ is a labeling function that labels each directed edge
between control states with an execution condition and a set of actions. 
\end{definition}

Throughout this paper we will assume that execution conditions on FMP edges
include as defaults the non-negativity condition $x_i\ge 0$, for all $x_i\in
\vars$. In practice each variable can have a different lower bound as the
methods presented in this paper  depend only on the existence of default bounds
on variables. The technical results presented in this paper can be easily
extended to settings where $x_i$'s also have upper bounds. They can also be
extended to settings where $x_i$'s are bounded only above by replacing arguments
about actions' effects on decreasing variables with their effects on increasing
variables and including the upper bounds as default execution conditions. 

This representation generalizes the existing forms of policies used with
qualitative as well as deterministic semantics. When needed for clarity, we will
use the term ``control states'' or ``qstates'' to refer to the states of an FMP
policy and ``problem states'' to refer to the states defined by a planning
domain. 

Execution of an FMP starts with $q_0$. At any stage during the execution where
the problem state is $s_1\in\states_\vars$ and the FMP is in control state
$q_1\in Q$, the agent can execute any action $a$ s.t. there exists a control
state $q_2\in Q$ for which $\kappa(q_1, q_2)=(\psi, A)$, $a\in A$, and
$s_1\models \varphi$. Following such an action, the FMP's control state
becomes $q_2$. Execution stops at a state $s$ if the FMP  is in a control
state $q$ that has no outgoing edges whose execution condition is consistent
with $s$. Thus, an FMP terminates if its execution reaches a terminal control state, or
if it reaches a control state such that none of its outgoing edge conditions is
satisfied in the current problem state.

An FMP is said to be \emph{deterministic} iff for every $q\in Q$, the set of
outgoing edges are labeled with mutually inconsistent execution conditions and
singleton action sets so that at most one action is possible at every step
during execution.

Existing notions of qualitative policies are closely related with FMPs. For
instance, an abstract policy $\pi$~\cite{srivastava11_qnp,srivastava15_qnp} that
maps abstract states to actions can be expressed as FMPs with one control state
$q_0$ and the labeling function $\kappa(q_0, q_0)=\set{(\tilde{s}, \set{a}):
\pi(\tilde{s})=a}$. ``Sketch'' based generalized policies~\cite{bonet21_general} can
be expressed as FMPs where the edge label consists of a condition $\psi$ but
$\mathcal{P}(A)$ is replaced by a set of integers denoting changes in the values
of a subset of the variables in $\mathcal{V}$. An edge in such a sketch can be
taken if the agent executes a sequence of actions that causes the corresponding
change. Since the framework for analysis developed in this paper considers the
changes induced by FMP edges rather than the action names, it offers a promising
direction for analyzing properties of sketch based policies as well.

We say that an FMP has a \emph{well-defined set of terminal states} $H\subseteq
Q$ if the states in $H$ have no outgoing edges and when started with any state
in $s\in\states$, execution continues until $H$ is reached. Execution of
policies that do not have well-defined halt states can end in arbitrary control
states when the policy does not have any outgoing edges corresponding to the
current state of the problem.     This represents situations that are in some
sense unforeseen according to the FMP and is especially common with FMPs that
are generated through learning over a limited training dataset.

\subsubsection{Solution Properties}
Given an initial state $s_0\in\states_\vars$, \emph{an execution of an FMP
policy} $\langle Q, q_0\in Q, \delta,\kappa \rangle$ is defined as a sequence
$(q_0, s_0), a_0, (q_1, s_1), a_1, \ldots$ such that $\kappa(q_i,
q_{i+1})=(\varphi_i, A_i)$, $a_i\in A_i$, $s_{i+1}\in a_i(s_i)$ and $s_i\models
\varphi_i$. 
Two criteria over the set of executions possible under an FMP policy  can be
used to define the quality of the policy. If a policy $\Phi$ is such that every
execution of $\Phi$ stops after a finite number of steps when started with an
initial state $s_0$, we say that $\Phi$ terminates for $s_0$. If $\Phi$
terminates for all $s\in\states_\vars$,  we say that it  is a \emph{terminating}
policy. A terminating policy whose execution ends only at states satisfying the
goal condition $g$ of the given domain is said to be \emph{goal achieving}.

\section{Determining Termination of Finite Memory Policies}
\label{sec:termination}
Recall that we adopt the fail-stop mode of  execution semantics for FMPs, where
 execution stops if the current control state has no outgoing edges consistent
 with the current problem state. As discussed under related work, several
 algorithms for generalized planning learn or iterate over multiple solution
 structures that can be expressed in the form of FMPs. To evaluate the utility
 of a candidate FMP as a solution, we need to consider two key properties:
 reachability and termination. Ideally, we would want to establish that every
 execution of an FMP reaches the goal in a finite number of steps for all the
 initial problem states of interest.  
 
 It is well known that decision problems of termination and reachability can be
 reduced to each other. Prior work~\cite{srivastava10_precons,srivastava12_aij}
 shows that this is undecidable in general, because showing that every execution
 terminates in a finite number of steps is equivalent to the halting problem for
 Turing machines. In this section we present our new approach for hierarchically
 decomposing and analyzing FMPs to determine termination. 
 
 We begin with a formal analysis of the relationship between qualitative
 semantics and deterministic semantics with respect to termination analysis.

\subsection{Termination Under Qualitative and Deterministic Semantics}
\label{sec:comparison}

\paragraph{Background on the Sieve family of algorithms} The Sieve family of
algorithms~\cite{srivastava11_qnp,srivastava15_qnp} are sound and complete tests
of termination under qualitative semantics, but only sound for deterministic
semantics. Although the correctness of these algorithms was proved for abstract
policies and transition graphs, the results carry over in a straightforward
manner to FMPs. The Sieve algorithm~\cite{srivastava11_qnp}
conducted this analysis for variables bounded below. The Progress-Sieve
algorithm~\cite{srivastava15_qnp} generalized this intuition to settings with
variables with discrete observable levels and upper or lower bounds, and is
listed here as Alg.\,\ref{alg:progress-sieve} for ease of reference. 

\begin{algorithm}[t] 
  \DontPrintSemicolon
    \caption{\small (Progress-Sieve) abstract policy termination test~\cite{srivastava15_qnp}\!\!} 
    \label{alg:progress-sieve}
  \begin{small}
    \KwIn{$g = \emph{ts}(\pi,\tilde{s_I})$}
      \nl Remove all edges $e$ of $g$ that have a progress variable
      w.r.t. their SCCs\;

    \nl \If{no edge was removed}{
    \nl Return ``Non-terminating''}
      \nl \For{$g'\in \emph{SCCs-of(g)}$}{
        \nl \If{Progress-Sieve($g'$)=
        ``Non-terminating''}{\nl Return
        ``Non-terminating''}}
    \nl Return ``Terminating''
  \end{small}
  \end{algorithm} 

Intuitively, this class of algorithms operate as follows:  for each strongly
connected component (SCC), the algorithm identifies ``progress'' variables that are
changed in only one direction (positively or negatively): the direction in which
they are bounded (above or below, respectively). Progress-Sieve  removes edges
modifying such a variable if the edge moves it towards the  bound on the
variable. If any edge was removed, the entire process is repeated. This
continues until either the graph is left with no strongly connected components
and the policy is reported as terminating, or strongly connected components
remain, but no progress variables are found and the policy is reported as
non-terminating. 

The following natural result relating termination in deterministic and
qualitative settings is useful to list out for completeness:

\begin{proposition}
Let $\Phi = \tuple{Q, q_0, \delta, \kappa}$ be an FMP. If $\Phi$ terminates for
$s_0\in \states_\vars$ under qualitative semantics, then it must terminate for
$s_0$ under deterministic semantics.
\end{proposition}

The proof is straightforward because increments and decrements by a constant are
a special case of qualitative effects. If $\Phi$ does not terminate under such
effects, then it clearly cannot terminate under the qualitative semantics with
the same action instantiations.

However, the other direction is not true: termination under deterministic
semantics does not imply termination under qualitative semantics, as shown in
the following counter-example.

\begin{figure} 
  \centering
  \includegraphics[scale=0.4]{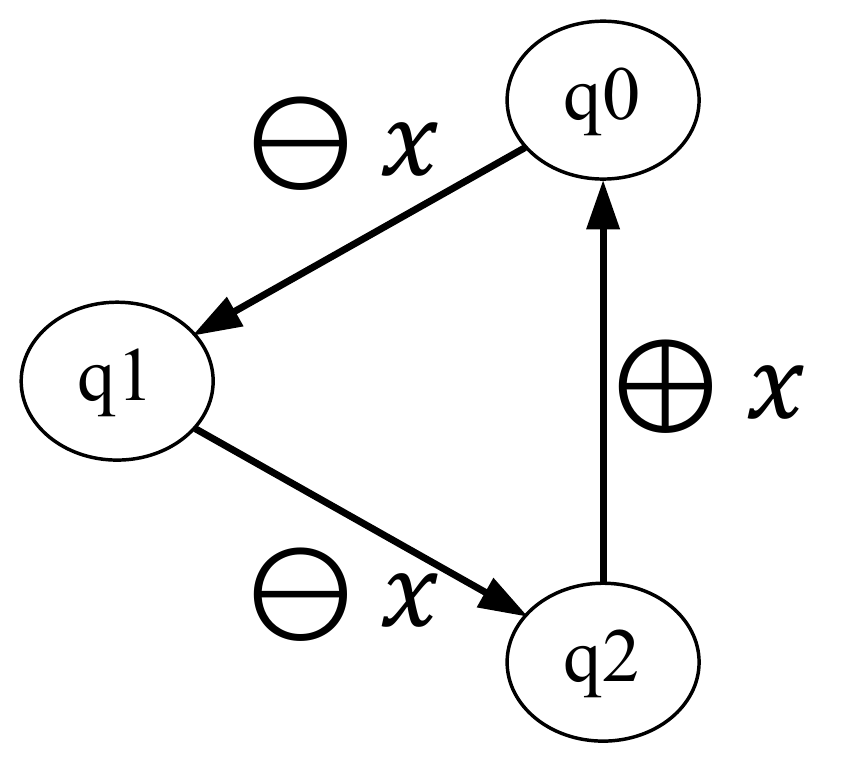}
  \caption{
    \small An FMP that terminates under deterministic semantics ($\oplus=+1$ and
    $\ominus = -1 $)  but not under qualitative semantics (see Example
    \ref{eg:counter_example}).} \label{fig:counter_example}
\end{figure} 

\begin{example}
\label{eg:counter_example}
Let $\Phi$ be an FMP with three qstates $q_0, q_1$ and $q_2$ as shown in
Fig.\,\ref{fig:counter_example}. $\Phi$ is a terminating policy under
deterministic semantics where $\ominus x = x-1, \oplus x= x+1$ (due to the
default edge conditions on non-negativity). However, it is not a terminating
policy under qualitative semantics because the $\ominus$'es may be small enough
to be overwhelmed by the $\oplus$ in every iteration of the cycle.
\end{example}

This example also shows why we are able to develop sound and complete tests for
termination for qualitative numeric planning: under qualitative semantics,
actions are sufficiently imprecise that the effective class of terminating
policies reduces to match the analysis conducted by the Sieve family of
algorithms. This also limits the types of behaviors that can be expressed using
qualitative semantics. \citet{srivastava15_qnp} study this boundary of
expressiveness further and identify conditions under which policies with
qualitative semantics are insufficient for expressing solutions to planning
problems.

\subsection{A Hierarchical Framework for Termination Analysis}
\label{sec:hierarchical_framework}
While the Sieve algorithms discussed above constitute  an efficient and sound
method for identifying a restricted class of terminating FMPs, they have
somewhat obvious limitations when considering deterministic semantics, as
illustrated in Example \ref{eg:counter_example}. Under deterministic semantics
with default edge conditions enforcing non-negativity, the execution of this
SCC is guaranteed to stop in a finite number of
steps regardless of the initial value of the variable being changed by its
actions. However, Sieve algorithms cannot determine that this SCC terminates
because it does not have a progress variable. This is consistent with
completeness under qualitative semantics, which would allow the $\oplus$
operation to increase the variable arbitrarily, although it misses rather
obvious patterns of termination under deterministic semantics. 

Example \ref{eg:counter_example} lends some intuition about the type of
reasoning that is required to develop a more general algorithm for determining
termination under deterministic semantics. Instead of requiring a non-monotonic
decrease over all edges in an SCC, we need a way to assess the net changes on a
variable over execution segments that can be repeated indefinitely in an
infinite execution that might occur when the FMP is interpreted as a finite
automaton without edge conditions. However, reasoning about possible execution
trajectories in an SCC becomes difficult when nested cycles are present:  any
number of iterations of one cycle may be interleaved with an arbitrary number of
iterations of subsequences another cycle within the same SCC.  A test for
termination would have to assess all the infinitely many possible permutations
of this form. 

We formalize this intuition by first defining a richer notion of progress
variables as follows. If variables are bounded above, a similar notion of
increasing progress variables can be defined and the subsequent analysis can be
extended to those cases. As with the default edge conditions, in this paper we
focus on the setting where variables are bounded only from below. We focus on
deterministic semantics in the rest of this paper.

\define{\label{def:decvars} Let $\Pi$ be a set of paths in an FMP and let
$\pi\in\Pi$. A variable $v$ is a  \emph{net-decrease variable} for a path $\pi$
w.r.t. $\Pi$ iff $v$ undergoes a net decrease in $\pi$ and $v$ does not undergo
a net increase in any path in $\Pi$. We use $\decvars^\pi_\Pi$ to denote the set
of all net-decrease variables for $\pi$ w.r.t $\Pi$ and extend this notation to
define $\decvars_\Pi$ as the set of sets of net-decrease variables for all paths
in $\Pi$: $\decvars_\Pi=\{\decvars^\pi_\Pi: \pi\in\Pi\}$. }

Notice that each element of $\decvars_\Pi$ is a set. Thus if the number of
non-empty sets in $\decvars_\Pi$ is the same as the cardinality of $\Pi$, then
every path in $\Pi$ creates a net decrease in some variable that does not
undergo a net increase by any   path in $\Pi$. Intuitively, if $\Pi$ denotes the
set of all possible paths in (including repetitions) in an SCC, and
$\decvars_\Pi$ has no empty sets then execution cannot continue indefinitely
within $\Pi$: every path creates a net decrease that cannot be undone in $\Pi$
and execution conditions include non-negativity. Unfortunately however, this
intuition is  computationally ineffectual because the set of possible execution
paths (which can include multiple executions of SCCs) in a graph with SCCs is
infinite.

\subsubsection{Structures for Hierarchical Decomposition of FMPs} 

To push the envelope on determining whether an FMP will terminate, we need to be
able to structure the possible executions of an SCC in an FMP.  Intuitively,  we
wish to develop a bottom-up process that analyzes ``inner'' loops before moving
on to ``outer'' loops that span the inner ones.  Although this intuition is
helpful, it is not sufficient because an infinite execution of an FMP can
include infinitely many non-consecutive occurrences of non-cycles in the form of
``bridge'' paths and shortcuts through SCCs. We develop this intuition by
building on McNaughton's notion  of \emph{analysis of a
graph}~\cite{mcnaughton69_loops}, which facilitates the construction of a
directed tree over the components of an SCC.  For clarity we will use Gruber's
version of this concept, formalized as Directed Elimination
Trees~\cite{gruber12_loops}. Let $G_X$ denote the subgraph of $G =(V, E)$ formed
using the vertices $X\subseteq V$.  

\define{
A \emph{directed elimination tree} (DET) for a non-trivially strongly connected digraph $G=(V,E)$ is a rooted tree $DET_G = (T_V, T_E)$ where $T_V \subseteq 2^V \times V$ with the root $\tuple{V,v}$ such that $DET_G$ satisfies the following properties:
\begin{itemize}
    \item If $\tuple{H, v}\in T_V$, then $v\in H$
    \item There is no pair of distinct vertices of the form $\tuple{H, x}$ and $\tuple{H,y}$
    \item If $\tuple{H, v}$ is a node in $T$ and $G_H\setminus \set{v}$ has $j\ge 0 $ non-trivial SCCs $G_1, \ldots, G_j$, then $\tuple{H,v}$ has exactly $j$ children denoted as $ch(H,v) = \set{\tuple{G_1, v_1}, \ldots, \tuple{G_j,v_j}}$, for some $v_i, \ldots, v_j\in V$. 
\end{itemize}
} 

If $\tuple{G,v}$ is a node in a DET, we refer to $v$ as the \emph{elimination
point} for that node. We will use the concept of directed elimination trees with
FMPs by interpreting an FMP as a graph in the usual manner of representing
controllers as graphs. We will use FMPs and their graphs interchangeably and
clarify the distinction when required for clarity. In general, an FMP will yield
a \emph{directed elimination forest (DEF)} consisting of one directed
elimination tree for each strongly connected component in the FMP.

\begin{example}
\label{eg:mcnaughton}Fig.\,\ref{fig:fmp_det} (left) shows an example of an FMP
with edge labels representing actions in deterministic semantics. This example
was created by adding additional edges and edge labels to McNaughton's classic
example of a strongly connected digraph~\cite{mcnaughton69_loops}. In this
example, edge conditions are the default non-negativity conditions. The  right
subfigure  shows a DEF for the same FMP. 
\begin{figure}
  \subcaptionbox{\small Finite-memory policy $\Phi_1$\label{fig:fmp}}
  {\includegraphics[height=2.8in]{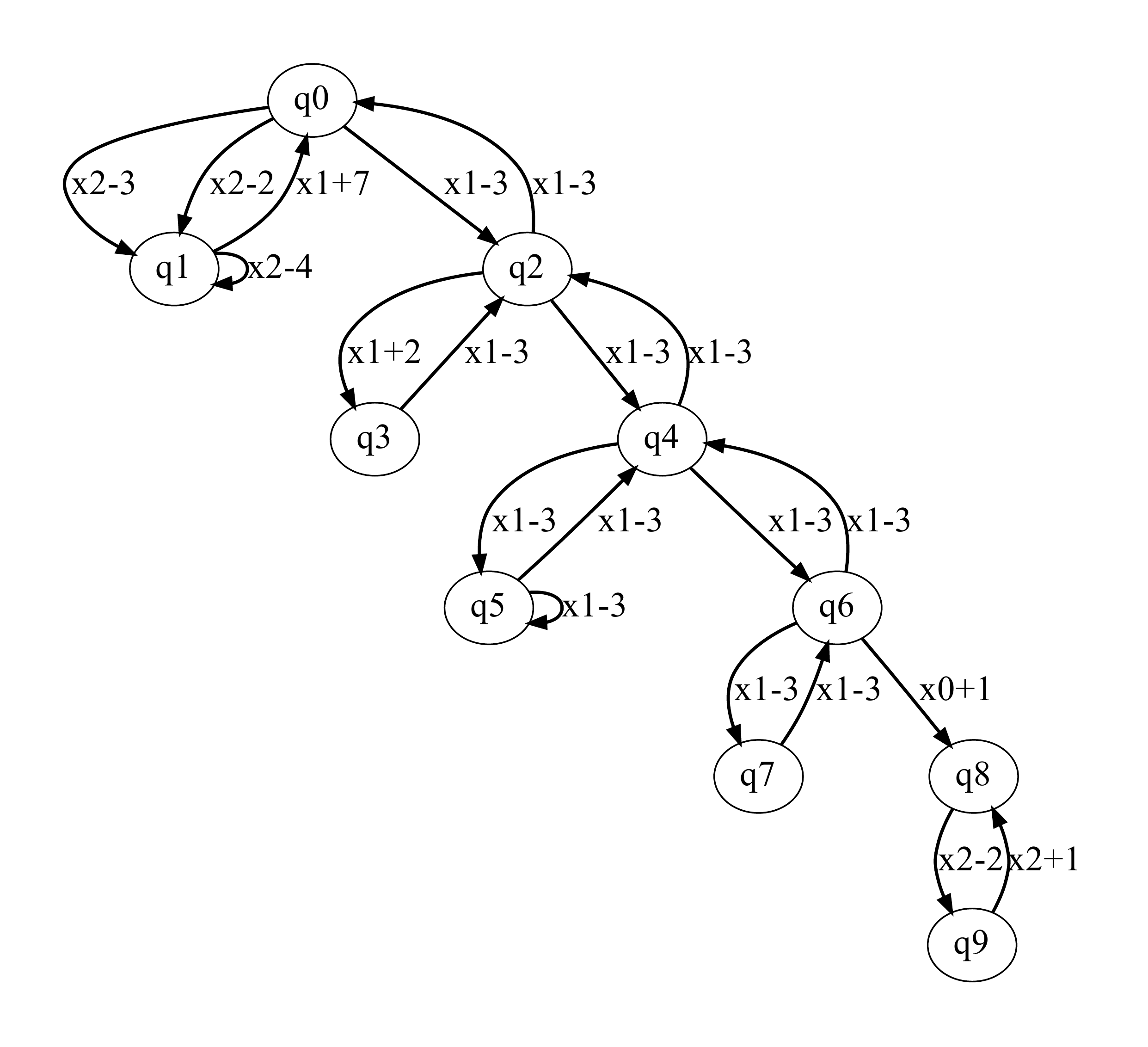}}
  \subcaptionbox{\small Directed elimination forest $D_{\Phi_1}$for
    $\Phi_1$\label{fig:det}}
   {\includegraphics[height=2.8in]{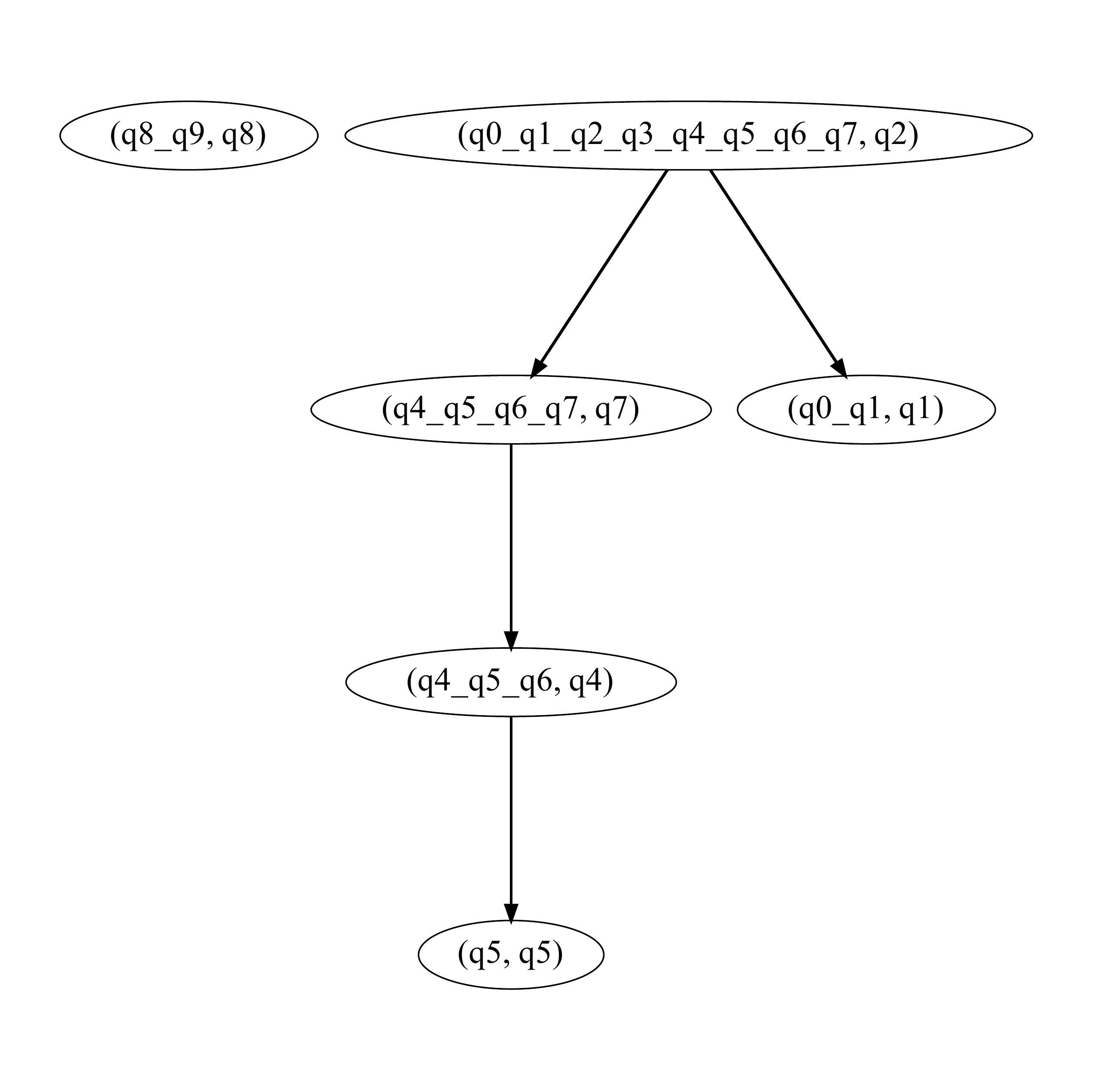}}
  \caption{\small A finite-memory policy and its directed elimination
    forest. Nodes in $D_{\Phi_1}$ are labeled $(V_G, v)$ where
    $V_G$ is the set of vertices for the component $G$. }\label{fig:fmp_det}
\end{figure}
\end{example}

Directed elimination trees are closely related to cycle ranks~\cite{eggan63_cr}.
In fact, the minimal height of directed elimination trees for a graph is  the
cycle rank of that graph~\cite{mcnaughton69_loops}. Cycle ranks are closely
related to multiple notions of the complexity of recurring patterns of behavior,
e.g., the star-height of regular expressions. 

Although the problem of computing a minimal directed elimination tree is
equivalent to the problem of computing the cycle rank of a graph and is
NP-complete, our algorithm does not require minimality. This is helpful because
polynomial-time divide-and-conquer algorithm can be used to compute a directed
elimination tree within a factor $O((log~ n)^{3/2})$ of the minimal height.
This result and a greater discussion of DETs
for structural categorizations of computational complexity can be found at
\cite{gruber12_loops}.

\paragraph{Termination Analysis Using DEFs} We now develop the formal concepts
required to develop a stronger test for termination under deterministic
concepts.

We wish to define an inductive, bottom-up process on the directed elimination
tree to derive a set of net-progress variables for the entire SCC. To do this,
we need to define the notion of a graph that considers sub-components as black
boxes that are guaranteed to have net-progress variables. Let $\tuple{G, v}$ be
a node in a directed elimination tree for an FMP policy, with children nodes
$\emph{ch}(G,v)=\set{\tuple{G_1, v_1}, \ldots, \tuple{G_k, v_k}}$. Recall that
by definition this implies that $G$ is an SCC such that upon removing $v$, $G$
is left with SCCs $G_1, \ldots, G_k$. 

We define the \emph{quotient graph of $G$} w.r.t. $\emph{ch}(G,v)$ (denoted as $\mod{G}{ch(G,v)}$) as the graph obtained by replacing each  $G_i$ in $\emph{ch}(G,v)$  with a new \emph{component vertex} $c_i$ while inheriting all the edges that led into, or out of $G_i$. Intuitively, $c_i$ encapsulates $G_i$ and $\mod{G}{ch(G,v)}$ can be thought of as an SCC of rank 1 over $G_i$'s.

\begin{example}
  \label{eg:phi_1_mod_graph}
  Fig.\,\ref{fig:mod_graph} shows the quotient graph
  $\mod{G}{\set{q4\_q5\_q6\_q7},\set{q0\_q1}}$ for the root node
  of the non-trivial tree in the DEF from Fig.\,\ref{fig:det}.
  Here $G$ is the
  subgraph of Fig.\,\ref{fig:fmp} induced by the qstates
  $\set{q0, q1, q2, q3, q4, q5, q6, q7}$. Since the root node
  $\tuple{G, q2}$ has two children nodes, these subgraphs are replaced
  by component vertices in the quotient graph. Quotient graphs help
  simplify reasoning about possible executions of a FMP policy. For
  instance, this quotient graph indicates that any infinite execution
  of $\Phi_1$
  that does not have an infinite suffix in $\set{q4, q5, q6, q7}$ and
  in $\set{q0, q1}$ (i.e., that does not have an infinite execution
  contained entirely within one of the two child components) must
  visit $q2$ infinitely often.

\begin{figure}
 \centering
    \includegraphics[height=2in]{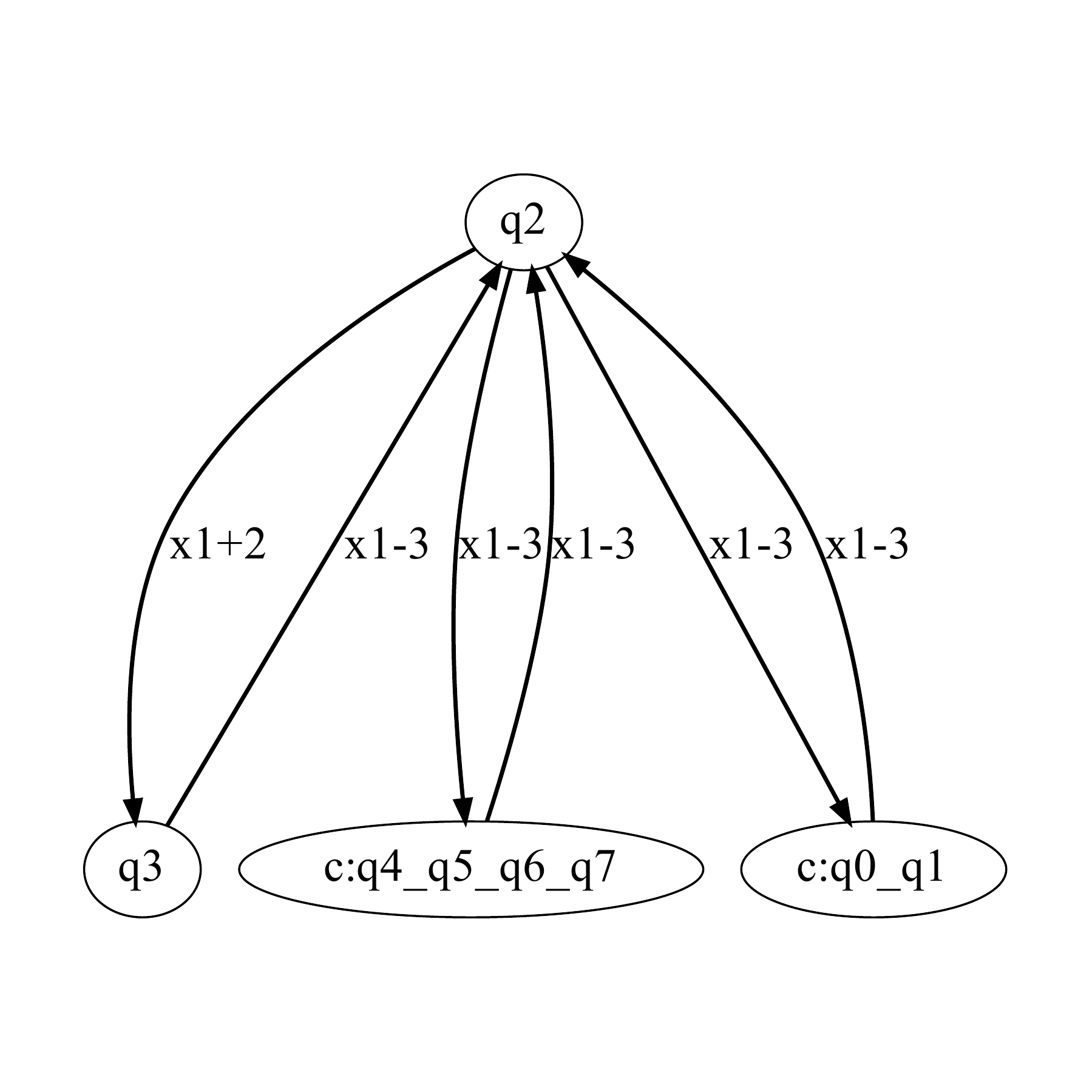}
  
  \caption{\small Quotient graph corresponding to the first node of the
    non-trivial tree in Fig.\,\ref{fig:det}. Component vertices are
  labeled using the prefix \texttt{c:}.}\label{fig:mod_graph}
\end{figure}
\end{example}

The hierarchical sieve algorithm developed in this paper formalizes the
intuition that if each path segment that can be executed infinitely often (while
ignoring edge conditions) has a net-progress variable that is not increased by
other inner loops, and the outer loop also has such a net-progress variable,
then the SCC cannot be executed indefinitely when edge conditions are
considered. More precisely, if $\mod{G}{\emph{ch}(G,v)}$ and each $G_i\in
\emph{ch}(G,v)$ has net-progress variables that are not increased by the other
children of $G$ or by $\mod{G}{\emph{ch}(G,v)}$ then $G$ itself has a
net-progress variable, which implies that executions within $G$ must terminate
or exit $G$. We develop this intuition by first computing a superset of paths
(not necessarily complete cycles) that can be executed infinitely often when
edge conditions are not considered.

Let $G^0_\Phi$ be the graph of an FMP $\Phi$ and  let $\tuple{G,v}$ be a node in
the DEF of $G^0_\Phi = (V_\Phi, E_\Phi)$.  Let  $\Pi_c(G,v)$ denote the set of
all cycle-paths in $\mod{G}{\emph{ch}(G,v)}$: paths that start and end at $v$ and visit $v$
exactly twice. $\Pi_c(G,v)$ denotes cycles in $\mod{G}{\emph{ch}(G,v)}$ that
start and end at $v$. No such path can visit any node (qstate or component
vertex) of   $\mod{G}{\emph{ch}(G,v)}$ other than $v$ twice
because by definition of the elimination tree and $\mod{G}{\emph{ch}(G,v)}$,
removing $v$ from $G$ renders $\mod{G}{\emph{ch}(G,v)}$ acyclic. 

We also need to consider paths that go through $G$ without encountering any
cycles. Let $I_G$ be the set of boundary vertices for  edges that enter $G$,
i.e., $\set{v\in G: \emph{ there is a } w \in V_\Phi\setminus G\emph{ such that
}  \tuple{w,v}\in E_\Phi }$. Similarly, let $O_G$ be the set of boundary
vertices for out-going edges from $G$. We define the set of through-paths for
$G$, $\Pi_{io}(G,G^0_\Phi)$, as the set of all cycle-free paths in $G$ that go
from a vertex in $I_G$ to a vertex in $O_G$.  Notice that this definition of
through paths considers all of the vertices in $G$ while the $\Pi_c$ considers
paths in $\mod{G}{\emph{ch}(G,v)}$. Considering all through-paths in this manner
allows for a more accurate algorithmic test for termination but this would be
difficult to do while considering all possible combinations of cycles in G, with
different possible numbers of iterations of each cycle. Thus we use quotient
graphs to hierarchically structure the cycles into a more manageable framework.
Let $\Pi(G,v) = \Pi_{io}(G,G^0_\Phi) \cup \Pi_c(G,v)$.

\begin{example} 
 \label{eg:paths} 
  Continuing with the running example, consider
$\mod{G}{\set{q4\_q5\_q6\_q7},\set{q0\_q1}}$, the quotient graph
(Fig.\,\ref{fig:mod_graph}) corresponding to the first node of the non-trivial
DET shown in Fig.\ref{fig:det}. In
this quotient graph the cycle paths are $(q2,q3,q2)$, $(q2,
c\!:\!q4\_q5\_q6\_q7, q2)$ and $(q2, c\!:\!q0\_q1, q2)$.  This node of the
DET represents the qstates
$\set{q0, q1, q2, q3, q4, \allowbreak q5, q6, q7}$. 
The only boundary
vertex of the subgraph defined by these qstates is q6, and thus this
subgraph has no through-paths. On the other hand, the subgraph defined by
$\tuple{\set{q4, q5, q6, q7}, q7}$,  a child node of
$\tuple{\set{q0, q1, q2, q3, q4, a5, q6, q7}, q2}$,  has $q4$ and $q6$ as boundary
vertices and two through-paths: $(q4, q6)$ and $(q6, q4)$.
\end{example}

Using the notion of net-decrease variables for a set of paths
(Def.\,\ref{def:decvars}), we define the net-decreased set for a node
$\tuple{G,v}$ in the elimination tree as the set of \emph{sets} of net-decrease
variables, with one set for each path in $\Pi(G,v)$. We denote this collection as
$\decvars_{\Pi(G,v)}$ (abbreviated as $\decvars_{G,v}$).  If any path in
$\Pi(G,v)$ has no net-decrease variables, then the set of net decreased
variables for that path is included in $\decvars_{G,v}$ as the empty set,
$\emptyset$. In the same vein, we define possibly increased variables for
$\tuple{G,v}$, $\incvarsa{G,v}$, as the set of variables that undergo a net
increase in at least one path in $\Pi(G,v)$.

\subsubsection{The \algo Algorithm} 
\label{sec:algo}
We now have all of the concepts required to
present the main algorithm of this paper.  Let $A \boxminus B$ denote the
element-wise set-minus operation where $A$ is a set of sets and $B$ is a set:
$A\boxminus B=\{x\setminus B: x\in A$\}. 

Let $G$ be the strongly connected graph of an FMP $\Phi$ and let $\tuple{G,v}$
be the root node of its DET. If the input FMP has multiple strongly-connected
components, we consider each component independently and prove termination by
proving that each component terminates. \algo (Alg.\,\ref{alg:determine})
inductively derives an estimate $\hat{\decvars}_{G,v}$ ($\eincvarsa{G,v}$) for the
set of variables that could undergo a net negative (positive) change in a path
in $\Pi(G,v)$  under default edge labels  in $G$ as follows.

\renewcommand{\mod}[2]{#1|_{#2}}

\begin{eqnarray}
\eincvarsa{G,v} &=& \incvarsa{G, v} \cup \bigcup_{(G_i, v_i)\in
\emph{ch}(G,v)}\eincvarsa{G_i, v_i} \\
 \hat{\decvars}_{G,v} &=&  \{\decvars_{G, v}\cup \bigcup_{(G_i, v_i)\in
 \emph{ch}(G, v)} \hat{\decvars}_{G_i, v_i} \} \boxminus \eincvarsa{G,v}
\end{eqnarray}

Since every directed elimination tree of a finite graph is finite and the base
case is defined in closed form, this process must terminate after $h$ recursions
where $h$ is the height of the directed elimination tree.

\begin{algorithm}
\small
  \DontPrintSemicolon
  \SetKwFunction{buildDecVars}{BuildDecVars}
  \SetKwFunction{buildIncVars}{BuildIncVars}
  \SetKwFunction{hsieve}{\algo}
  \SetKwProg{myalg}{Algorithm}{}{}
  \SetKwProg{myfunc}{Function}{}{}

\KwIn{FMP policy $G$}
\tcc*[l]{IV: set of variables increased by a path in $\Pi(G)$}
\tcc*[l]{pDV: set of sets of potential net-decrease variables per path in $\Pi(G)$}
\tcc*[l]{DV: set of sets of net-decrease variables per path in $\Pi(G)$}
\tcc*[l]{ZV: set of variables that undergo zero net change in a path in $\Pi(G)$}  

\nl progress=True\;
\nl \While{progress=True}{
    \nl $DET_G, \tuple{G,v} \gets $ DET of $G$ with root $\tuple{G,v}$ \;
    \nl  \emph{IV, ZV}$\gets$ \buildIncVars{$G, v, DET_G$}\;

    \nl decreasedVars $\gets$ \buildDecVars{$G, v, DET_G, \rm{\textit{IV}}$}\;
    \nl decreasingEdges $\gets$ edges that reduce decreasedVars not
    in ZV\;
    \nl remove decreasingEdges from $G$\;
    \nl \lIf{decreasingEdges $= \emptyset$ or $\set{} \not \in \emph{decreasedVars}$}{progress=False}
    }
\nl \eIf{$\{\}\in \emph{decreasedVars}$}{
\nl   \KwRet  ``unknown''}{
\nl    \KwRet ``terminating''}
  \setcounter{AlgoLine}{0}
\;

\myfunc{\buildIncVars{G, v, DET}}{
\nl \emph{IV}$\gets \emptyset$; \emph{ZV}$\gets \emptyset$\;

\nl \ForEach{$\pi \in \Pi(G,v)$}{
  \nl $\emph{IV} \gets \emph{IV} \cup \set{x: x \emph{ undergoes a net
      increase in }\pi}$\;
  \nl $ZV \gets ZV \cup \set{x: x \emph{ undergoes net
      zero change in } \pi}$\;
}

\nl \ForEach{$\tuple{G_i, v_i}\in ch(G,v)$}{
expand sets IV, ZV with output pair from \buildIncVars{$G_i, v_i, DET$}}

\nl \KwRet \emph{IV}, \emph{ZV}}
\;

\myfunc{\buildDecVars{G, v, DET, \emph{\textit{IV}}}}{
\nl $pDV \gets \emptyset$; $DV\gets \emptyset$; \;

        \nl \ForEach{$\pi \in \Pi(G,v)$}{
            \nl $pDV\gets pDV \cup \set{\set{x: x\emph{ undergoes net
                  decrease in } \pi}} $\;
 
        }

\nl \lForEach{$\tuple{G_i, v_i}\in ch(G,v)$}{
    \emph{pDV} $\gets$ \emph{pDV} $\cup$ \buildDecVars{$G_i, v_i, DET, \rm{\textit{IV}}$}
    }

\nl $\emph{DV}\gets \emph{pDV} \boxminus \emph{IV}$ \;

\nl \KwRet $DV$\;
}
  \caption{\algo}\label{alg:determine}
\end{algorithm} 

Intuitively, \algo generalizes the intuition behind Sieve and Progress-Sieve
algorithms. It attempts to identify every minimal \emph{path} that can occur
infinitely often in an execution in the form of $\Pi(G,v)$. The computation of
every such path is difficult because of the variable nature of executions
possible in FMPs with cycles, and the DET helps impose a hierarchical structure
for extracting them. 

\algo (Alg.\,\ref{alg:determine}) uses functions \texttt{BuildIncVars} and
\texttt{BuildDecVars} to compute $\eincvarsa{G,v}$ and $\edecvarsa{G,v}$ using
equations (1) and (2) respectively.  In cases where $\hat{\decvars}_{G,v}$
includes an empty set, it means that there is a path in $\Pi(G,v)$ whose
infinite execution cannot be ruled out yet. However, the computation of
$\edecvarsa{G,v}$ is conservative and can overestimate the impact of paths that
increase variables. Some of these paths can become unreachable after finitely
many steps of execution. To accommodate such situations, \algo computes the set
of variables that are only decreased by elements of $\Pi(G,v)$ (lines 5-6). It
then removes \texttt{decreasingEdges}, or edges that decrease such variables
(line 7), and reinvokes the computation of $\eincvarsa{G,v}$ and
$\edecvarsa{G,v}$ (the \texttt{while} loop starting in line 2). If in any
iteration of this loop, each set in $\hat{\decvars}_{G,v}$ is found to be
non-empty, this means that every path that can be executed infinitely often
includes at least one variable that undergoes a net negative change that is not
increased in any path that can be executed infinitely often.  This implies that
the FMP must terminate. Theorem \ref{thm:soundness} asserts this result
formally in Sec.\,\ref{sec:formal}. On the other hand, if the algorithm reaches
a stage where $\edecvarsa{G,v}$ contains an empty set and no edges were removed
in line 7, \algo terminates without asserting termination (lines 8 -- 10). In
practice, we check $\edecvarsa{G,v}$ for the absence of an empty set before
removing \texttt{decreasingEdges}  to allow the algorithm to assert termination
early if possible.

The Sieve family of algorithms constitute a special case of this general
principle: they consider each \emph{edge} as possibly occurring infinitely often
in an execution. Example 1 illustrates how this approach is unnecessarily
conservative under deterministic semantics -- even though each $\oplus$ can be
executed only in conjunction with two $\ominus$   operations, which leads to a
minimal recurring path with a net change of $-1$ in deterministic semantics,
Progress-Sieve and Sieve consider the $\oplus$ edge as an independently
executable edge that can undo the negative change. The hierarchical structure
developed in our approach makes it possible to extract minimal paths that
may occur infinitely often in the execution of FMPs with higher cycle ranks.

\begin{example}We continue with the running example $\Phi_1$ from Examples
  \ref{eg:mcnaughton} through \ref{eg:paths}.  \algo asserts that $\Phi_1$
  terminates, after two iterations of the   \texttt{while} loop
  shown in Alg.\,\ref{alg:determine}. Edges reducing the variable $x2$ are removed
  after the first iteration. Fig.\,\ref{fig:algo_phi1} shows the FMP obtained
  after this operation along with the DEF for this FMP. As seen in
  Fig.\,\ref{fig:algo_phi1}(a), every path involving x1 that could be executed
  infinitely often is a decreasing execution over x1 because the only positive
  change in x1 is accompanied with a larger negative change. The total time
  taken was less than 1.2s. The sieve algorithm fails to assert termination of
  $\Phi_1$ due to the absence of a net decrease variable.

\begin{figure}
  \subcaptionbox{\small Finite-memory policy $\Phi_{1}^2$ obtained
    after removing edges that decrease x2 from $\Phi_1$. \label{fig:phi1_fmp2}}
  {\includegraphics[height=2.8in]{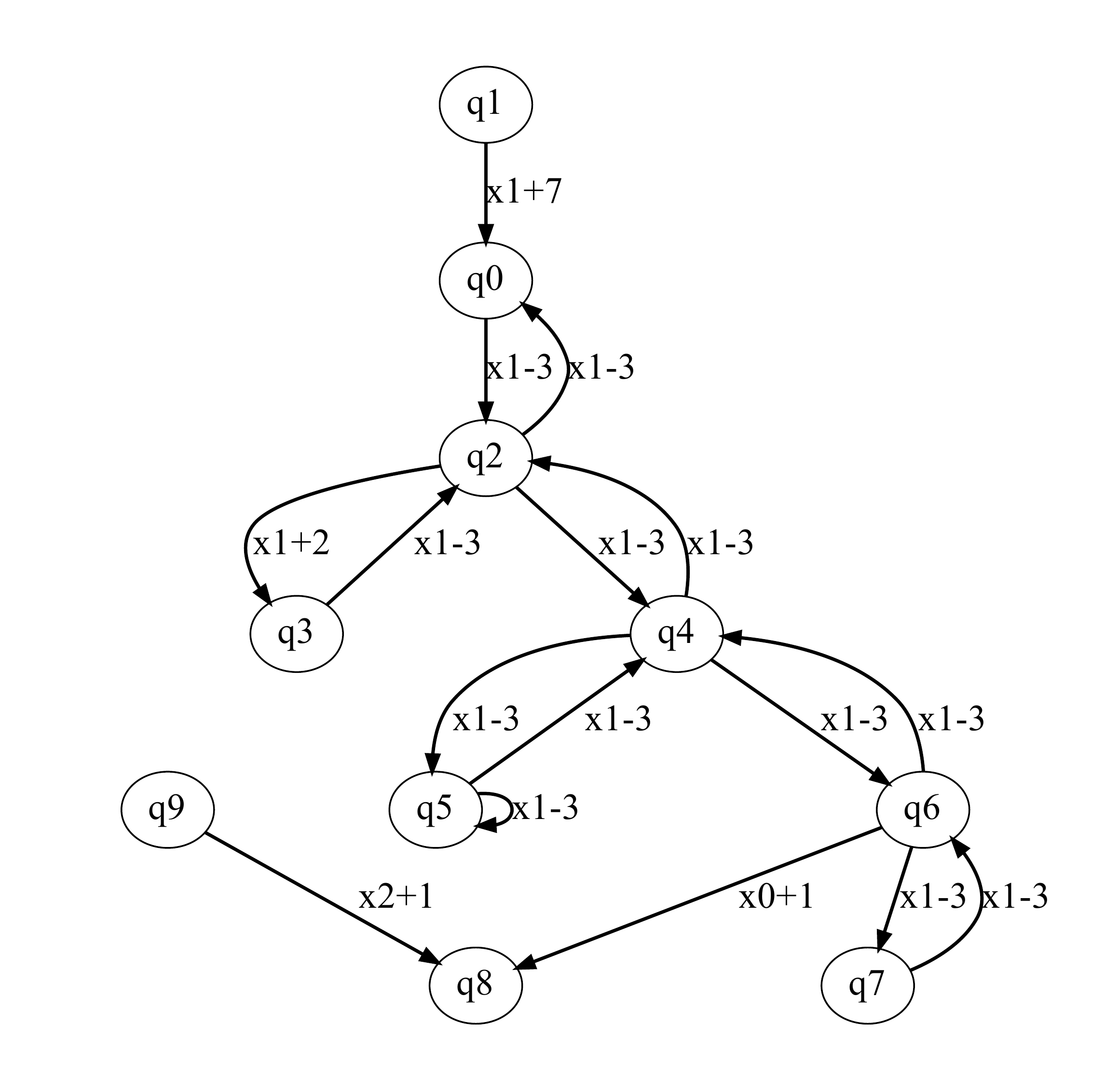}}
  \subcaptionbox{\small Directed elimination forest $D_{\Phi_1^2}$ for
    $\Phi_1^2$.\label{fig:phi1_det2}}
   {\includegraphics[height=2.8in]{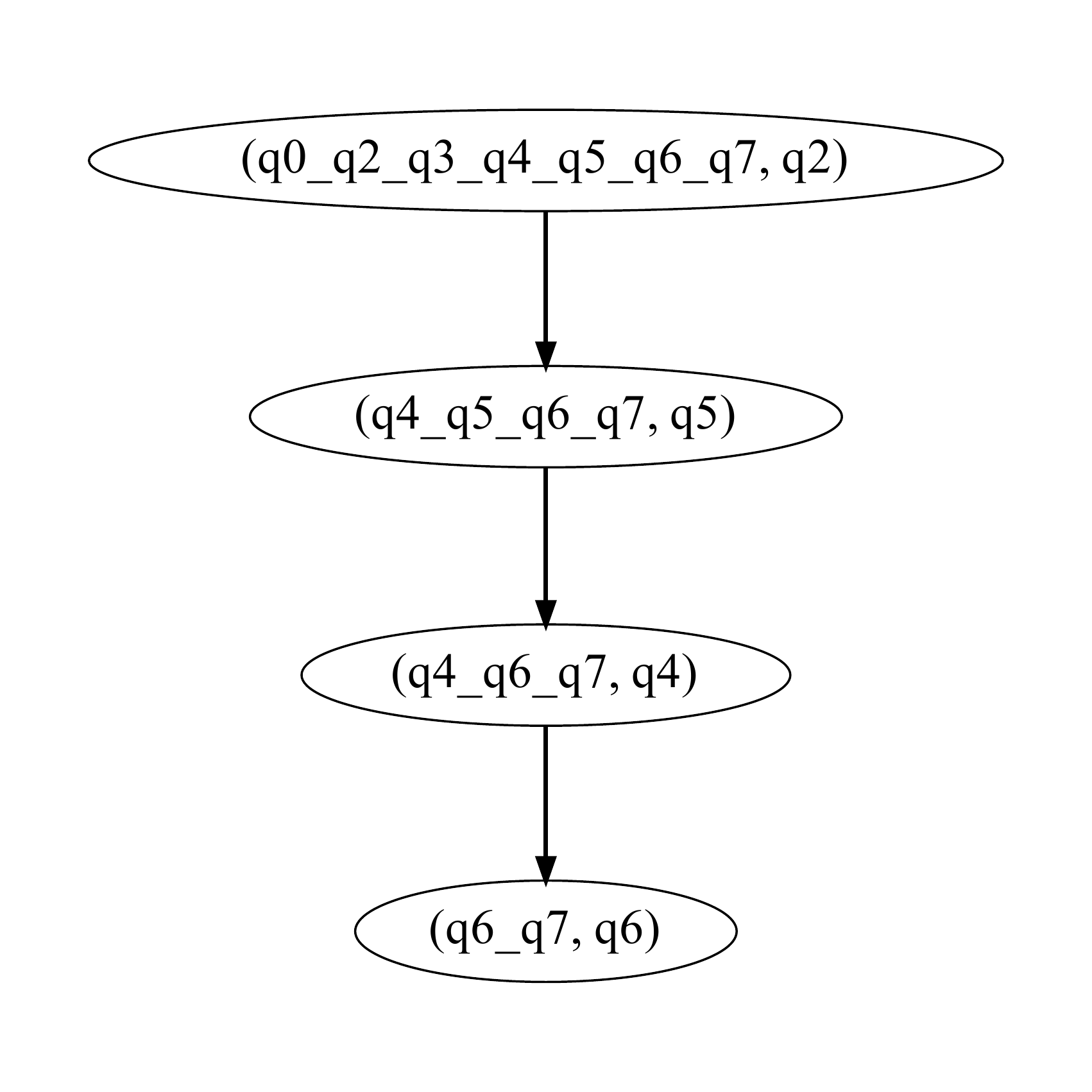}}
  \caption{\small Steps in \algo's execution on $\Phi_1$. }\label{fig:algo_phi1}
\end{figure}

\end{example}

The following sections formalize the key notions required to concretize these
intuitions and present our main formal and empirical results.

\section{Results}
\label{sec:results}
This section presents the key formal and empirical results obtained using the
hierarchical analysis framework and the \algo algorithm developed in
Sec.\,\ref{sec:hierarchical_framework}. We begin with formal results
(Sec.\,\ref{sec:formal}) followed by empirical results
(Sec.\,\ref{sec:empirical}) including  results from an implementation of of
\algo, illustrations of intermediate steps of its execution, and several
examples of complex, terminating FMPs that are beyond the scope of prior methods
for determining termination but were found to be terminating using the methods
developed in this paper. 

\subsection{Theoretical Analysis and Results}
\label{sec:formal}
In this section we develop new concepts and methods for formalizing the
intuitive reasoning behind \algo and its soundness. We also illustrate the proof
techniques made possible with this framework. These concepts refine  the
intuitive notions presented with the specification of \algo in
Sec.\,\ref{sec:algo}.

We consider the general case of FMPs with default edge conditions. Additional
edge conditions can limit the set of possible executions. Therefore, any FMP
that is a terminating policy without considering additional edge conditions
remains a terminating policy when they are considered. We focus on the case
where the default edge conditions lower bound all variables at zero. The
same arguments can be transposed to apply to situations where the lower bound is
different, as well as to situations where all variables have an upper bound. 

The main result of this section, Thm.\,\ref{thm:soundness}, establishes the
soundness of \algo. To obtain this result we need to develop additional
terminology that allows us to build inductive arguments about the desired
properties of $\eincvarsa{G}$ and $\edecvarsa{G}$ sets.

\define{
Let $\pi$ be an execution of an FMP policy. If  $\pi$ is finite, we say that
\emph{$\pi$ is an increasing (decreasing) sequence for $x$} iff the sum of the
effects of actions on $x$ in $\pi$ is  positive (negative). If $\pi$ is
infinite, we say that \emph{$\pi$ is an  increasing (decreasing) sequence for
$x$} iff there is an index $i$ such that for every $j>i$, there exists $k>j$
such that the net change on $x$ due to $\pi_{j, \ldots, k}$ is positive
(negative).}

Intuitively an infinite increasing sequence for $x$ will increase it
indefinitely while an infinite decreasing sequence for $x$ will decrease it
indefinitely.

For any path $\pi$ in a component $G$ of a FMP with a DEF $T$, we define
$\mod{\pi}{ch(G,v)}$ as a version of $\pi$ where every maximal contiguous
segment of nodes in $\pi$ that belong to a single child component $G_j$ is
replaced by the component vertex $c_j$ of $\mod{G}{ch(G,v)}$ defined using $T$.
We use the notation $v \rightsquigarrow w$ to denote the set of paths of the
form $v, v_1, v_2, \ldots, v_n, w $ in an underlying graph made clear from the
context, such that each $v_i\ne v_j$, $v_i\ne v$ and $v_i\ne w$ for distinct
$i,j\in \set{1, 2,\ldots, n}$.

To simplify notation, we use $\eincvarsa{G}$ to denote $\eincvarsa{G,v_0}$ where
$\tuple{G, v_0}$ is the root node of the DET for $G$. The following result
establishes that $\eincvarsa{G}$ is a superset  of variables that may be
increased indefinitely by $\Phi$.

\lem{\label{lem:increase_computation}Let $H$ be a FMP $\Phi$ for a domain $\dom= \tuple{\vars, \ell,
\mathcal{A}}$, let $T_H$ be the directed elimination forest for $H$  and let
$\tuple{G,v}$ be a node in $T_H$. Let $I_G$ and $O_G$ be the set of incoming and
outgoing boundary vertices in $G$.

Suppose that $\pi$ is an execution of $\Phi$ starting from $I_G$ such that
every qstate in $\pi$ is in $G$ and either $\pi$ ends in $O_G$ or $\pi$ is an
infinite execution sequence. If $\pi$ causes a net increase on a variable in
$x\in\vars$ under default edge conditions, then $x\in \eincvarsa{G,v}$. }

\prf{We prove the result by induction on the height of $\tuple{G,v}$. In the
proof we will consider  execution paths over control states (qstates) in $\Phi$
regardless of the problem states because we wish to prove the result under
default edge conditions. 

\paragraph{Base case} Suppose $\tuple{G,v}$ is a leaf node and assume for a
proof by contradiction that $x$ is a variable such that $\pi$ causes a net
increase in $x$ but $x\not\in\eincvarsa{G,v}$. We first consider the case that
$\pi$ is a finite path that reaches $O_G$. In such a situation, $\pi$ may or may
not include repetition of control states. If there are no repetitions, $\pi\in
\Pi_{io}(G,v)$ and every variable that undergoes a net increase is included in
$\incvarsa{G,v}$ by definition. If $\pi$ includes repetition, then $\pi$ is of
the form  $q_i\rightsquigarrow(v \rightsquigarrow v)^*  \rightsquigarrow q_o$
where $q_i\in I_G$ and $q_o\in O_G$, and $\rightsquigarrow$ represents paths
without repetitions as noted above. This is because $G$ is a leaf node, which
implies that removing $v$ eliminates all cycles in $G$. In other words, every
cycle in $G$ must pass through $v$. If any of the paths going from $v$ to $v$ in
$G$ created a net increase in $x$, $x$ would have been included in
$\incvarsa{G,v}$ by definition. Thus, no segment of $\pi$ of the form $v
\rightsquigarrow v $ can create a net increase in $x$. This implies $\pi$ must
be a finite path of the form $q_i \rightsquigarrow v \rightsquigarrow q_o$ that
has a net positive change on $x$. However, such paths are included in
$\Pi(G,v)$, which implies that $x$ should be in $\incvarsa{G,v}$, leading to a
contradiction.

Suppose on the other hand that $\pi$ is an infinite path that is an increasing
sequence for $x$. Then $\pi$ must be of the form: $q_i \rightsquigarrow
(v\rightsquigarrow v)^*$. Since variables that undergo a net positive change in
$v \rightsquigarrow v$ are included in $\incvarsa{G,v}$ and $x\not\in
\incvarsa{G,v}$, the only component that can increase $x$ is the finite prefix
$q_i \rightsquigarrow v$. But this implies that after the index $i$ denoting the
length of this prefix, $x$ stops increasing in $\pi$ and $\pi $ is not an
increasing sequence for $x$.

\paragraph{Inductive case} Suppose the hypothesis holds for DEF nodes at height
at most $k$, i.e., if $\pi$ is an execution that continues indefinitely within
$G$ or reaches $O_G$ and $\pi$ is an increasing sequence in a variable $x\in
\vars$, then $x\in \eincvarsa{G,v}$ for all tuples of the form $\tuple{G,v}$ with
height at most $k$. 

Let $\tuple{G,v}$ be at height $k+1$. We need to prove that if a variable
$x$ undergoes a net increase as a result of a path $\pi$ with the properties
noted in the premise, then $x\in \eincvarsa{G,v}$. 
Suppose this is not true and there exists a path $\pi$ in $G$ that starts from
$q_i\in I_G$, and is an increasing sequence for $x$ but $x\not\in
\eincvarsa{G}$. Since all paths of the form $q_i \rightsquigarrow q_o$ in $G$
where $q_o\in O_G$ are considered in the $\Pi_{io}(G,v)$ component of
$\incvarsa{G,v}$, $\pi$ cannot be such a path and $\pi$ must include a cycle.

In general, $\pi$ may include qstates from its child SCCs. However, since each
child is of height at most $k$ and $x\not\in\eincvarsa{G,v}$, the inductive
hypothesis and the fact that $\eincvarsa{G,v}$ includes $\eincvarsa{G_i,v_i}$ for its
children SCCs (by Eq. 1) implies that no path lying entirely within a child
component can be increasing for $x$.  
Thus, if $\pi$ is an increasing sequence for  $x$, $\mod{\pi}{ch(G,v)}$  must be
an infinite sequence of the form $q_i \rightsquigarrow (v \rightsquigarrow v)^*
$ in $\mod{G}{ch(G,v)}$ or a finite sequence of the form $q_i \rightsquigarrow
(v \rightsquigarrow v)^* \rightsquigarrow q_o$ in $\mod{G}{ch(G,v)}$. 

In the former case,  $\pi$'s suffix of the form $(v \rightsquigarrow v)^*$ must be an increasing sequence for $x$.  The induction hypothesis allows us to conclude that each component of $\pi$ that is within $c_i$ must create a net zero or a net negative change on $x$ because $x\not\in\eincvarsa{G,v}$. Thus we can think of each $c_i$ as a pseudo-qstate such that paths within $c_i$ cannot possibly create a net increase in $x$. But this implies that $x$ must undergo a net positive change in at least one path of the form $v\rightsquigarrow v$ in $\mod{G}{ch(G,v)}$, which implies $x\in \incvarsa{G,v}$ by definition of $\incvarsa{G,v}$, and we reach a contradication.

On the other hand, if $\pi$ is a finite sequence of the form $q_i
\rightsquigarrow (v \rightsquigarrow v)^* \rightsquigarrow q_o$ where $q_o\in
O_G$ in $\mod{G}{ch(G,v)}$, the reasoning is similar. Neither the segments of
$\pi$ within $G$'s child components nor any segment of the form $v
\rightsquigarrow v$ can be increasing for $x$ because of the inductive
hypothesis and the consideration of $\Pi_{io}(G,v)$ in $\incvarsa{G,v}$. This
implies that a sequence of the form $q_i \rightsquigarrow v \rightsquigarrow
q_o$ must create a net positive change in $x$, but that implies that $x\in
\incvarsa{G,v}$ by definition. Again, this contradicts $x\not \in
\eincvarsa{G,v}$. Thus, $x$ must be in $\eincvarsa{G,v}$. }

The result above shows that the estimated set $\eincvarsa{G,v}$ is a superset of
variables  that may be increased in a possible execution of the FMP. For
decreases, we require a tighter result: that \emph{every} possible execution
creates a net decrease on some variable. If a subset of the variables that
undergo a net decrease in every possible execution does not include the superset
of variables that may be increased ($\eincvarsa{G,v}$) in any possible execution,
then we know that every execution must terminate in a finite number of steps due
to the default edge conditions. The following theorem establishes what we need
with net-decreased variables.

\lem{\label{lem:decrease_computation}Let $H$ be the graph of a FMP $\Phi$ for a domain $\dom=
\tuple{\vars, \ell, \mathcal{A}}$, let $T_H$ be the directed elimination forest
for $H$  and let $\tuple{G,v}$ be a node in $T_H$.  Let $I_G$ and $O_G$ be the
set of incoming and outgoing boundary vertices in $G$. 

If $\emptyset \not \in \edecvarsa{G,v}$ and $\pi$ is an execution of $\Phi$ starting
from $I_G$ such that  every qstate in $\pi$ is in $G$ and either $\pi$ ends in
$O_G$ or $\pi$ is an infinite execution sequence, then $\pi$ is a net-decreasing
execution for at least one variable in $\vars$.}

\begin{proof} By induction on height of the tuple $\tuple{G,v}$ in $T_H$.

\paragraph{Base Case} Let $\tuple{G,v}$ be a leaf tuple such that the empty set,
$\emptyset$, is not in $\edecvarsa{G,v}$ and let $\pi$ be an execution sequence of the form
described in the premise.  By Eq. (1), the premise of the theorem and the fact
that Eq. (2) cannot remove empty-sets from $\edecvarsa{G,v}$, we know that
$\emptyset \not \in \edecvarsa{G,v}$ and consequently, that $\emptyset \not \in
\decvarsa{G,v}$. Suppose $\pi$ is of the form $q_i \rightsquigarrow q_o$ with
$q_i\in I_G$ and $q_o\in O_G$. By definition of $\decvarsa{G,v}$, this means
that all paths of the form $q_i \rightsquigarrow q_o$ create a net negative
change in at least one variable and we have the desired result. Suppose on the
other hand that $\pi$ is a finite sequence of the form $q_i \rightsquigarrow (v
\rightsquigarrow v)^* \rightsquigarrow q_o$. Because $\emptyset \not \in
\decvarsa{G,v}$, $\decvarsa{G,v}$ includes,  for each through path from $I_G$ to
$O_G$ as well as for each $v \rightsquigarrow v$ path in $\mod{G}{ch(G,v)}$ a
non-empty set of variables that undergo a net decrease under that path. In this
case $\mod{G}{ch(G,v)} = G$. Thus each of the three types of segments in $\pi$
($q_i \rightsquigarrow v$, $v\rightsquigarrow v$, and $\rightsquigarrow q_o$)
must create a net negative change in  at least one variable that is not in
$\eincvarsa{G}$ (recall that we use the abbreviated form $\eincvarsa{G}$ to
refer to $\eincvarsa{G,v_0}$ where $v_0$ is the root of the DET) by Eq. (2) and the premise that $\emptyset\not\in \edecvarsa{G,v}$.
This implies that each of these segments creates a net negative change in at
least one variable for which $\pi$ is not an increasing sequence, and thus $\pi$
must create a net negative change on at least one variable. Finally, suppose
$\pi$ is an infinite sequence of the form $q_i \rightsquigarrow (v
\rightsquigarrow v)^*$. Through arguments identical to those for the previous
form of $\pi$, 
all  segments of the form $v \rightsquigarrow v$ must create net negative changes in at least one variable. This implies that $\pi$ must be a decreasing sequence for at least one variable.

\mysssection{Inductive Case} Suppose the result holds for tuples at height at
most $k$. Let $\tuple{G,v}$ be at height $k+1$ and $\emptyset \not \in
\edecvarsa{G,v}$. We need to prove that $\pi$ is a net decreasing sequence for at
least one variable.

We consider the finite case first. If $\pi$ is finite, it may be of form $q_i
\rightsquigarrow q_o$ in $G$ or of the form  $q_i \rightsquigarrow (v
\rightsquigarrow v)^*\rightsquigarrow q_o$ in $\mod{G}{ch(G,v)}$ with $q_i\in
I_G$ and $q_o\in O_G$. Let $\pi$ be of the form $q_i \rightsquigarrow (v
\rightsquigarrow v)^*\rightsquigarrow q_o$. The argument for $\pi$ of the form
$q_i \rightsquigarrow q_o$ is similar. Since $\emptyset\not\in \edecvarsa{G,v}$,
and the empty set cannot be removed as a result of the $\boxminus$ operation,
Eq. (2)  implies $\emptyset \not \in \decvarsa{G,v}$ and $\emptyset \not \in
\edecvarsa{G_i,v_i}$ for each child $\tuple{G_i, v_i}$. This implies that the
components of $\pi$ within child SCCs must be decreasing sequences because of
the inductive hypothesis. Furthermore, all paths of the form $q_i
\rightsquigarrow v \rightsquigarrow q_o$ and all paths of the form $v
\rightsquigarrow v$ in $\mod{G}{ch(G,v)}$  must decrease at least one variable
because the set of variables that undergo a net negative change due to each such
path is added as a set into $\decvarsa{G,v}$ and  $\emptyset\not\in
\decvarsa{G,v}$. Furthermore, $\emptyset \not \in \edecvarsa{G,v}$ implies that
removing $\eincvarsa{G}$ from each of these sets leaves them non-empty. By Lemma
\ref{lem:increase_computation}, $\pi$ cannot result in a net increasing change
on any variable not in $\eincvarsa{G}$. Thus $\pi$ is composed of segments, each
of which causes a net decrease in at least one variable and each such variable
is such that it does not undergo a net increase in the entire finite sequence
represented by $\pi$. This implies that $\pi$ is a net decreasing sequence for
each of the variables in $\hat \tau_G$.

 If $\pi$ is infinite, it may be such that it has an infinite suffix in one of
 the child components of $G$ or  such that $\mod{\pi}{ch(G,v)}$ is of the form
 $q_i \rightsquigarrow (v \rightsquigarrow v)^*$ in $\mod{G}{ch(G,v)}$, because
 these are the only possible infinite execution sequences in $\mod{G}{ch(G,v)}$
 that do not have an infinite suffix within a single child component. Suppose
 $\pi$ has an infinite suffix in a child component $\tuple{G_i, v_i}$ of $G$.
 Since $\edecvarsa{G_i,v_i} \subseteq \edecvarsa{G,v}$ by Eq. (2) and $\eincvarsa{G}$
 can never contain $\emptyset$ as an element, if $\emptyset \not \in \edecvarsa{G,v}$
 then by Eq. (2),  $\emptyset \not \in \edecvarsa{G_i,v_i}$. The inductive
 hypothesis implies that such a suffix must be a net-decreasing sequence for at
 least one variable.

Suppose $\mod{\pi}{ch(G,v)}$ is of the form  $q_i \rightsquigarrow (v
\rightsquigarrow v)^*$ in $\mod{G}{ch(G,v)}$. Consider the infinite segment
$\pi_{vv}$ that is of the form $(v \rightsquigarrow v)^*$. By inductive
hypothesis we know that every finite segment of $\pi$ that lies within a child
component $G_i$ decreases at least one variable because such a segment
constitutes an $I_{G_i}$ to $O_{G_i}$ path, and Eq. 2 implies that
when $\emptyset\not \in \edecvarsa{G,v}$, $\emptyset \not \in \edecvarsa{G_i,v_i}$
because $\eincvarsa{G}$ cannot contain the set $\emptyset$ as an element.
Furthermore, each path of the form $v \rightsquigarrow v$ in $\mod{G}{ch(G,v)}$
decreases a variable because such paths are included in $\Pi(G,v)$ and
$\emptyset\not\in \decvarsa{G,v}$;  each of these decreased variable sets remain
non-empty when $\eincvarsa{G}$ is removed from $\edecvarsa{G,v}$. This implies
that when $\pi$ enters the $(v \rightsquigarrow v)^*$ segment, each  $v
\rightsquigarrow v$ segment in $\mod{G}{ch(G,v)}$ and each foray into a child
component of $G$ results in a net decrease on at least one variable that is not
increased by $\pi$ after a finite prefix because $\eincvarsa{G}$ contains all
variables that could possibly  undergo net increases in $\pi$ (Lemma 1). Since
there are only finitely many variables, the segment of $\pi$ in $v
\rightsquigarrow v$ must create a decreasing sequence on at least one such
variable.
\end{proof}

The following result provides the first test of termination using the methods
developed in this paper. This result shows that assertions of termination
obtained without removing any edges (in executions of \algo without using lines
6 and 7) are sound. 

\thm{\label{thm:direct_termination}
Let $\Phi$ be a FMP for a domain $\dom= \tuple{\vars, \ell, \mathcal{A}}$.
If $\emptyset \not \in \edecvarsa{G,v}$ then $\Phi$ is a terminating policy. }

\prf{We prove the result assuming that $\Phi$ has only default edge conditions
because if $\Phi$ has additional edge conditions, they will further limit the
set of executions possible, thereby maintaining termination.  By Lemma 2, we
know that every execution $\pi$ of $\Phi$ is a decreasing sequence on at least
one variable $x\in \vars$. Suppose $\pi$ is an infinite sequence.  When started
with a finite value for $x$, every execution of $\pi$ will  eventually lead to
an action that attempts to reduce $x$ below zero, at which point the default
edge condition will lead to the end of that execution and we arrive at a
contradiction. }

In cases where $\emptyset$ belongs to $\edecvarsa{G,v}$, we cannot assert
termination directly but the quantities computed in Equations 1 and 2 allow us
to simplify the input policy by remove certain edges (lines 6 and 7 in \algo)
and then continue the analysis. The following results  establish that certain
edges can be executed only finitely many times, setting up the stage for
removing them and creating a simpler policy  for the purpose of termination
analysis. In particular, if some variables satisfy stronger conditions than
membership in an element of $\edecvarsa{G,v}$, then the following two results
show that  edges involving such variables can be executed only finitely many
times, and
thus removed from consideration when we wish to focus on infinite executions.

\lem{
 \label{lem:finite_increase} Let $\Phi$ be a FMP for a domain $\dom= \tuple{\vars, \ell, \mathcal{A}}$.
Let $H_\Phi$ be its graph and let $T_H$ be its DEF and let $\Pi(H) =
\bigcup_{\tuple{G,v}\in T_H} \Pi(G,v)$. Suppose $x\in\vars$ is such that every
path  $\rho \in \Pi(H)$ that includes an edge with an action that affects $x$ is
such that $\rho$ creates a net negative change in $x$. 

Let $\tuple{G,v}$ be a node in $T_H$.  Let $I_G$ and $O_G$ be the set of
incoming and outgoing boundary vertices in $G$. If $\pi$ is a finite path that
starts in $I_G$ and ends in $O_G$, then $\pi$ cannot increase $x$.} \prf{By
induction on the height of $\tuple{G,v}$ in $T_H$.

\paragraph{Base case} Let $\tuple{G,v}$ be a leaf node. Since $\pi$ is a finite
path, it must be of the form $q_i \rightsquigarrow (v \rightsquigarrow v)^*
\rightsquigarrow q_o$ in $G$ where $q_i\in I_G$ and $q_o \in O_G$. Since all
segments of the form $v\rightsquigarrow v$ are considered in $\Pi(G,v)$. This
leaves a segment of the form $q_i \rightsquigarrow  q_o$, which is also
considered in $\Pi(G,v)$ as a part of $\Pi_{io}(G,v)$.  Since $\Pi(G,v)$ is
included in $\Pi(H)$, and no such segments increase $x$, $\pi$ cannot increase
$x$. 

\paragraph{Inductive case} Suppose the result holds for all nodes at height at
most $k$ in $T_H$ and $\tuple{G,v}$ is at height $k+1$. $\pi$ must be of the
form $q_i \rightsquigarrow (v \rightsquigarrow v)^* \rightsquigarrow q_o$ in
$\mod{G}{ch(G,v)}$ with $q_i \in I_G$ and $q_o\in O_G$. All segments of
$\mod{\pi}{ch(G,v)}$ of the form $q_i \rightsquigarrow q_o$ and
$v\rightsquigarrow v$ in $\mod{G}{ch(G,v)}$  are included in $\pi(G,v)$ and thus
these segments cannot create an increase in $x$. Furthermore, the segments of
$\pi$ within each child component are finite and satisfy the premises of the
theorem because $\Pi(H)$ includes $\Pi(G_i, v_i)$ for all nodes $\tuple{G_i,
v_i} \in T_H$.  Since these child components are at height at most $k$, no such
segment of $\pi$ can increase $x$ and we have the result. }

\thm{\label{thm:finite_decreases} Let $\Phi$ be a FMP for a domain $\dom= \tuple{\vars, \ell, \mathcal{A}}$.
Let $H_\Phi$ be its graph and let $T_H$ be its DEF. Let $\Pi(H) =
\bigcup_{\tuple{G,v}\in T_H} \Pi(G,v)$. If $x\in\vars$ is such that every path
$\rho \in \Pi(H)$ that includes an edge with an action that affects $x$ is such
that $\rho$ creates a net negative change in $x$,  then edges in $\Phi$ that
decrease $x$ can be executed only finitely many times in every execution of
$\Phi$. } 

\prf{By Eq. 1 and the definitions of $\incvarsa{}$ and $\eincvarsa{}$,
$x$ cannot be in $\eincvarsa{H}$. By Lemma 1, no infinite execution of $\Phi$ can
be an increasing sequence for $x$. Suppose an edge that decreases $x$ is
executed infinitely often (perhaps because a subsequent edge  increases $x$).
Let $\pi$ be such an infinite execution sequence in a subgraph $G$ corresponding
to the tuple $\tuple{G,v}$ in $T_H$.

\paragraph{Base case} $\pi$ must be of the form $q_i \rightsquigarrow (v
\rightsquigarrow v)^*$. $x$ cannot possibly have a net positive or net zero
change in segments of the form $v\rightsquigarrow v$ because such segments are
considered in $\Pi(G,v)$ and such segments create a net negative change on $x$
by the premise of the theorem. Thus $\pi$ can have only finitely many
occurrences of edges that reduce $x$ in all of its segments of the form
$v\rightsquigarrow v$ due to the default edge conditions. $x$ may undergo a net
positive change in the segment $q_i\rightsquigarrow v$ but this segment can have
only finitely many decreases. Thus we have a contradiction.

\paragraph{Inductive case} Suppose the premise holds for $\tuple{G_i, v_i}$ at
height at most $k$ and $\tuple{G,v}$ is at height $k+1$. $\mod{\pi}{ch(G,v)}$
must be of the form $q_i \rightsquigarrow c_i$ in $\mod{G}{ch(G,v)}$ with an
infinite segment within a child SCC $G_i$, or of the form $q_i\rightsquigarrow
(v\rightsquigarrow v)^*$. In the former case, the induction hypothesis implies
that the segment of $\pi$ within $G_i$ will have only finitely many occurrences
of edges that decrease $x$. 

Suppose $\mod{\pi}{ch(G,v)}$ is of the form $q_i\rightsquigarrow
(v\rightsquigarrow v)^*$ in $\mod{G}{ch(G,v)}$. No  segment within child
components will increase $x$ because $x$ is reduced by every path in $\Pi(H)$
that affects it, by the premise of this theorem  and Lemma \ref{lem:finite_increase}. Segments of $\pi$
that lie within child components will include only finitely many executions of
edges that reduce $x$ because these segments will be finite (execution returns
to $v$ infinitely often in this case).  Furthermore, paths of the form $v
\rightsquigarrow v$ are considered in $\Pi(G,v)$ and all such paths create a net
reduction in $x$, as noted in the premise. Thus $\pi$ can decrease $x$ only
finitely many times before further decreases are prevented due to the default
edge conditions.}

\cor{When the premise of Theorem 4 holds for a FMP $\Phi$ and a variable $x$,
every execution $\pi$ of $\Phi$ has a last step $i$ after which $\pi$ does not
use edges that decrease $x$. }

Theorem \ref{thm:finite_decreases} and Corollary 1 above allow us to design the
complete \algo, which proceeds by iteratively computing $\edecvarsa{G,v}$, and
if $\emptyset \in \edecvarsa{G,v}$, removing edges that reduce variables that
satisfy the premise of Theorem \ref{thm:finite_decreases} (lines 6 and 7 in
\algo), and repeating the process with the reduced graph. This process
generalizes the intuition behind Sieve and Progress-Sieve, which are special
cases obtained by replacing $\Pi(G,v)$ with the set of all  edges of $G$.

\thm{\label{thm:soundness} If \algo returns ``terminating'' when called with a
FMP $\Phi$ then all executions of $\Phi$ are finite.} 

\prf{ \algo computes the
set $\edecvarsa{G,v}$ for the graph of $\Phi$. Theorem \ref{thm:finite_decreases}
establishes the desired result if line 11 is reached without executing lines 6
and 7 (pruning of decreasing edges). Execution of lines 6 and 7 results in a
smaller policy $\Phi'$. Suppose \algo returns ``terminating'' for $\Phi'$ and
yet $\Phi$ has an infinite execution. By Corollary 1, after a finite prefix, this infinite execution
never decreases variables that are removed in lines 6 and 7. The infinite suffix
after this prefix does not use any edges that were removed from $\Phi$ to create
$\Phi'$, so this must be an infinite execution for $\Phi'$ as well, but this
contradicts the consequence of Theorem \ref{thm:direct_termination} for $\Phi'$.
Therefore if \algo returns ``terminating'' after finitely many iterations of the
main loop (line 2), $\Phi$ and all of the intermediate pruned policies are
terminating policies. }

The analysis presented above can also be generalized to apply to FMPs under
qualitative semantics where the absolute value of all decreases ($\ominus$) is
bounded below to be at least $\delta_{\ominus}$ and all increases ($\oplus$) are
bounded above  to be at most $\delta_{\oplus}$, so that we can obtain
conservative estimates of the net change due to a sequence with $x$ decreases
and $y$ increases as $\le x\delta_{\oplus} - y\delta_{\ominus}$. In such a case
we would replace all net changes with $x\delta_{\oplus} - y\delta_{\ominus}$.
The analysis conducted above uses the same value $  x\delta_{\oplus} -
y\delta_{\ominus}$ but for deterministic semantics with $\oplus =
\delta_\oplus=+1$  and $\ominus= -\delta_\ominus=-1$.

The time complexity of \algo depends on  the number of nodes in the DET and the
number of vertices in the quotient graphs. The process of computing all paths
between two vertices in a graph is $O(2^V)$ and this is done  $O(k)$ times in
\emph{BuildDecVars} and $\emph{BuildIncVars}$ where $k$ is the number of nodes
in the DET. Let   $q$ be the maximum number of vertices in  quotient graphs
and let $l$ be the number of segments included in the $\Pi(G,v)$ sets, which are
subsets of non-repeating paths in graphs of at most $q$ vertices. Thus the
runtime complexity is $O(k2^q + kl) \equiv O(k2^q)$.  In practice we found $q$
to be significantly less than the total number of qstates in FMPs with multiple
strongly connected components. 

The approach presented in this section is complementary to methods based on the
synthesis of ranking functions for linear arithmetic simple-while loop
programs~\cite{terminator} and the two approaches can be used in conjunction by
using our approach to produce a finite set of possible paths that need to be
validated.

\subsection{Empirical Analysis and Results}
\label{sec:empirical}
We developed a preliminary implementation of \algo in Python. All experiments
were carried out on a laptop with a 3.1GhZ Quad-Core Intel Core i7 processor and
16GB of RAM. We tested the implementation using custom generated FMPs as well as
randomized, auto-generated FMPs.  This implementation caches quotient graphs
with DEF nodes but several other optimizations are possible, e.g., by computing
the sets constructed in BuildIncVars and BuildDecVars simultaneously for each
DEF node. DEFs are computed in a straightforward manner by  identifying all
SCCs, removing a randomly selected vertex from each of the SCCs and repeating
this process on each of the resulting graphs. All  reported times include the
time taken for generating DEFs in this manner.

We present several  randomly generated FMP policies with increasing numbers
of nodes where the sieve algorithm is unable to assert termination due to the
absence of net decrease variables but \algo asserts termination. We limited the
randomly generated policies to use a small number of variables to make the
analysis of termination harder as with larger numbers of variables there tend to
be fewer instances of the same variable being increased and decreased in the
same strongly connected component. Further analysis of the ratio of variables to
control states and its relationship with difficulty of asserting termination is
a promising direction in the study of termination assessment of FMPs.

The runtime for this Python implementation was less than 2-3s in all of our
experiments. However it can be difficult to  randomly generate interesting
policies that can be manually verified as terminating policies, especially with
more than 5-6 control states. Currently, generating such policies (with and
without consideration of termination properties) is one of the major thrusts in
research on generalized planning. \algo can be used for analysis of candidate
FMPs in algorithms for the synthesis or learning of generalized plans.

\begin{figure}[h]
\centering
    \subcaptionbox{\small $\Phi_{2}$ \label{fig:phi_fmp2}}
        {\includegraphics[height=1.5in]{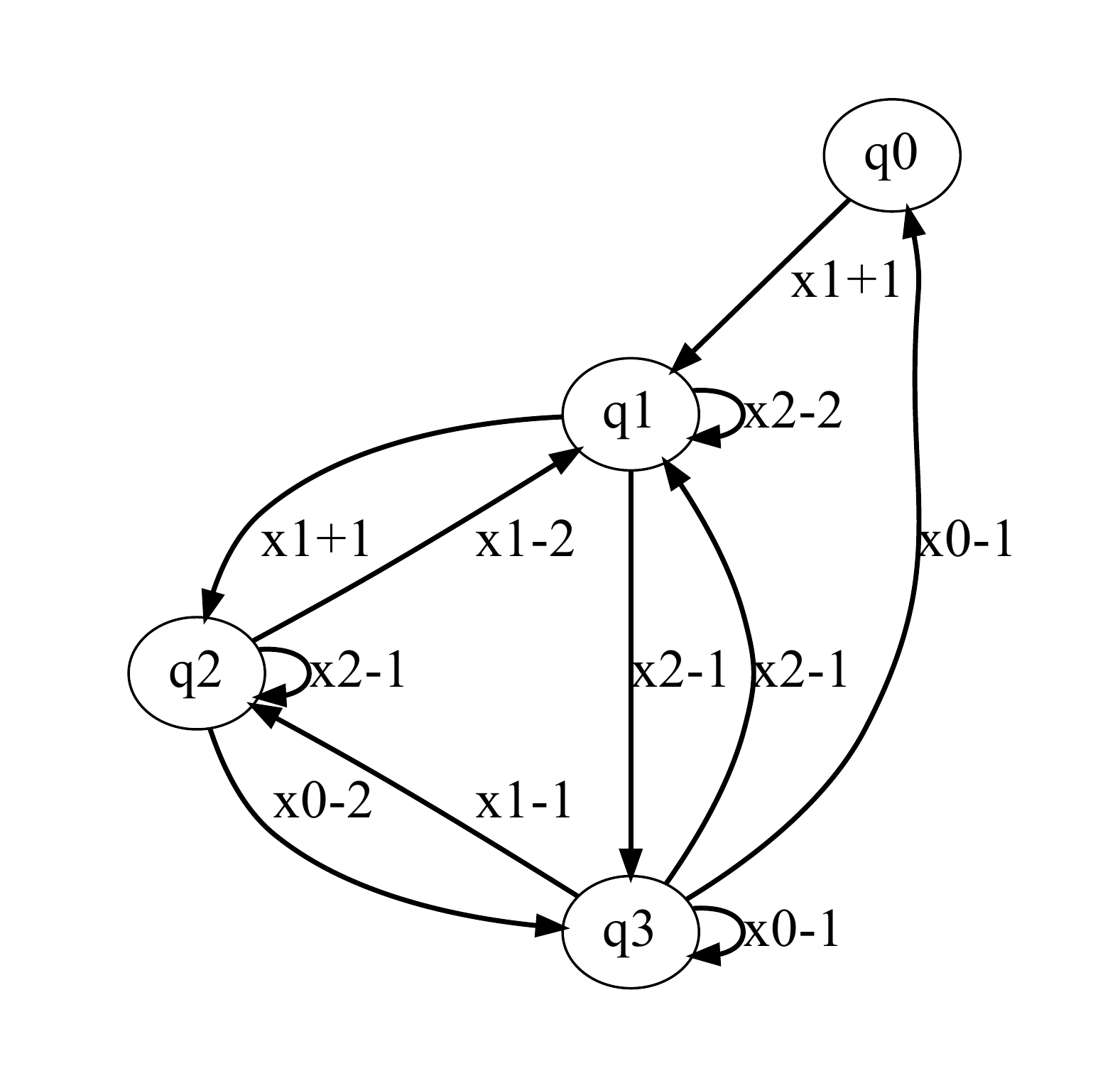}}
    \subcaptionbox{\small $D_{\Phi_2}$\label{fig:phi2_det1}}
        {\includegraphics[height=1.5in]{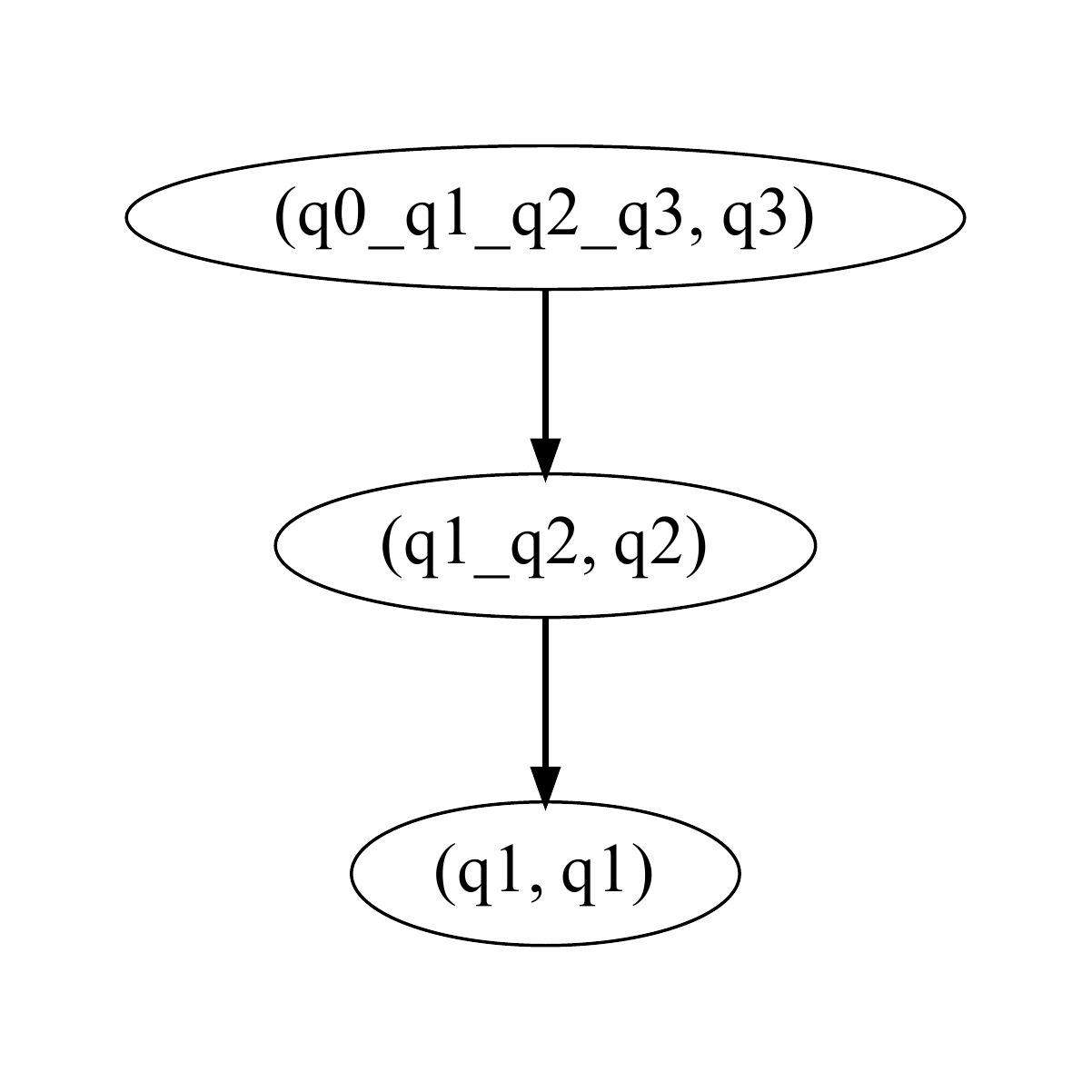}}
    \subcaptionbox{\small $\Phi_2^2 = \Phi_2^1 \setminus \set{x0-, x2-}$ \label{fig:phi12_fmp2}}
        {\includegraphics[height=1.5in]{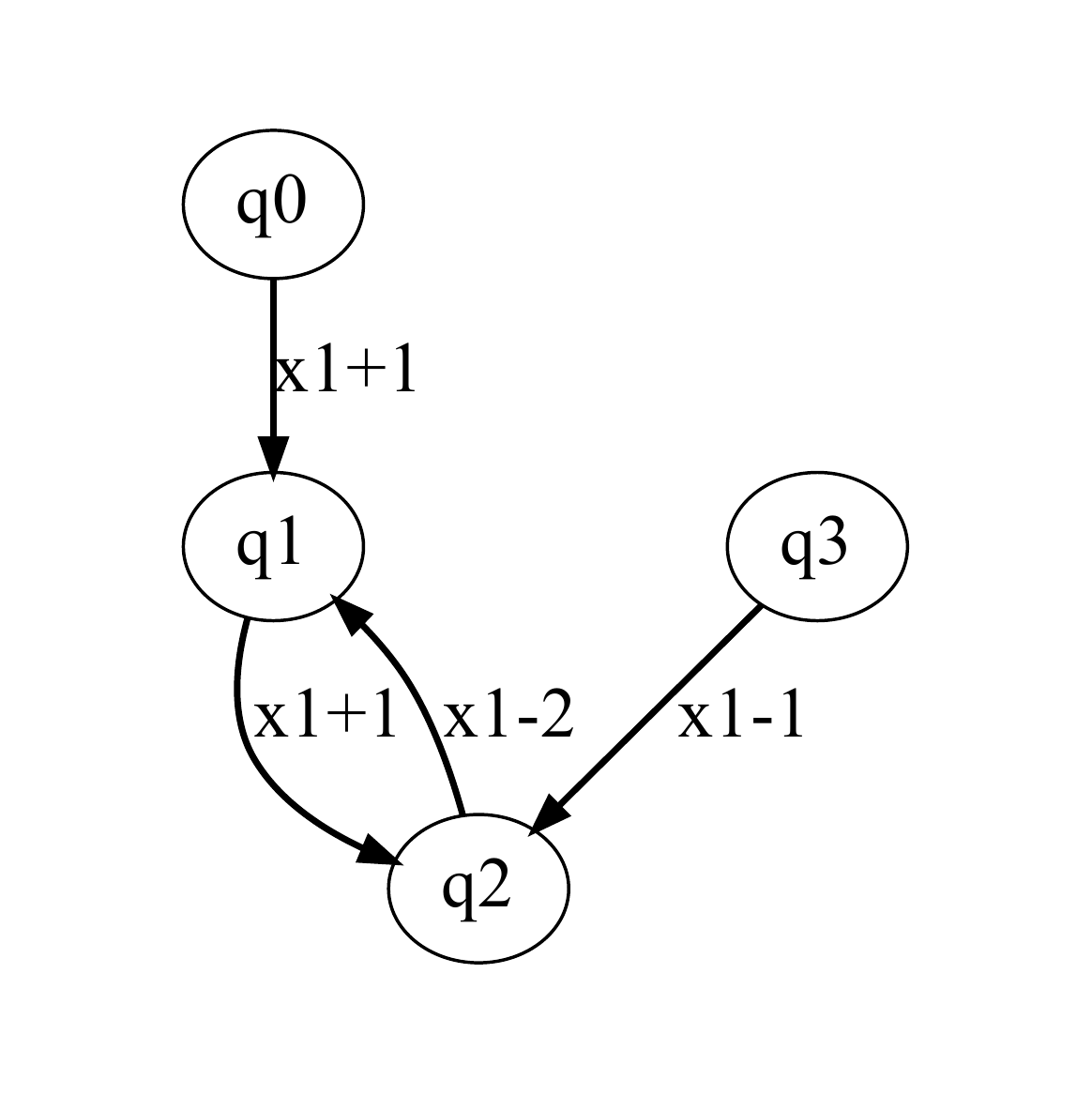}}
    \subcaptionbox{\small $D_{\Phi_2^2}$.\label{fig:phi2_det2}}
        {\includegraphics[height=0.5in]{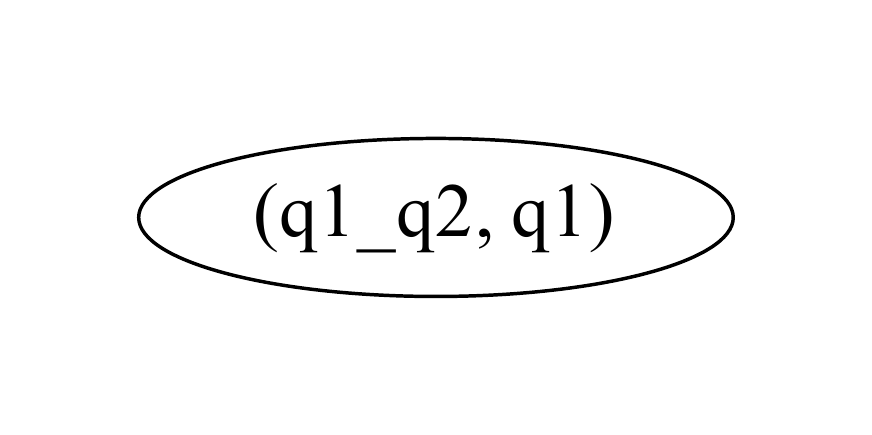}}

   \caption{\small Steps in \algo's assertion of termination on $\Phi_2$ (4 control states). Total runtime: 1.5s. }\label{fig:algo_phi2}
\end{figure}

\begin{figure}[h]
  \centering
  \subcaptionbox{\small  $\Phi_{3}$ \label{fig:phi_fmp3}}
  {\includegraphics[width=1.5in]{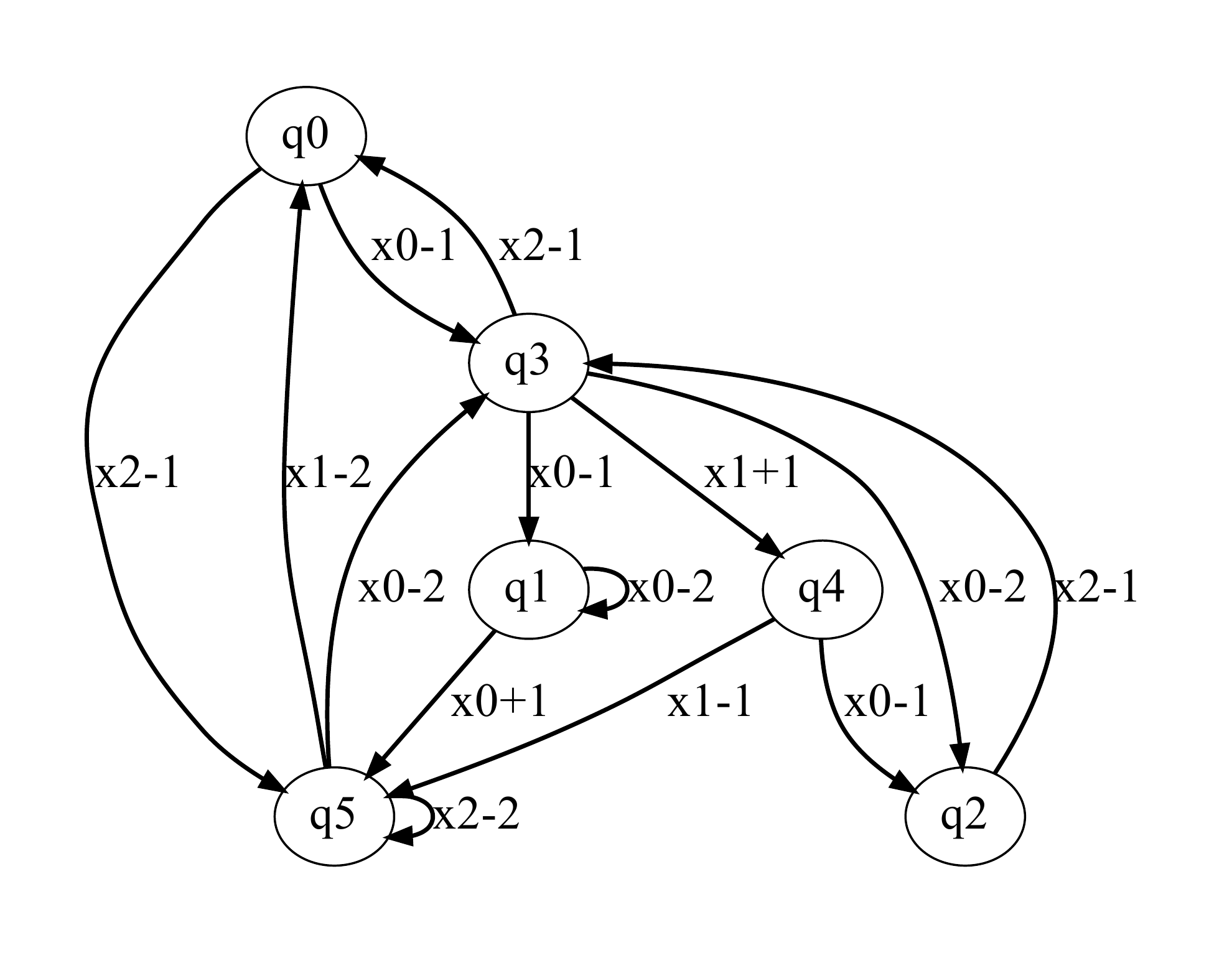}}
    \subcaptionbox{\small $D_{\Phi_3}$\label{fig:phi3_det1}}
   {\includegraphics[width=1.5in]{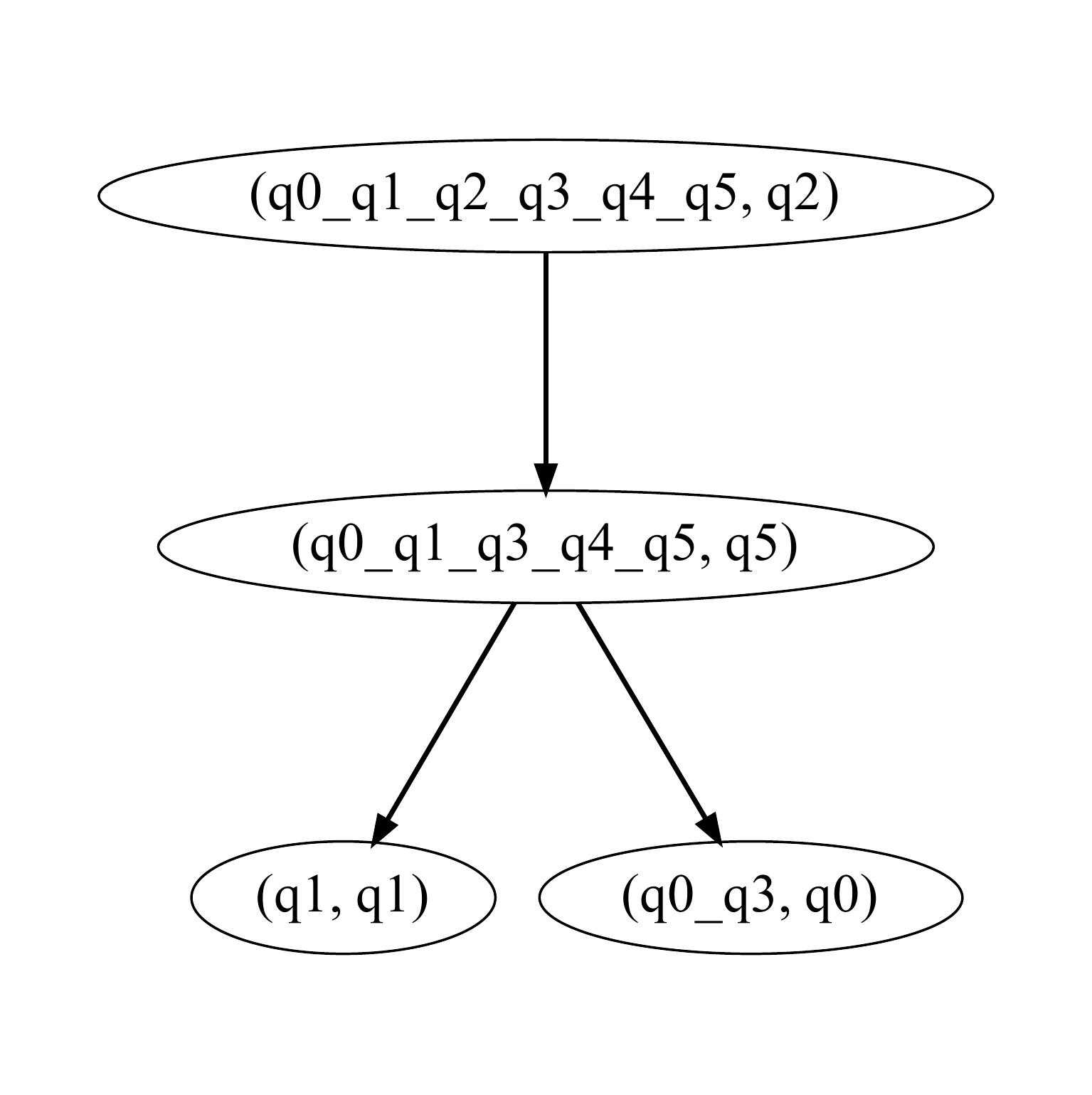}}
\subcaptionbox{\small ${\Phi_3^2}=\Phi_3 \setminus \set{x0-,x2-}$\label{fig:phi3_fmp2}}
   {\includegraphics[width=1.7in]{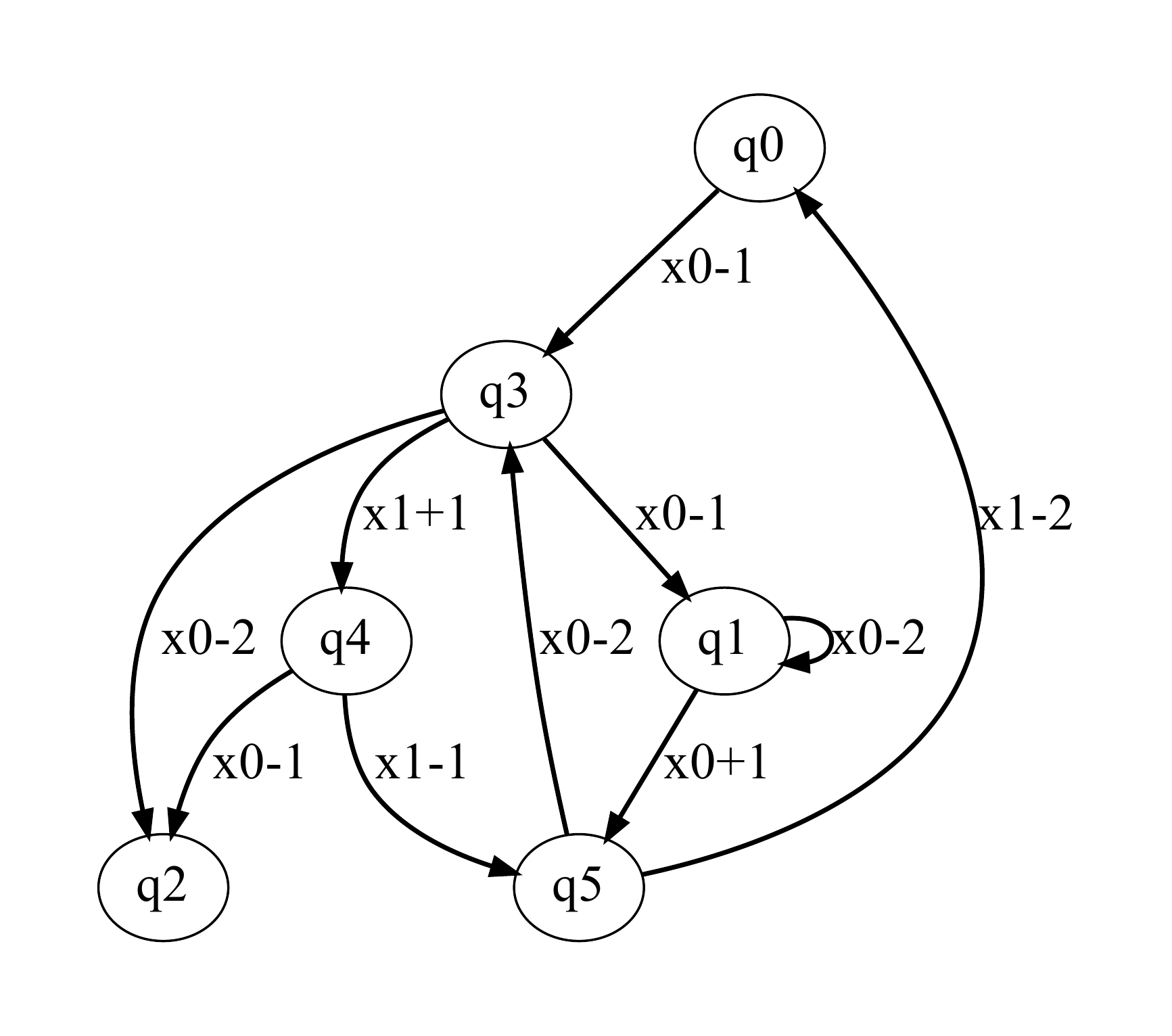}}
\subcaptionbox{\small  $D_{\Phi_3^2}$\label{fig:phi3_det2}}
   {\includegraphics[width=1in]{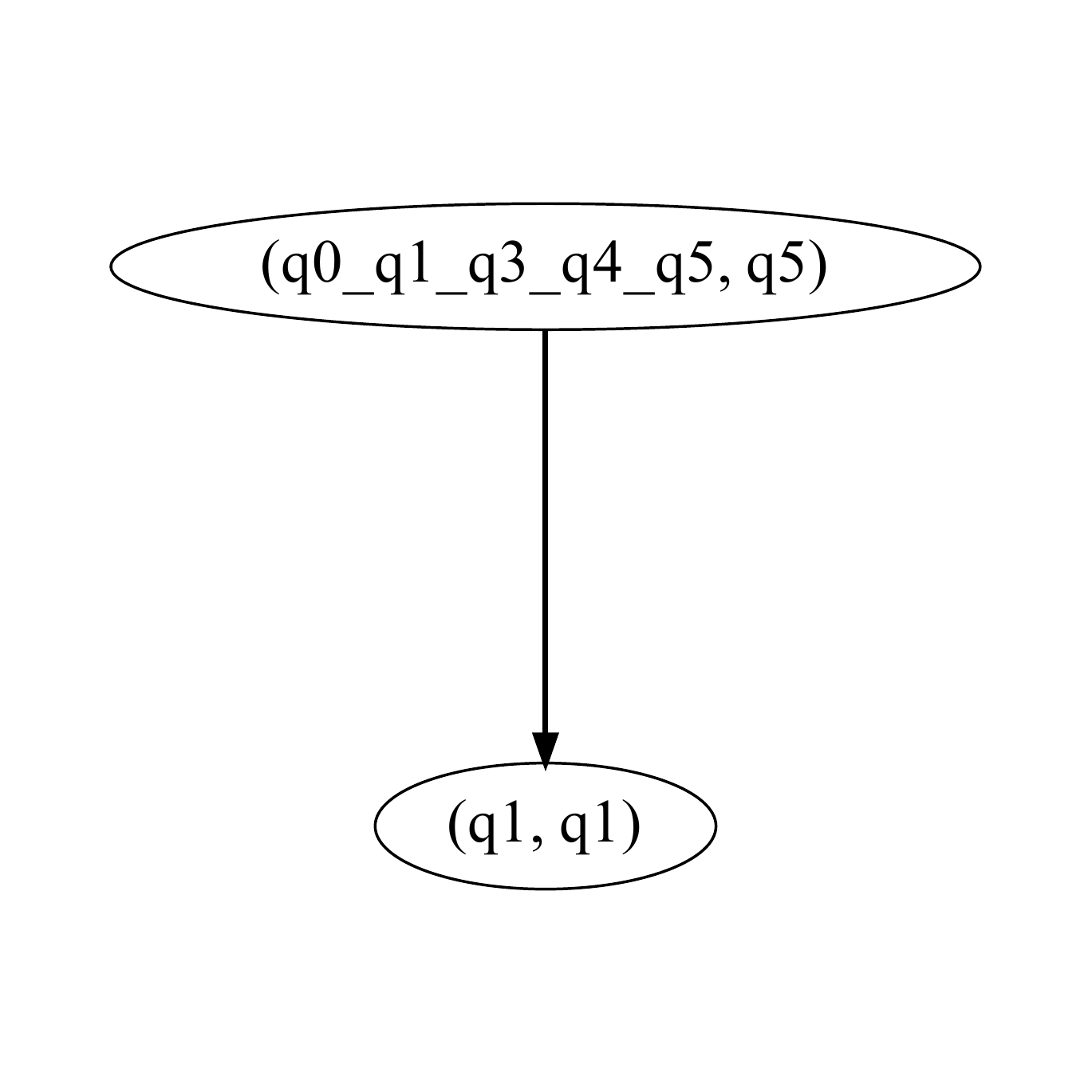}}

   \caption{\small Steps in \algo's assertion of termination on $\Phi_3$ (5 control states). Total runtime: 1.5s. }\label{fig:algo_phi3}
\end{figure}

\begin{figure}[h!]
  \centering
  \subcaptionbox{\small  $\Phi_{4}$ \label{fig:phi_fmp4}}
  {\includegraphics[width=2.5in]{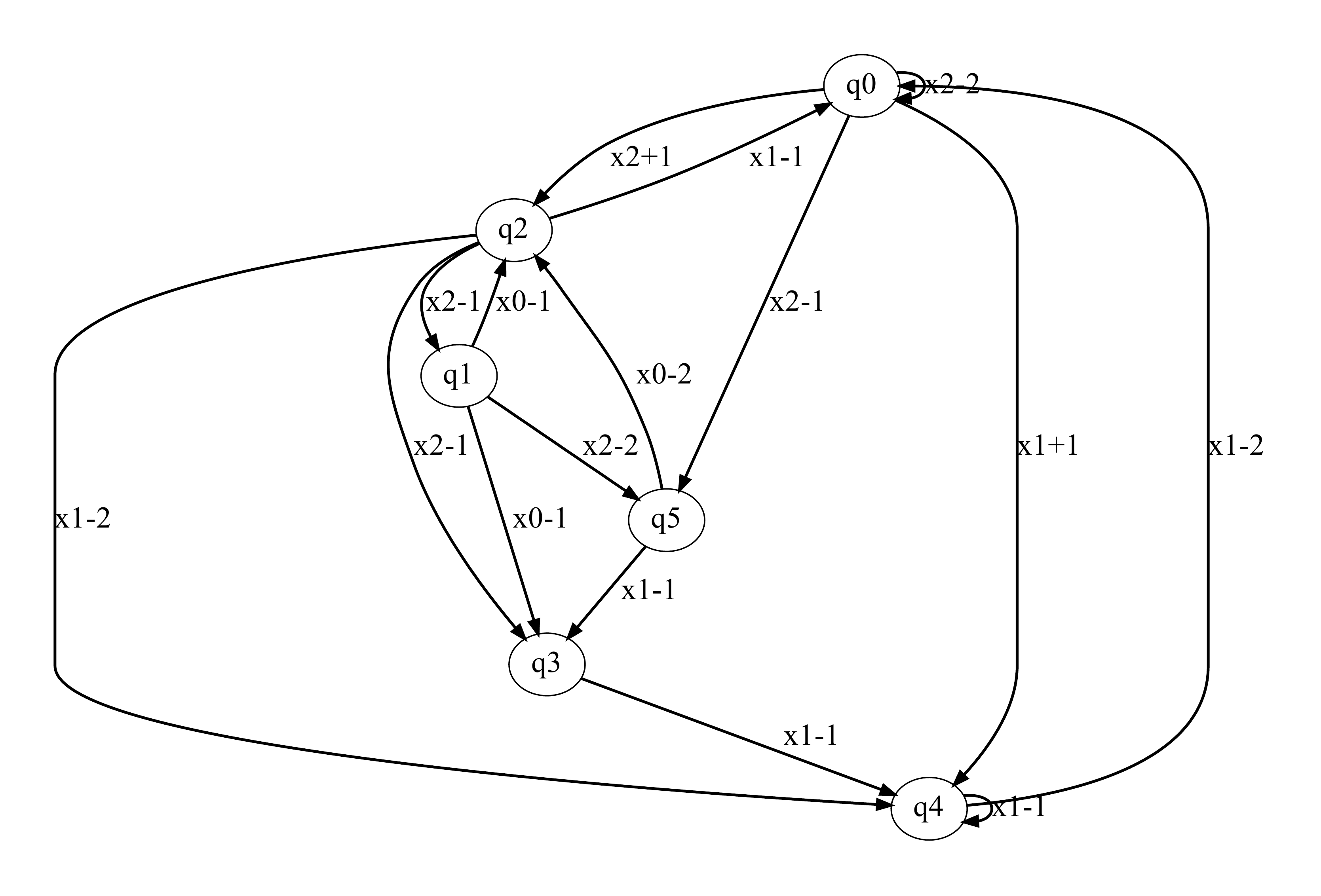}}
  \hspace{-1cm}
    \subcaptionbox{\small $D_{\Phi_4}$\label{fig:phi4_det1}}
   {\includegraphics[width=1.5in]{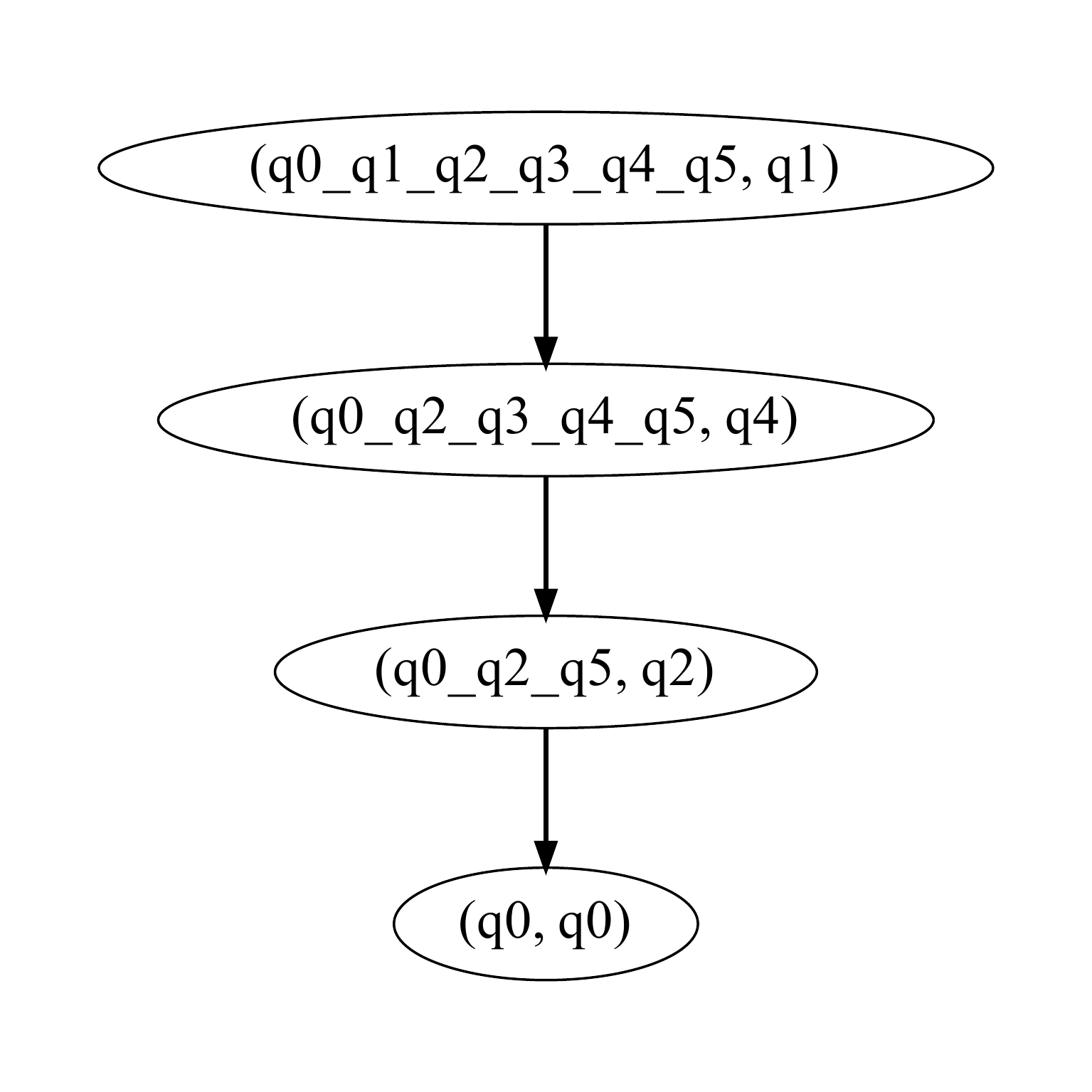}}
\subcaptionbox{\small ${\Phi_4^2}=\Phi_4 \setminus \set{x0-,x1-}$\label{fig:phi4_fmp2}}
   {\includegraphics[width=1.5in]{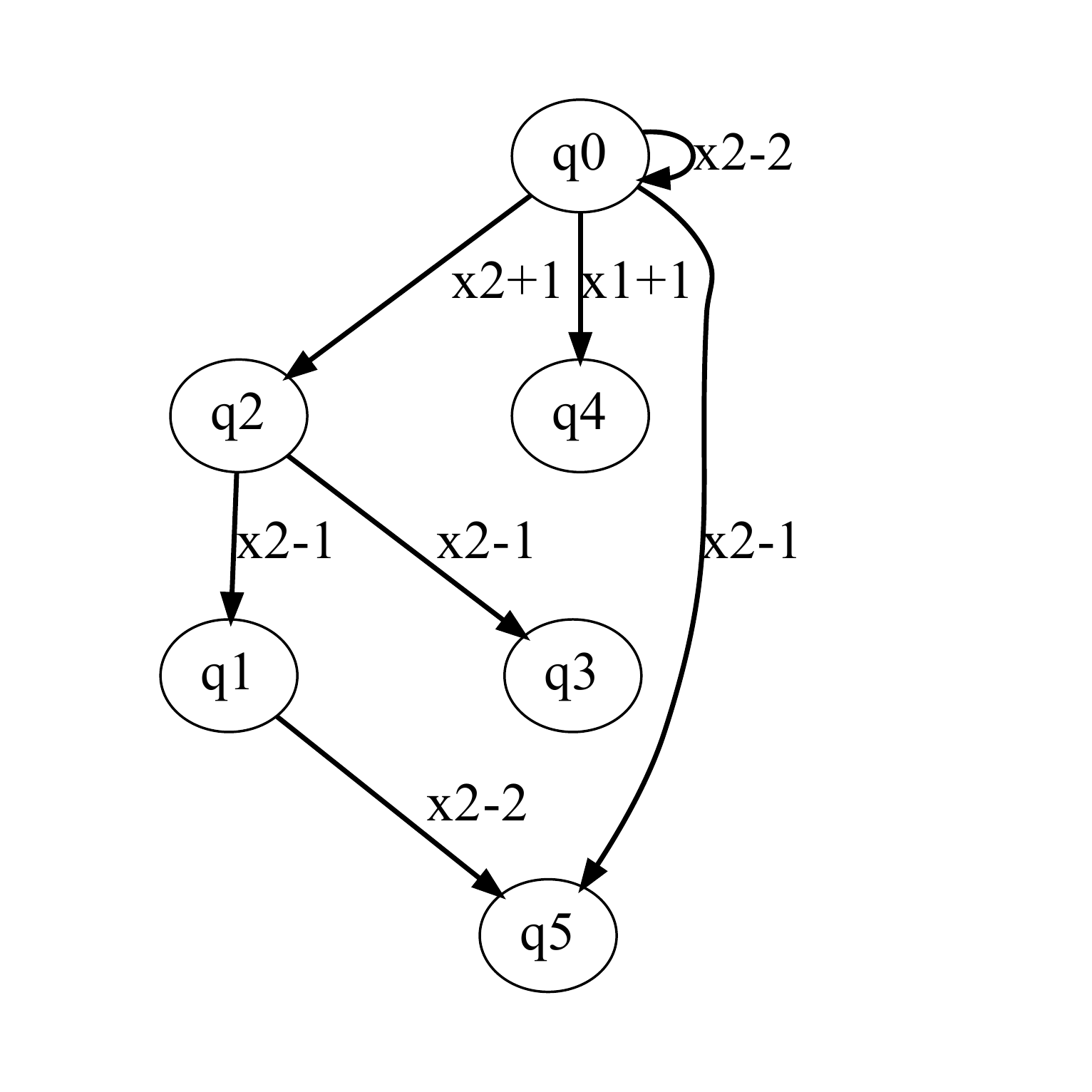}}
\subcaptionbox{\small  $D_{\Phi_4^2}$\label{fig:phi4_det2}}
   {\includegraphics[width=0.7in]{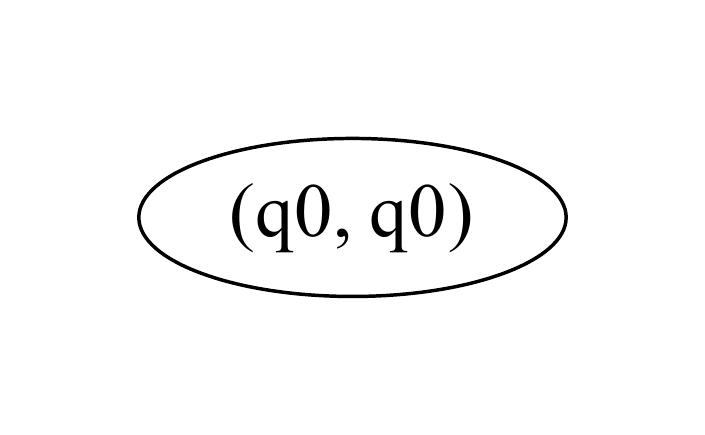}}
   \caption{\small Steps in \algo's assertion of termination on $\Phi_4$ (6 control states). Total runtime: 1.2s. }\label{fig:algo_phi4}
\end{figure}

\begin{figure}[b!]
  \centering
  \subcaptionbox{\small  $\Phi_{5}$ \label{fig:phi_fmp5}}
        {\includegraphics[width=3in]{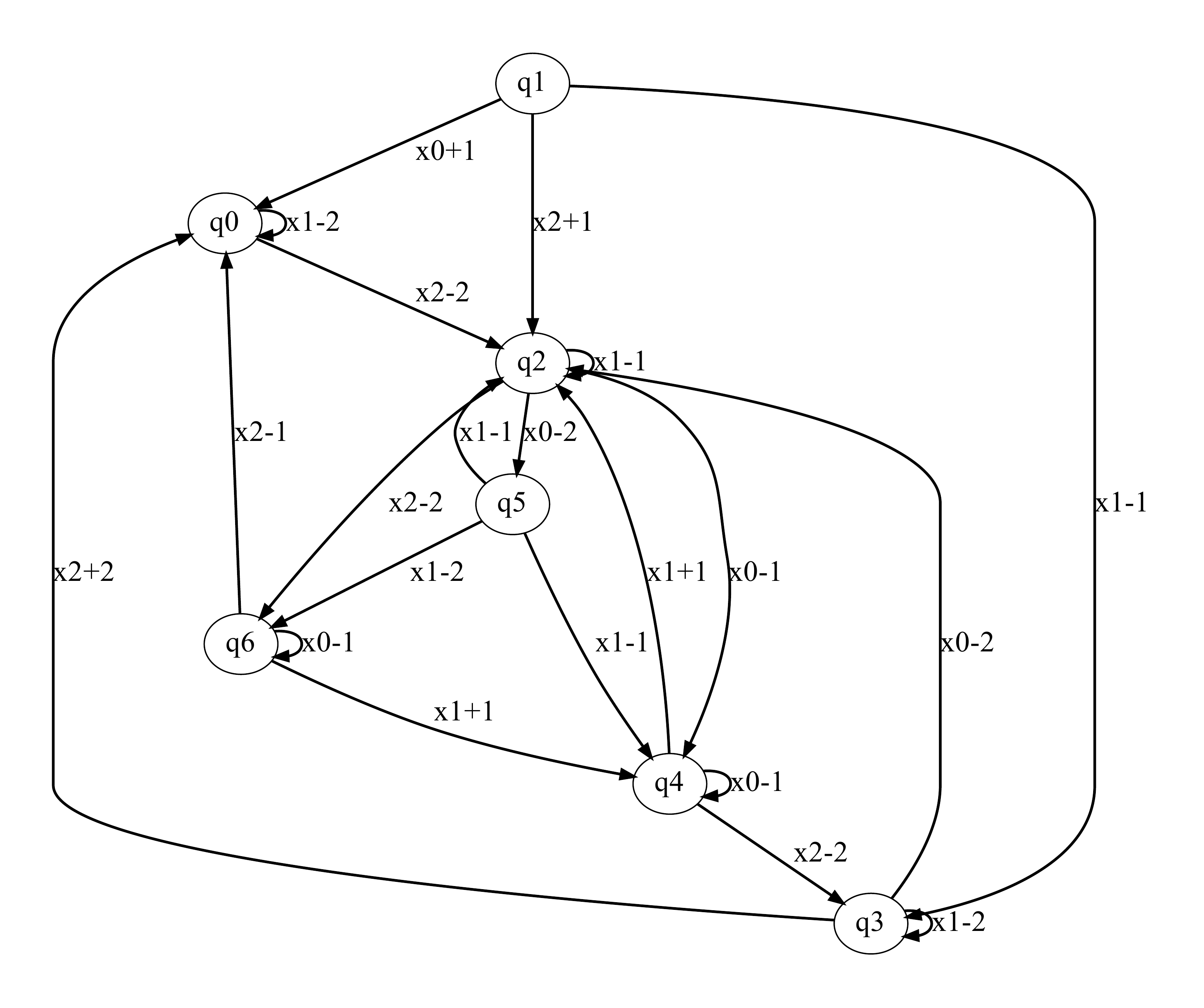}}
    \subcaptionbox{\small $D_{\Phi_5}$\label{fig:phi5_det1}}
        {\includegraphics[width=1.5in]{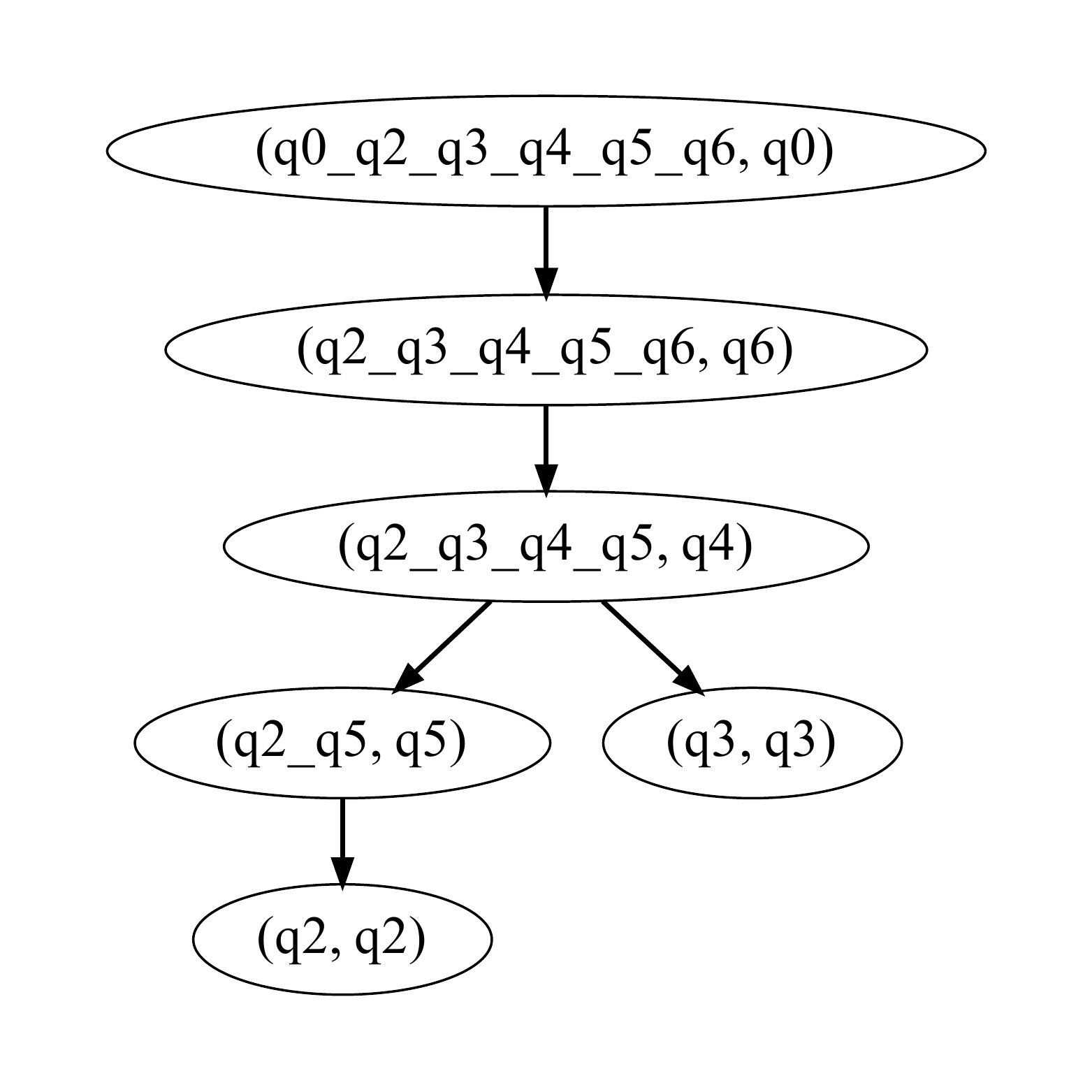}}
    \subcaptionbox{\small ${\Phi_5^2}=\Phi_5 \setminus \set{x0-}$\label{fig:phi5_fmp2}}
        {\includegraphics[width=2.5in]{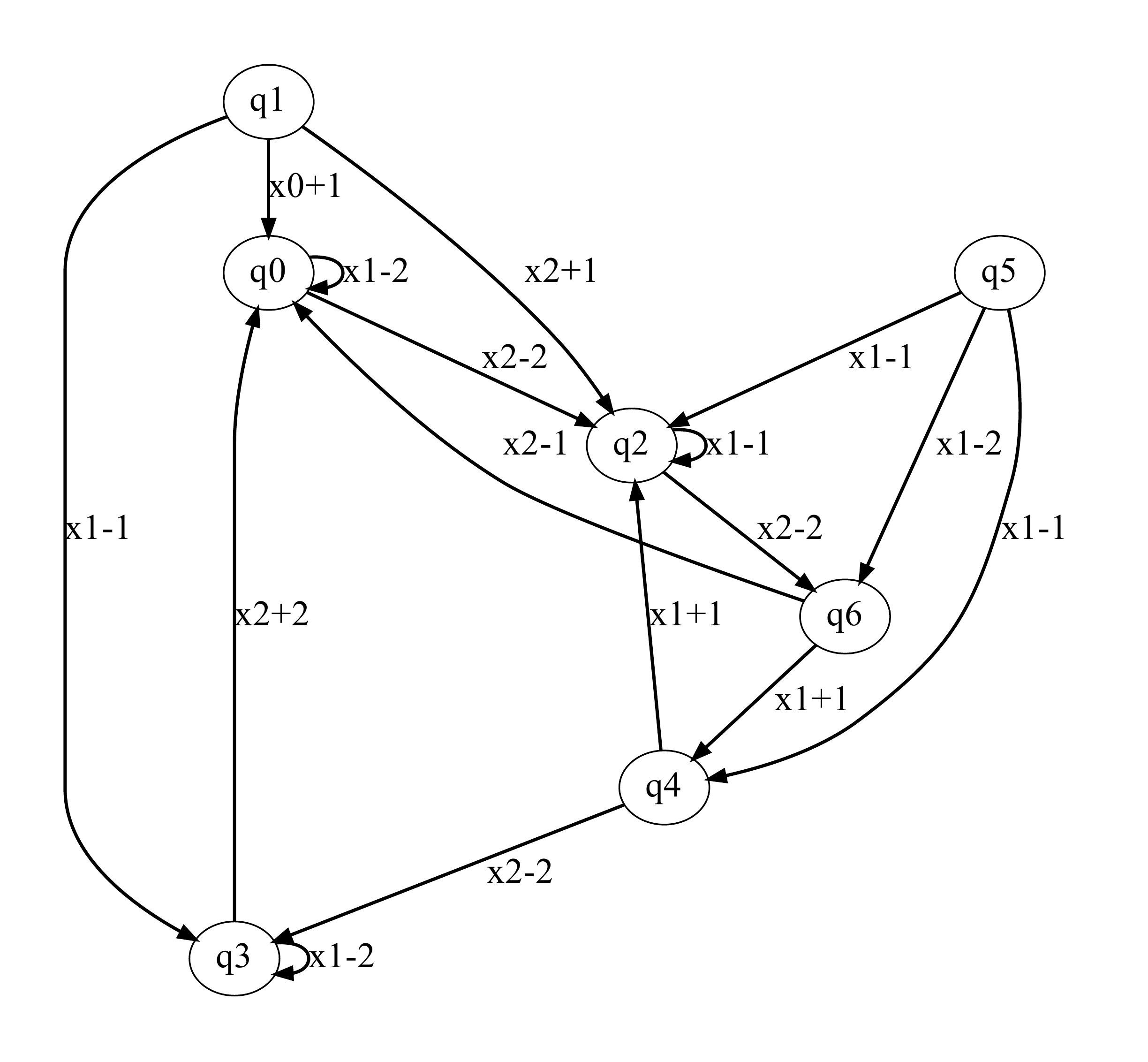}}
    \subcaptionbox{\small  $D_{\Phi_5^2}$\label{fig:phi5_det2}}           {\includegraphics[width=2in]{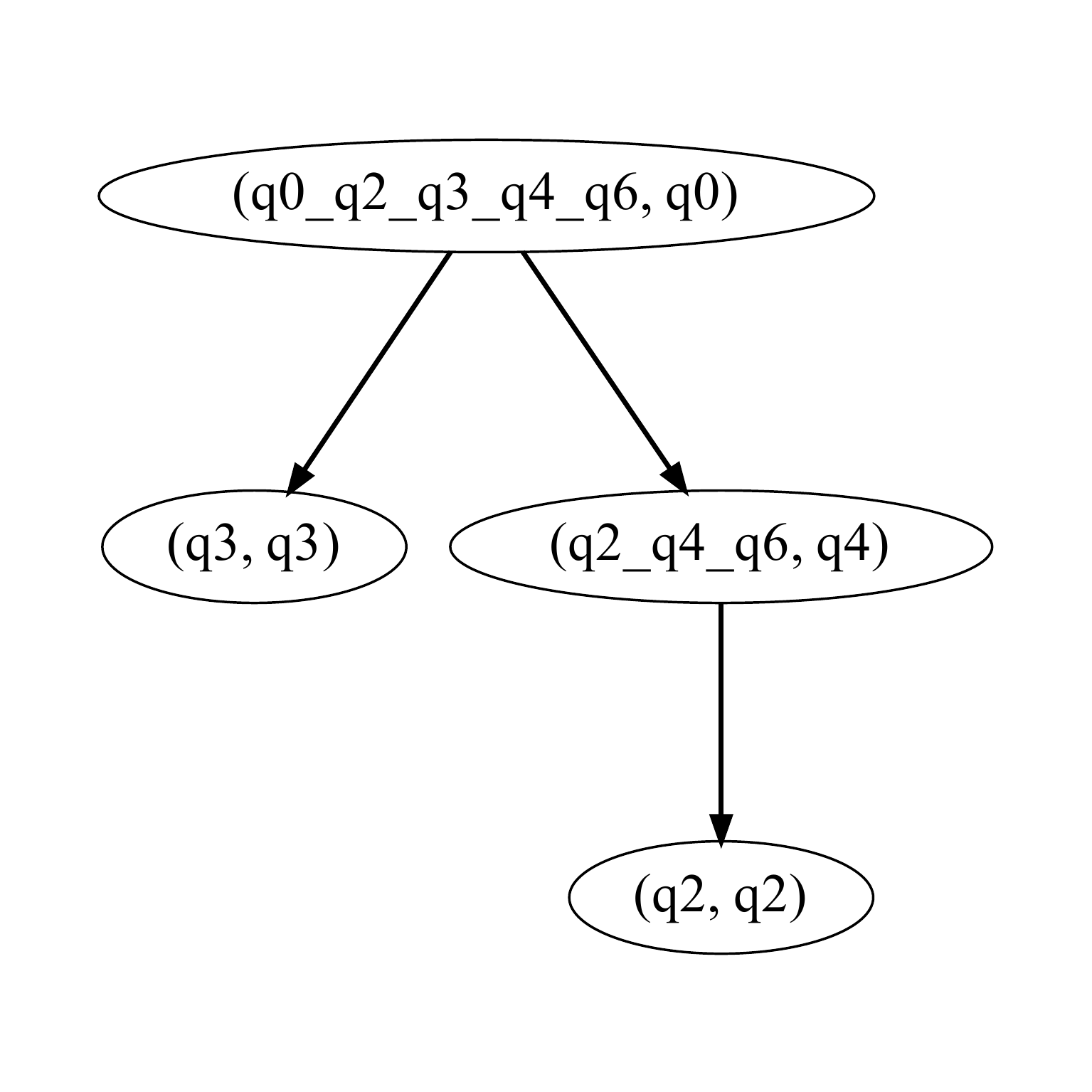}}    \subcaptionbox{\small  $\Phi_5^3 = \Phi_5^2\setminus\set{x2-}$\label{fig:phi5_fmp3}}
        {\includegraphics[width=2.5in]{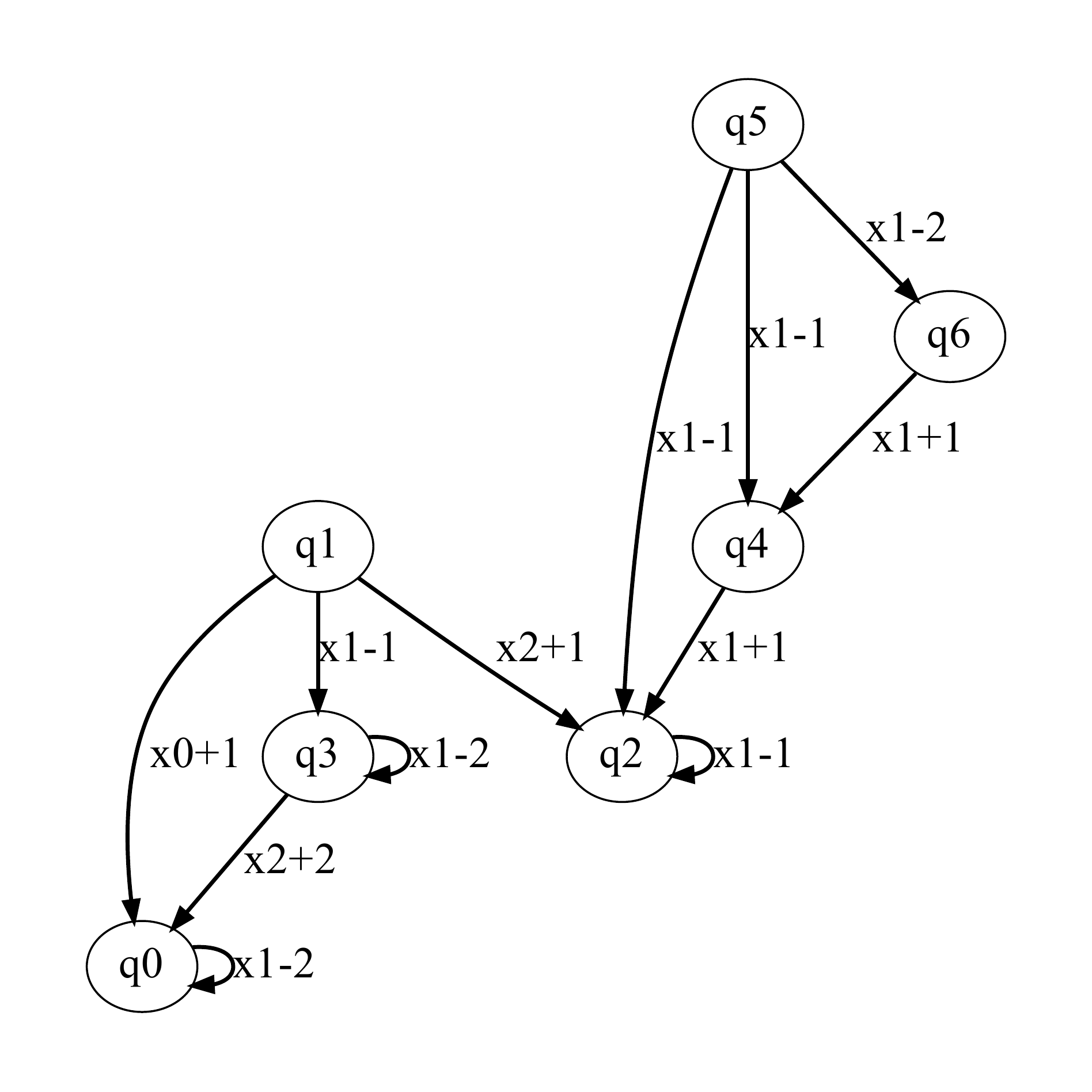}}
    \subcaptionbox{\small  $D_{\Phi_5^3}$\label{fig:phi5_det3}}
        {\includegraphics[width=2in]{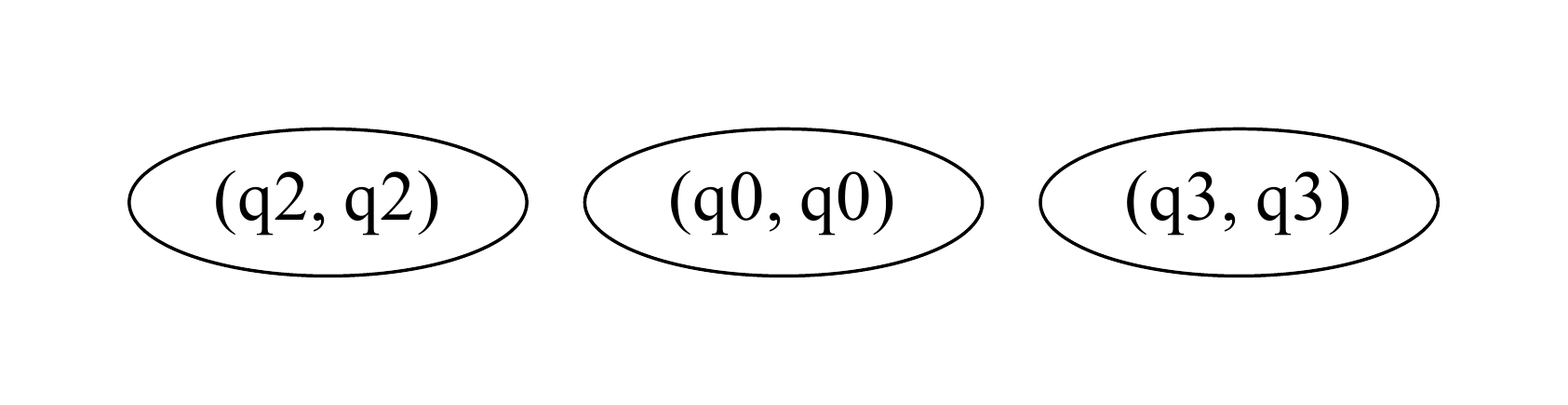}}

   \caption{\small Steps in \algo's assertion of termination on $\Phi_5$ (7 control states). Total runtime: 1.2s. }\label{fig:algo_phi5}
\end{figure}

\begin{figure}[h!]
  \centering
  \subcaptionbox{\small  $\Phi_{6}$ \label{fig:phi6_fmp1}}
        {\includegraphics[width=3.2in]{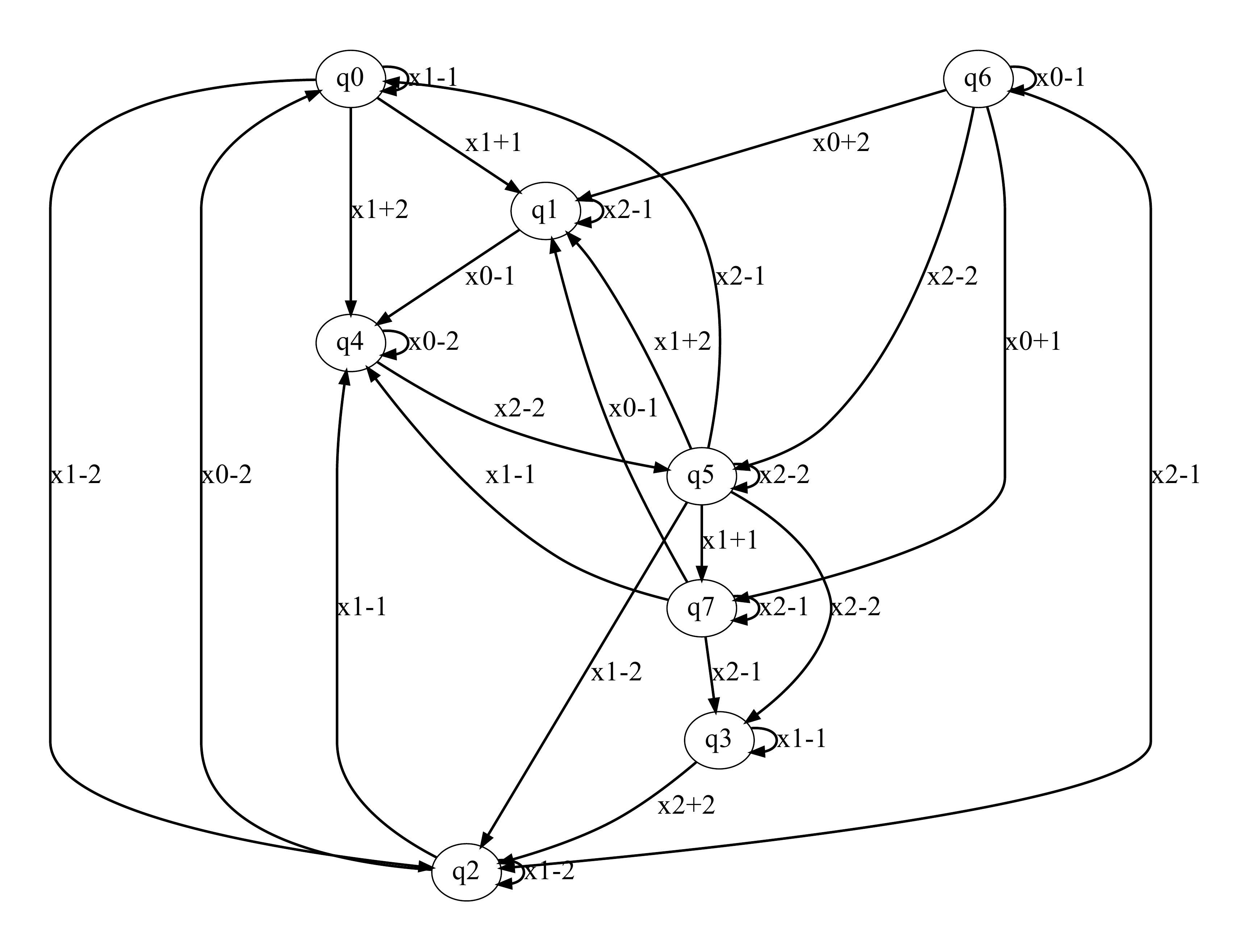}}
    \subcaptionbox{\small $D_{\Phi_6}$\label{fig:phi6_det1}}
        {\includegraphics[width=2.5in]{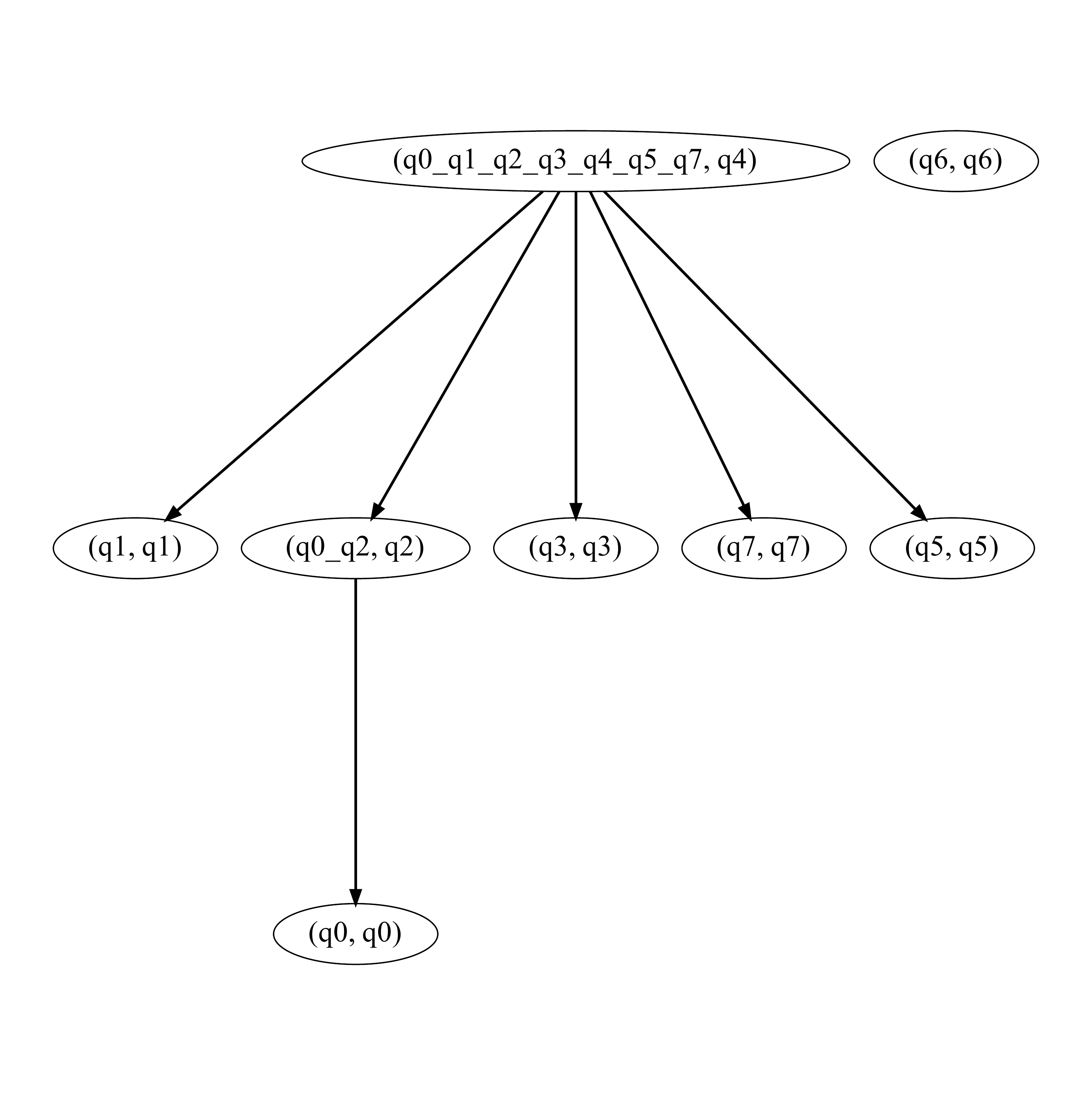}}
    \subcaptionbox{\small ${\Phi_6^2}=\Phi_5 \setminus \set{x0-}$\label{fig:phi6_fmp2}}
        {\includegraphics[width=2.5in]{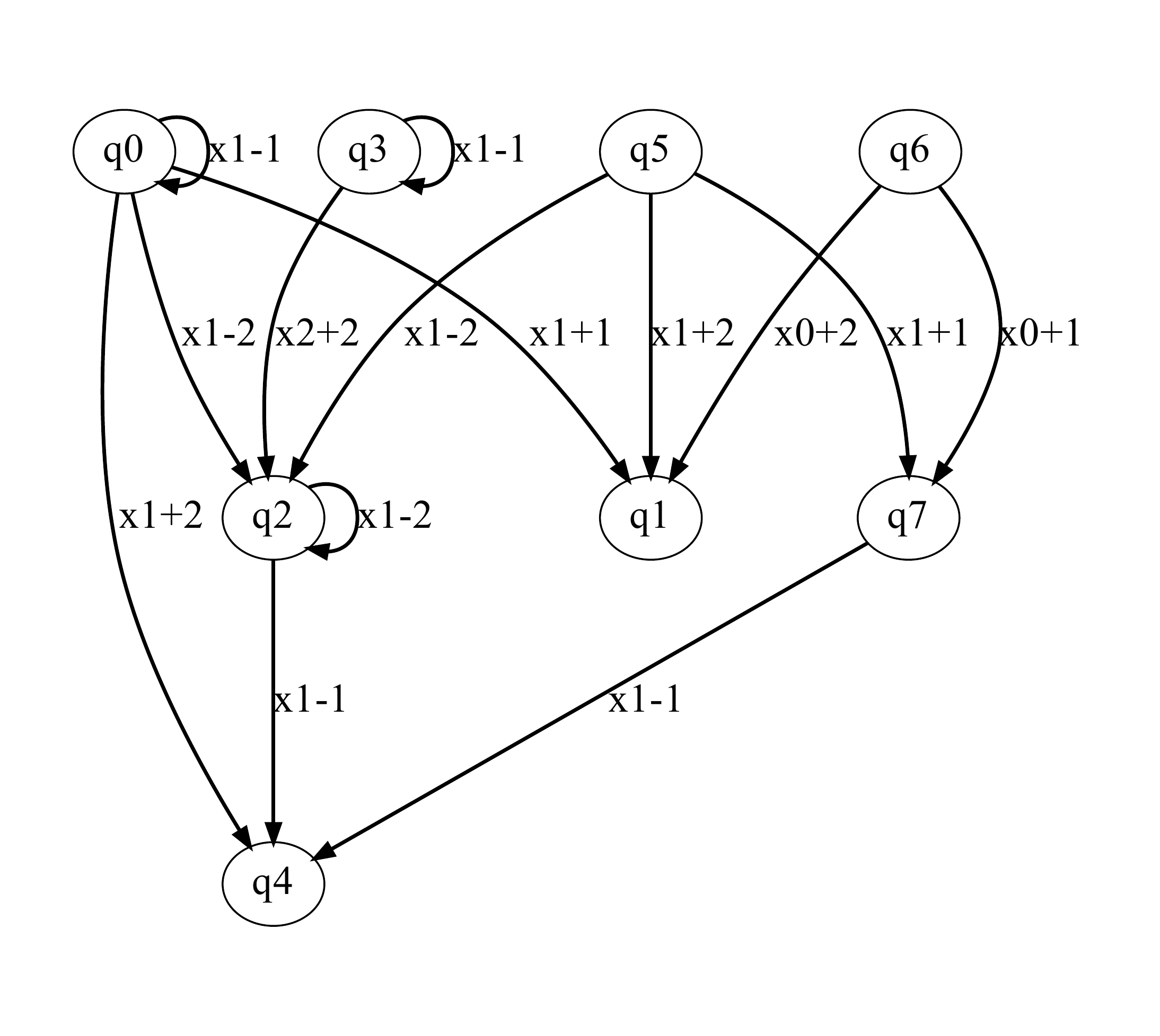}}
    \subcaptionbox{\small  $D_{\Phi_6^2}$\label{fig:phi6_det2}}           {\includegraphics[width=2in]{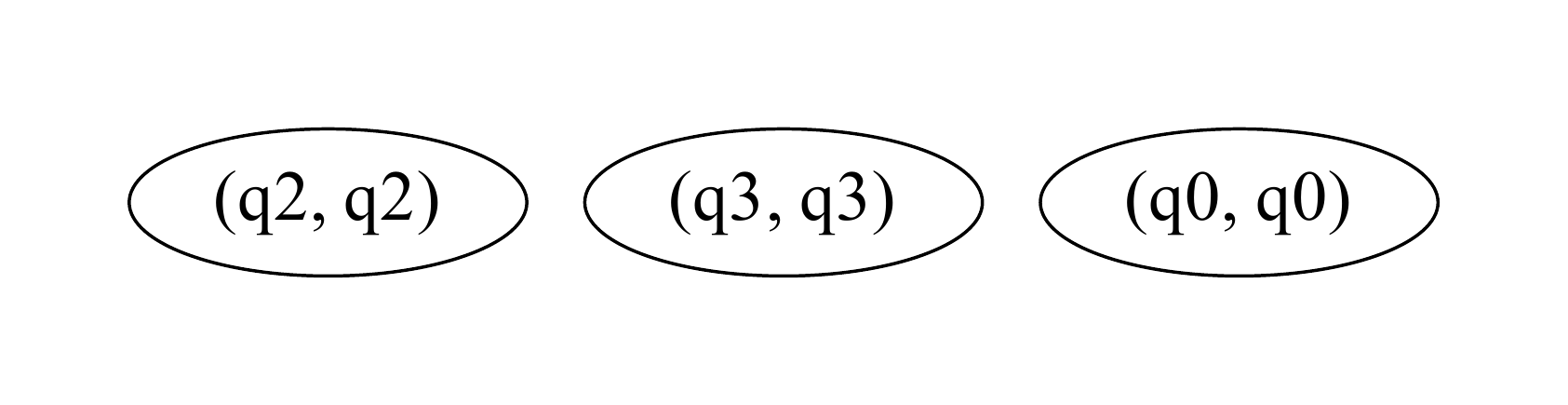}}    
    \caption{\small Steps in \algo's assertion of termination on $\Phi_6$ (8 control states). Total runtime: 1.2s. }\label{fig:algo_phi6}
\end{figure}

\begin{figure}[h!]
  \centering
  \subcaptionbox{\small  $\Phi_{7}$ \label{fig:phi7_fmp1}}
        {\includegraphics[width=4in]{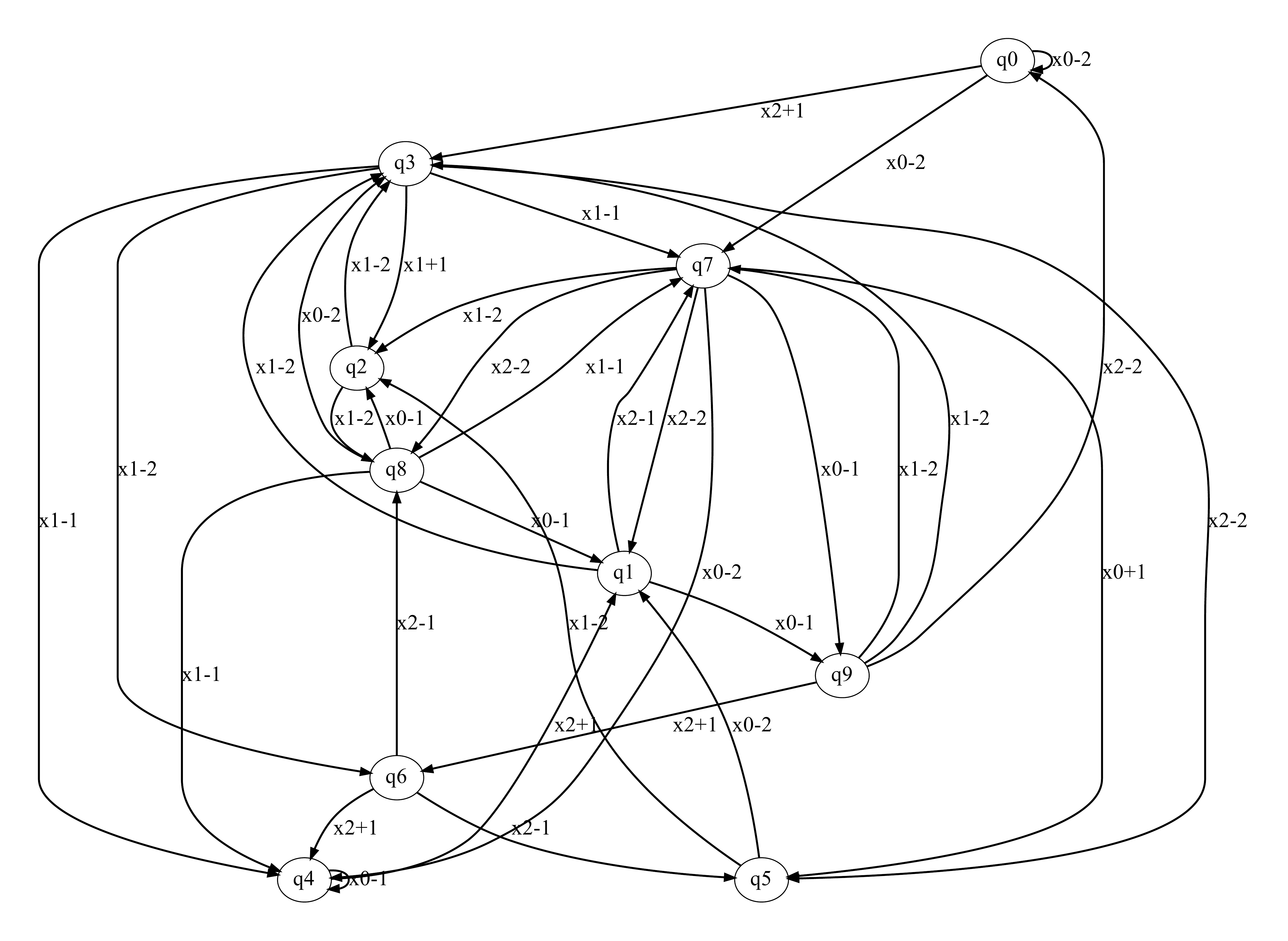}}
    \subcaptionbox{\small $D_{\Phi_7}$\label{fig:phi7_det1}}
        {\includegraphics[width=1.7in]{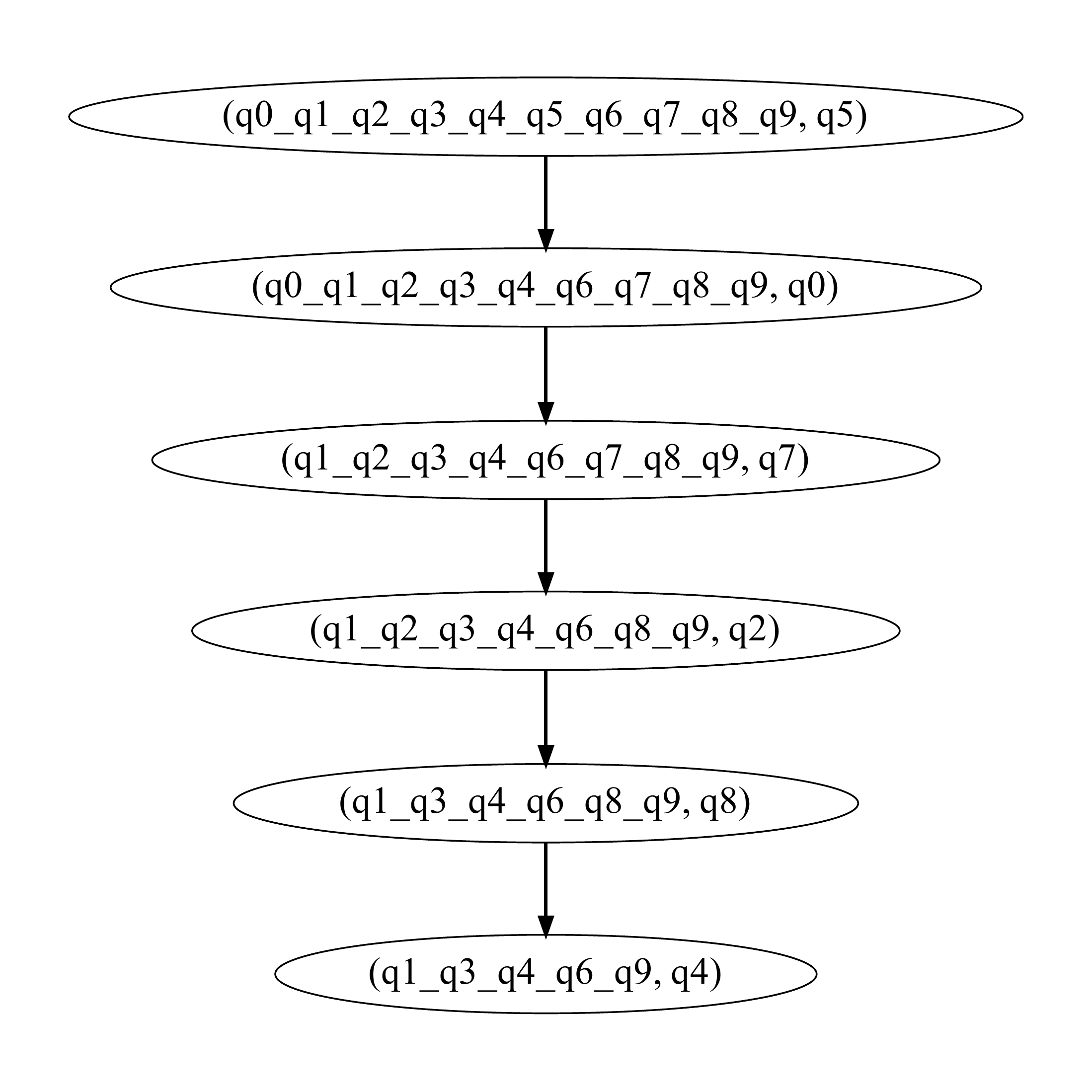}}
    \subcaptionbox{\small ${\Phi_7^2}=\Phi_7 \setminus \set{x1-}$\label{fig:phi7_fmp2}}
        {\includegraphics[width=2.5in]{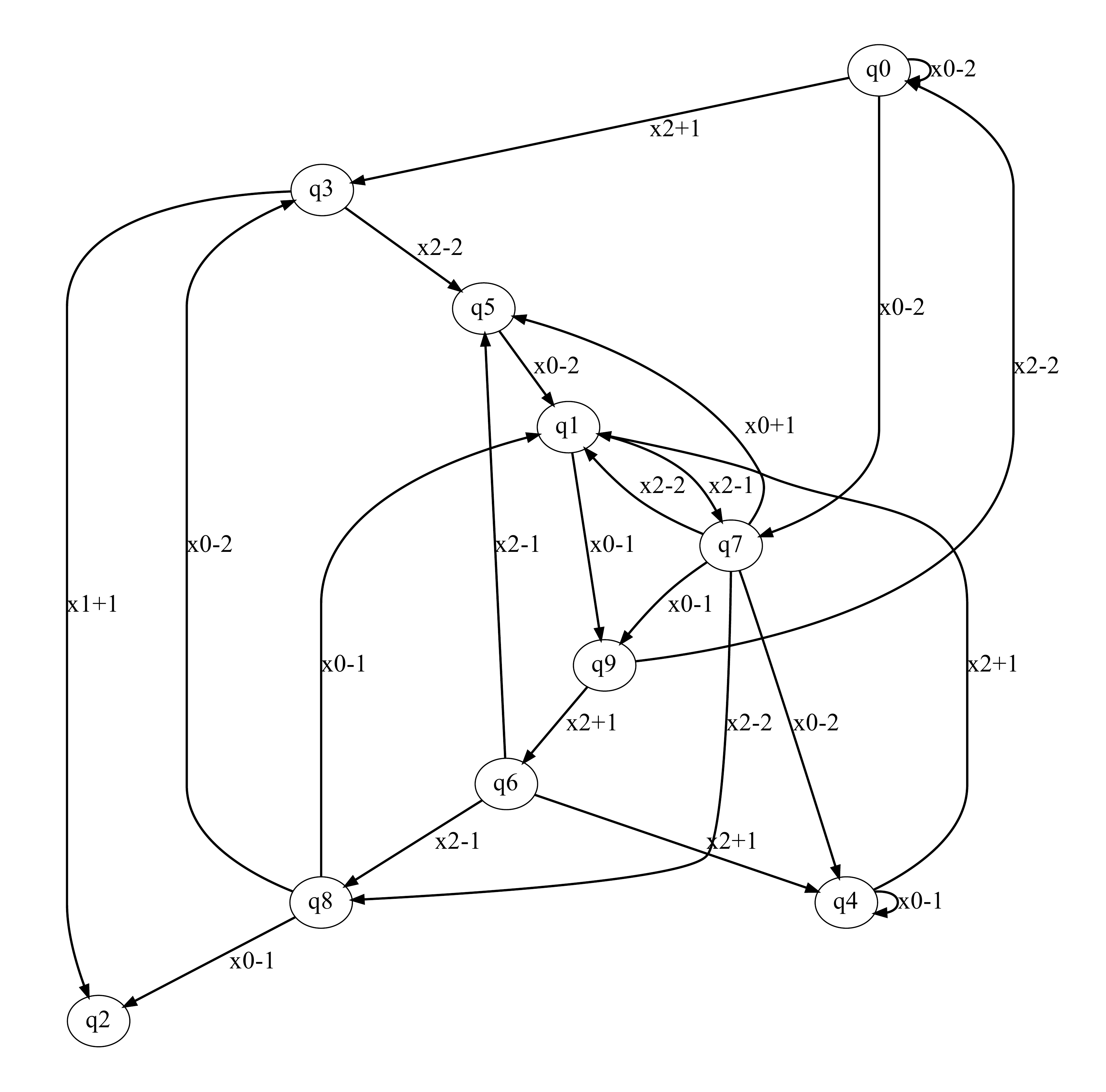}}
    \subcaptionbox{\small  $D_{\Phi_7^2}$\label{fig:phi7_det2}}           {\includegraphics[width=2in]{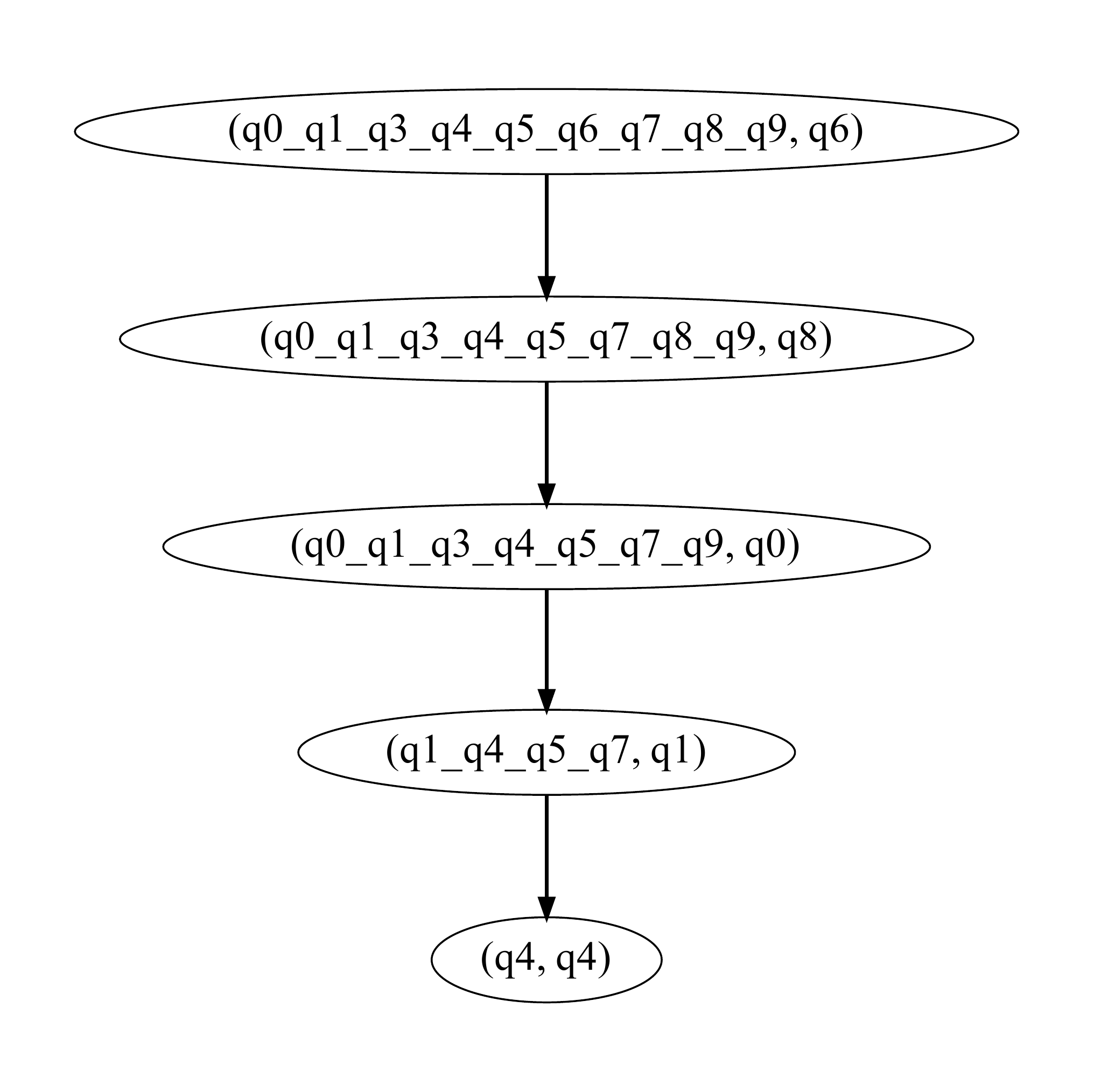}}    \subcaptionbox{\small  $\Phi_7^3 = \Phi_7^2\setminus\set{x0-}$\label{fig:phi7_fmp3}}
        {\includegraphics[width=2in]{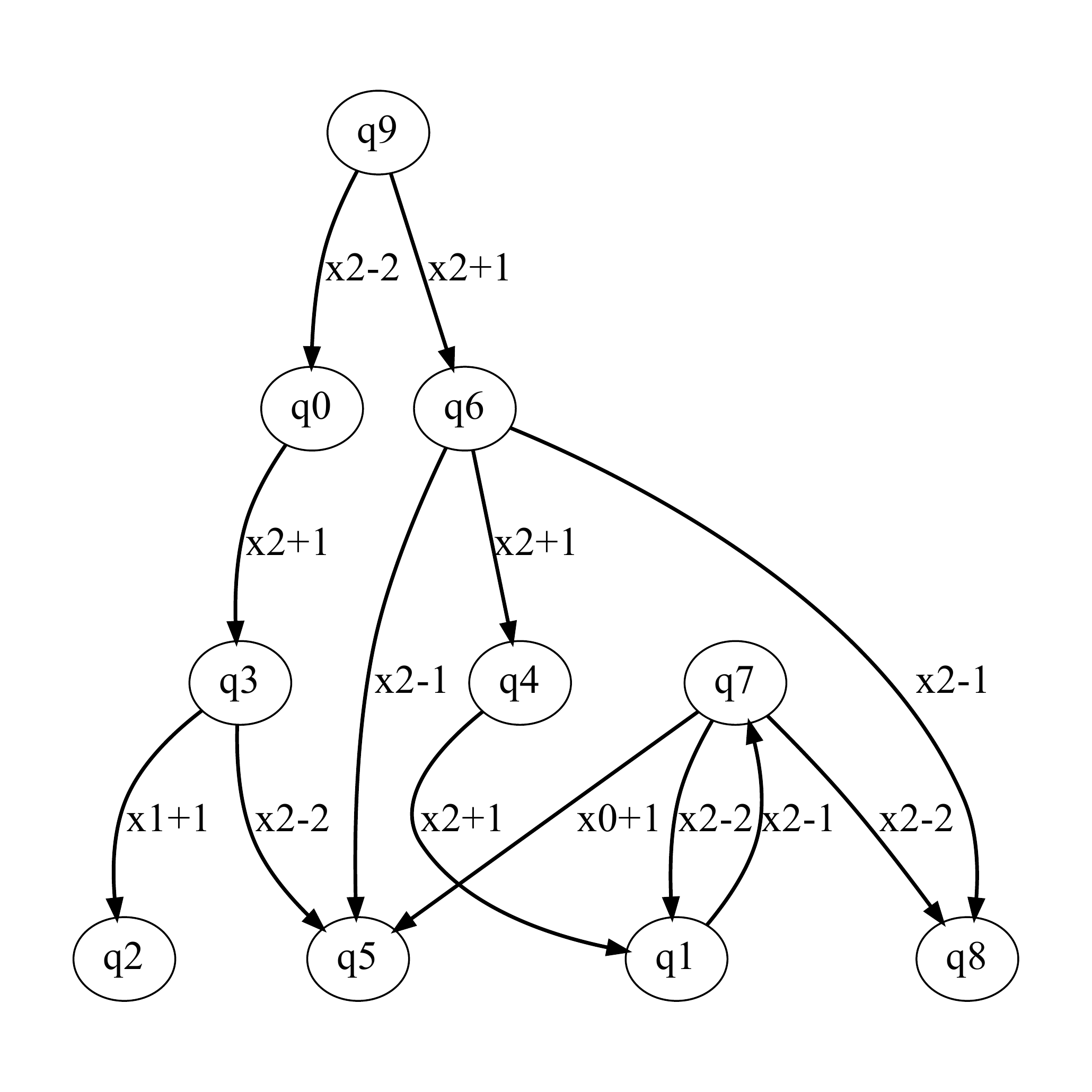}}
    \subcaptionbox{\small  $D_{\Phi_7^3}$\label{fig:phi7_det3}}
        {\includegraphics[width=1in]{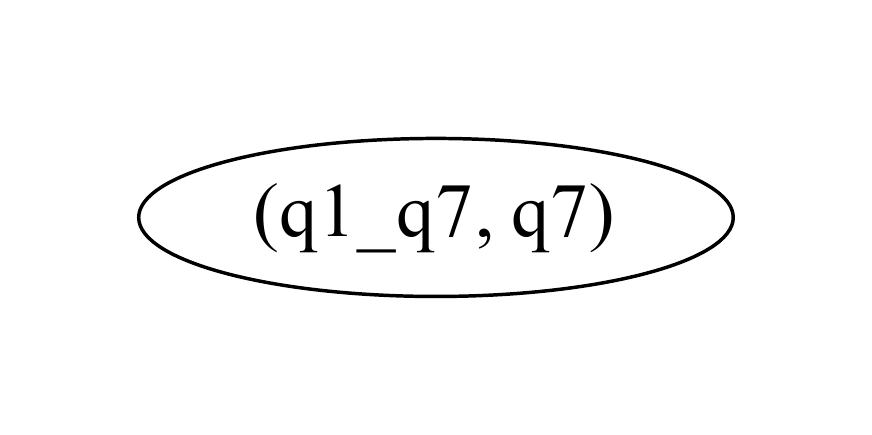}}

   \caption{\small Steps in \algo's assertion of termination on $\Phi_7$ (10 control states). Total runtime: 2s. }\label{fig:algo_phi7}
\end{figure}

\Cref{fig:algo_phi2,fig:algo_phi3,fig:algo_phi4,fig:algo_phi5,fig:algo_phi6,fig:algo_phi7}
show randomly generated policies and the DEFs generated for them initially as
well after every round of edge removal. All of these policies  were found to be
terminating by \algo. The Sieve algorithm  could not assert termination for any
of these policies.  Some of the policies required multiple rounds of edge
removals. An additional policy that required four iterations of the main while
loop in \algo  is 
presented in Appendix~\ref{app:more}.

\clearpage

\subsubsection{Dependence on Directed Elimination Forests}
\begin{figure}[h]
  \centering
  \subcaptionbox{\small  $\Phi_{9}$ \label{fig:phi_fmp9}}
        {\includegraphics[width=2.5in]{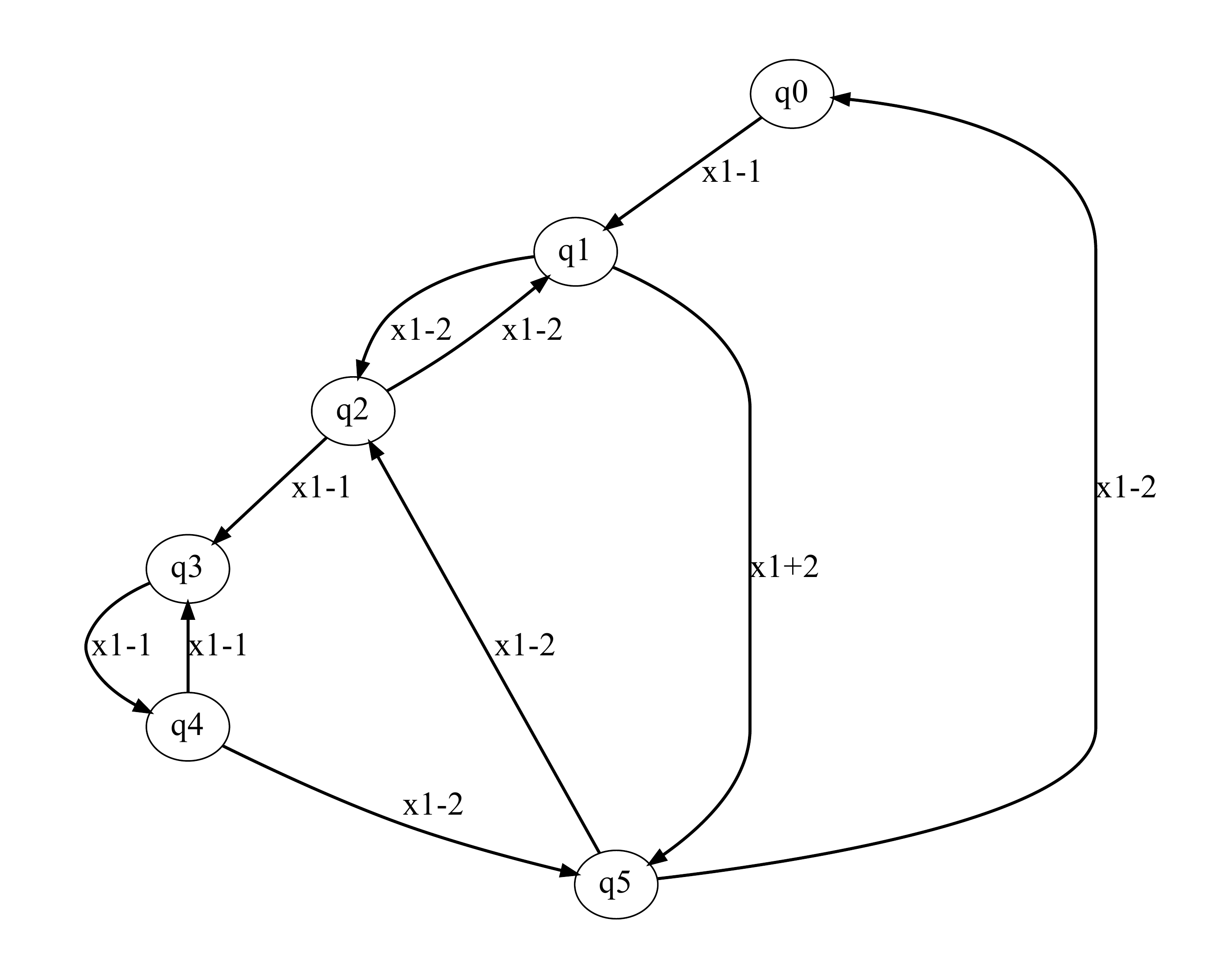}}
    \subcaptionbox{\small $D_{\Phi_9}$\label{fig:phi9_det1}}
        {\includegraphics[width=1.5in]{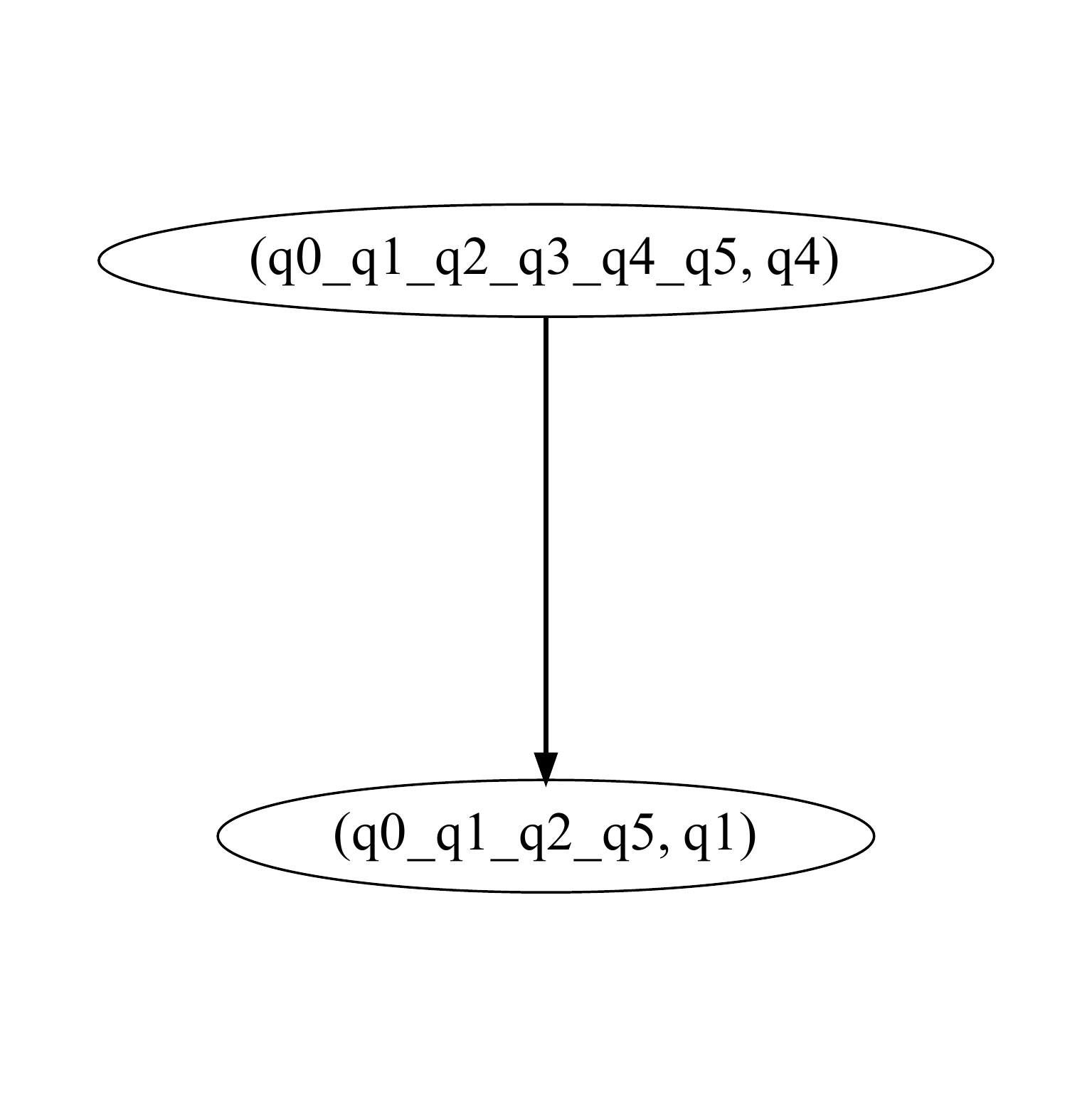}}
    \subcaptionbox{\small $D_{\Phi_9}'$\label{fig:phi9_det2}}
        {\includegraphics[width=1.5in]{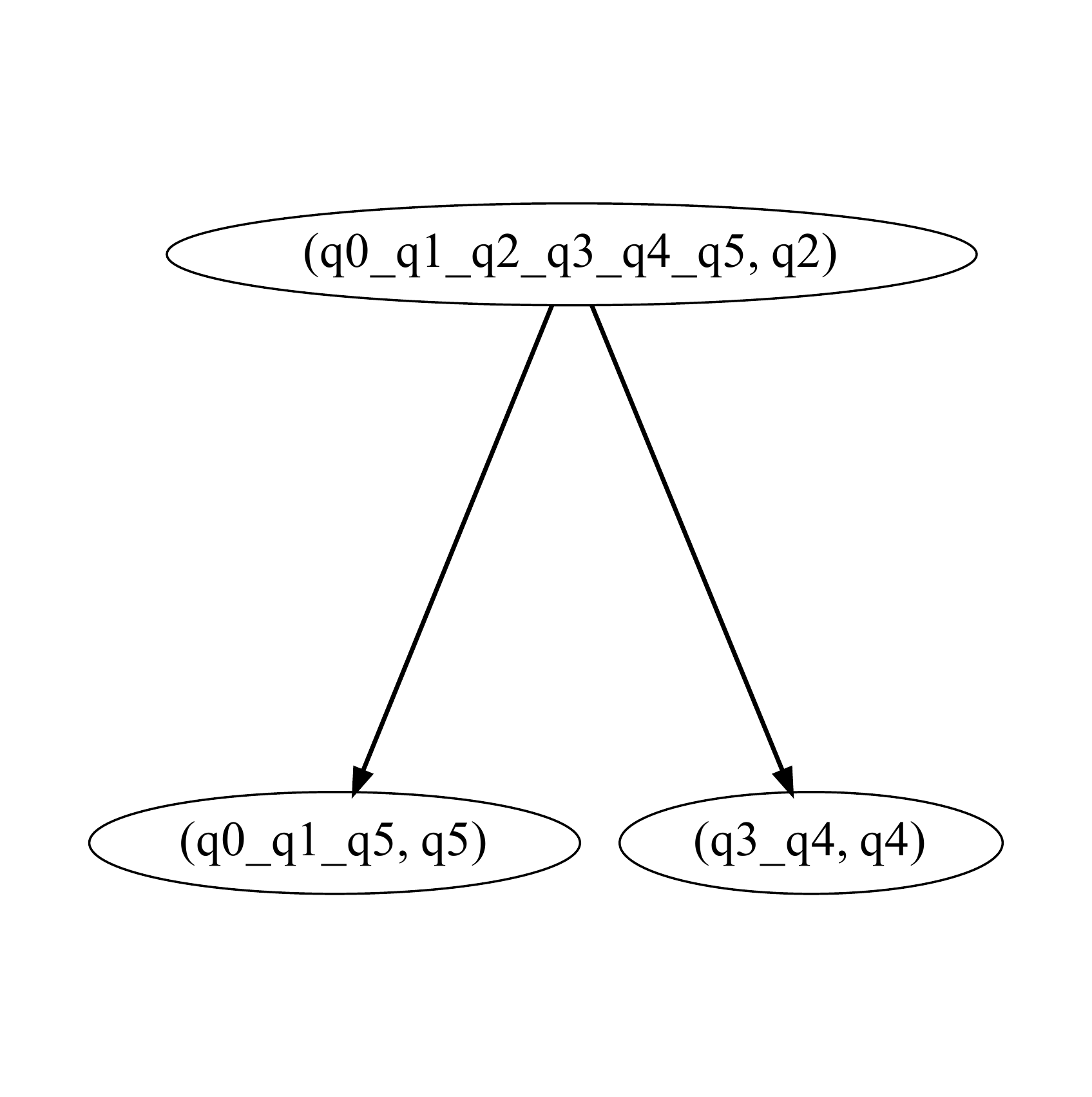}}

   \caption{\small $\Phi_9$ (six control states). When used with $D_{\Phi_9}$, \algo returns ``terminating'' in 2s. 
   When used with $D_{\Phi_9}'$, \algo returns ``unknown'' in 2s.}\label{fig:det_dependence}
\end{figure}

The analysis conducted by \algo is dependent on the DEF because DEF nodes guide
the construction of quotient graphs and the path sets $\Pi(G,v)$. We observed
that in rare cases \algo can return ``unknown'' with one DEF and ``terminating''
with another DEF for the same FMP. 
Fig.\,\ref{fig:det_dependence} shows an example of an FMP ($\Phi_9$) that
exhibits this feature. When run with $DET_{\Phi_9}$, \algo returns
``terminating'' but it returns ''unknown'' when run with $DET_{\Phi_9}'$.

The formal analysis presented in
Sec.\,\ref{sec:formal} ensures that the input policy terminates if \algo returns
``terminating'' for any DEF for the policy. The set of possible DEFs is finite,
albeit exponential in the number of qstates in the FMP and running \algo with
all possible DEFs for an FMP  could be viewed as a reasonable cost of
determining termination. An alternative approach could develop a probabilistic
by allocating a fixed computational budget towards assessing termination,
and iteratively sampling a DEF and running \algo with the sampled DEF until the
computational budget is exhausted.

\section{Conclusions}
\label{sec:conclusions}

We presented a new approach that uses graph theoretic decompositions of
finite-memory policies to efficiently determine whether they permit
non-terminating executions. In contrast to prior approaches, this framework
neither requires qualitative semantics nor does it place a priori restrictions
on the structure of FMPs that it can analyze. Empirical evaluation shows that
these methods go beyond the scope of existing approaches for this problem.
Several optimizations and extensions are possible with this new framework for
analyzing generalized plans. Parallelized implementations for per-DET analyses,
better bookkeeping and compiled language implementations could be used to speed
up the presented algorithm. The current methods could be refined to incorporate
more precise estimates of graph connectivity to refine the set of paths that
could be composed together in an execution. Edge conditions can be also be
included in this analysis. Heuristic search techniques could be developed to
create DETs that are more likely to identify termination and mitigate the impact
of the order of node elimination in the creation of DETs and in the analysis
carried out by \algo. Finally, the hierarchical approach developed in this paper
could be used in heuristics and early pruning
strategies  for learning and synthesizing generalized finite memory policies.

\acks{This work was supported in part by the National Science Foundation under
grant IIS 1942856. We thank the anonymous reviewers for their helpful comments
and suggestions.
}

\appendix
\section{Additional Results}
\label{app:more}
Figs.\,\ref{fig:algo_phi8_1} and \ref{fig:algo_phi8_2} show the execution of
\algo on a policy that resulted in a longer sequence of reductions and
iterations of the main while loop. The total run time for analysis was 1.2s. 

\begin{figure}[h!]
  \centering
  \subcaptionbox{\small  $\Phi_{8}$ \label{fig:phi8_fmp}}
        {\includegraphics[width=6in]{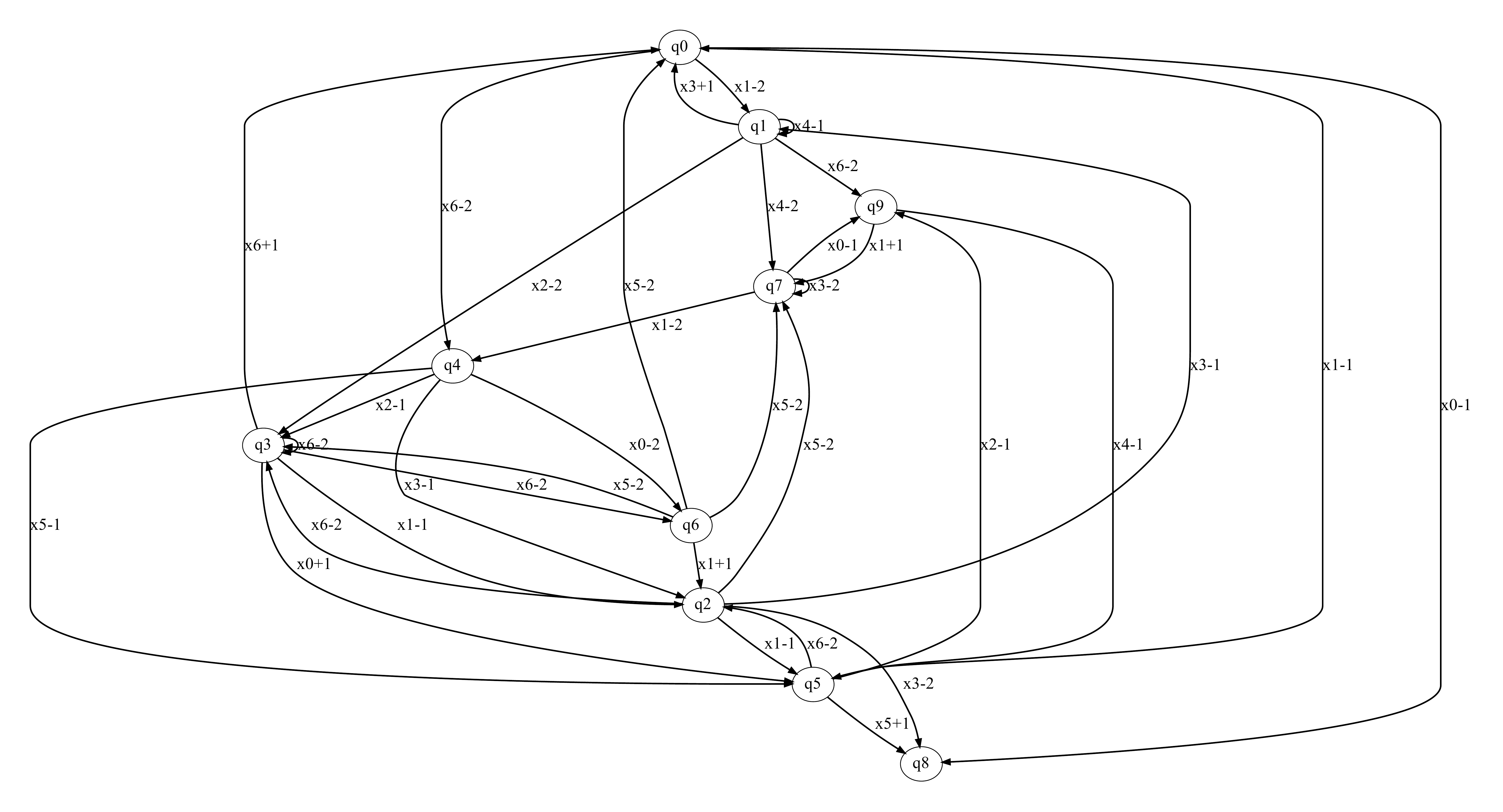}}
    \subcaptionbox{\small $D_{\Phi_8}$\label{fig:phi8_det}}
        {\includegraphics[width=2.5in]{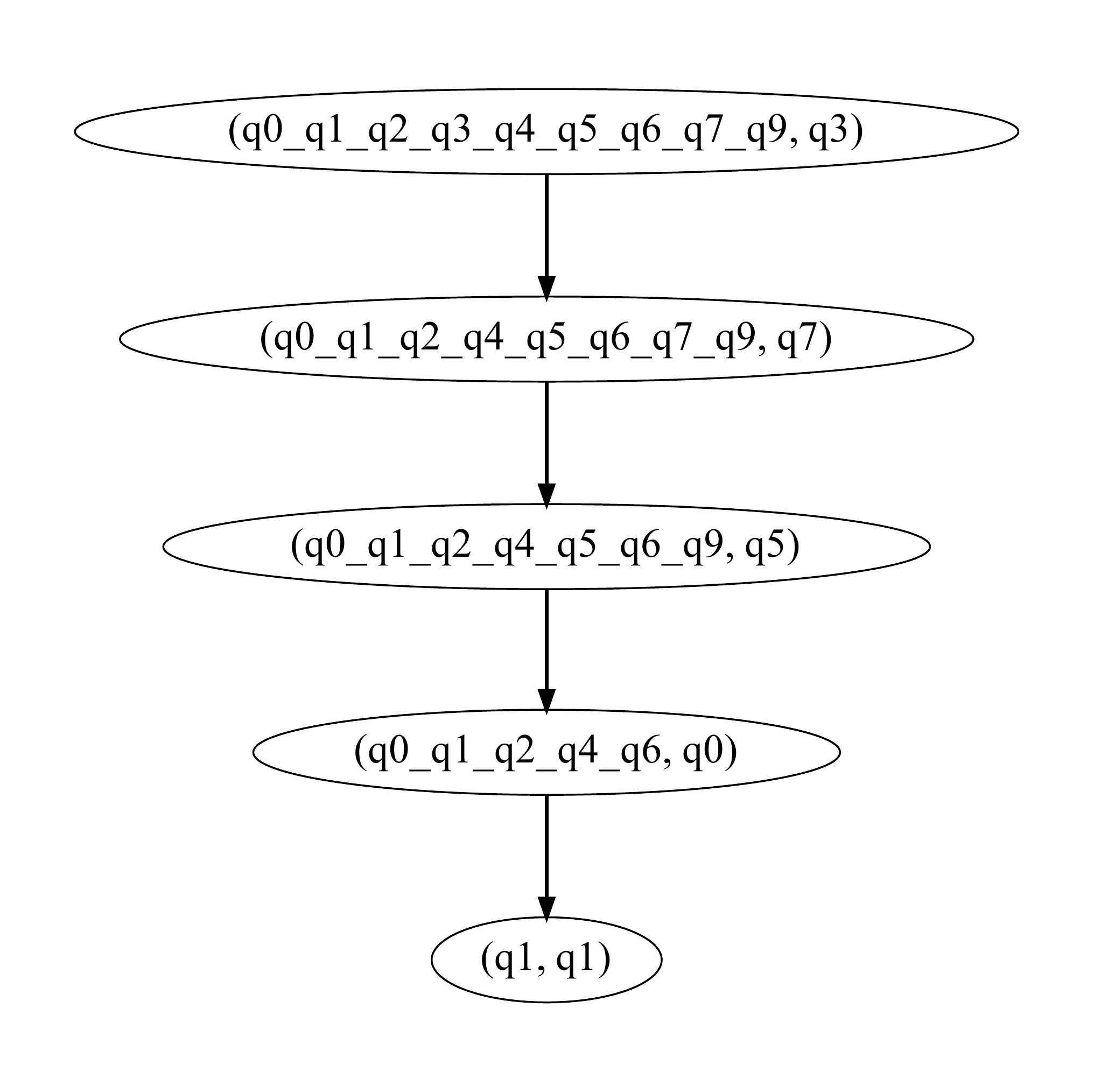}}

   \caption{\small Steps in \algo's execution on $\Phi_8$ (ten control states
   and seven variables). Continued on
   Fig.\,\ref{fig:algo_phi8_2}.}\label{fig:algo_phi8_1}
\end{figure}        
        
\begin{figure}[h!]
  \centering   
      \subcaptionbox{\small ${\Phi_8^2}=\Phi_8 \setminus \set{x2-, x4-, x5-}$\label{fig:phi8_fmp1}}
        {\includegraphics[width=3.5in]{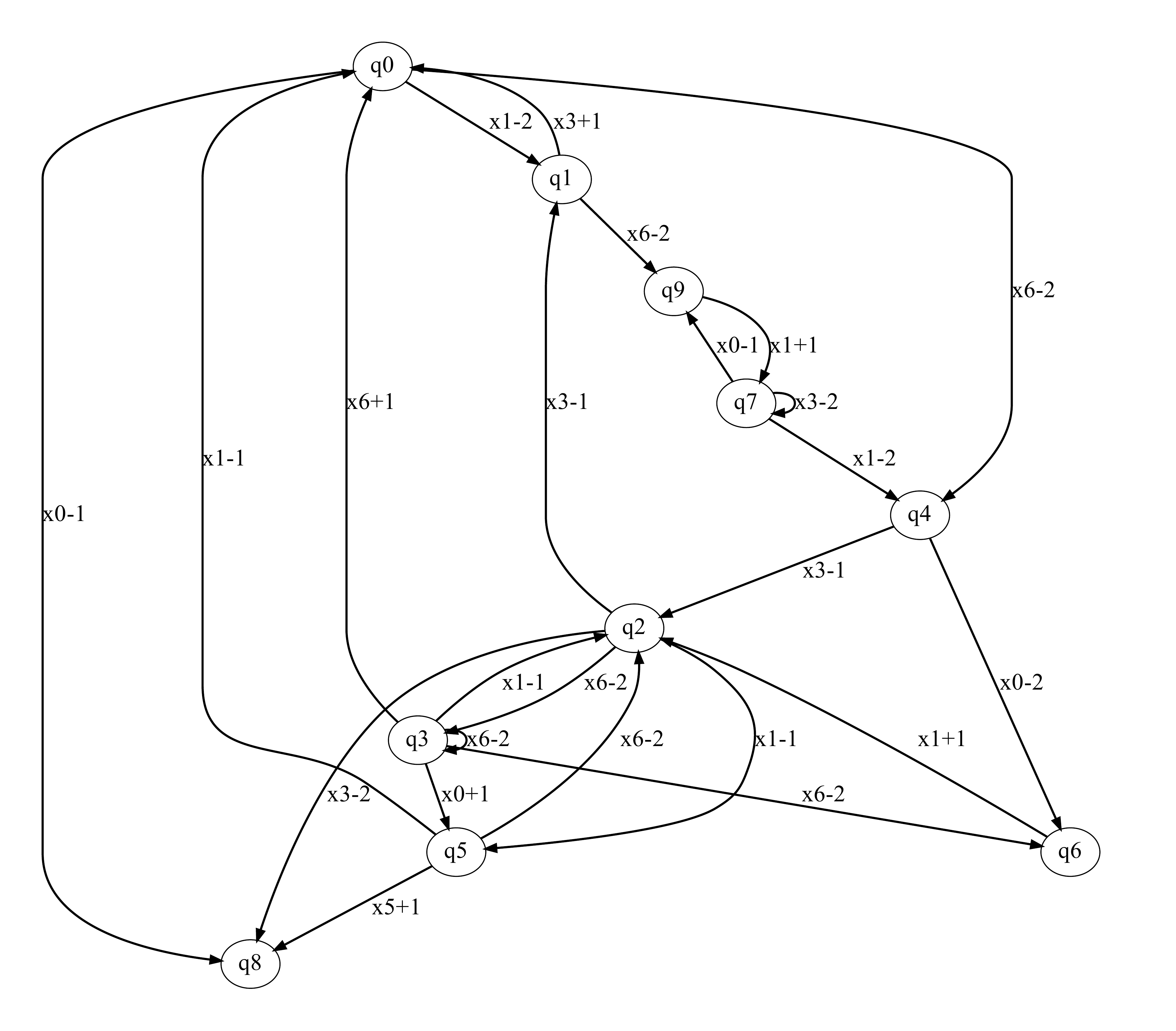}}
    \subcaptionbox{\small  $D_{\Phi_8^2}$\label{fig:phi8_det1}}           {\includegraphics[width=2in]{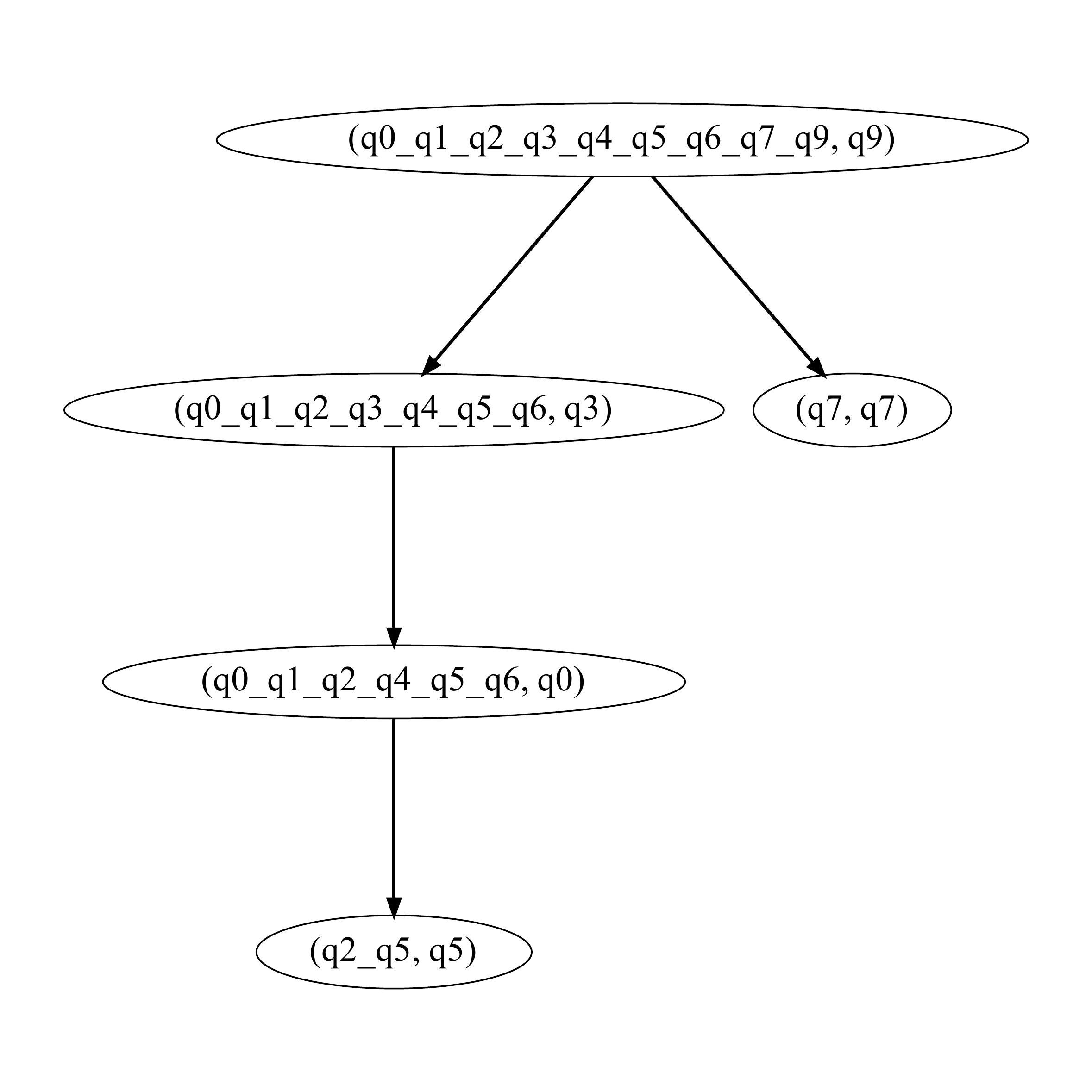}}    
    \subcaptionbox{\small  $\Phi_8^3 = \Phi_8^2\setminus\set{x6-}$}
        {\includegraphics[width=2in]{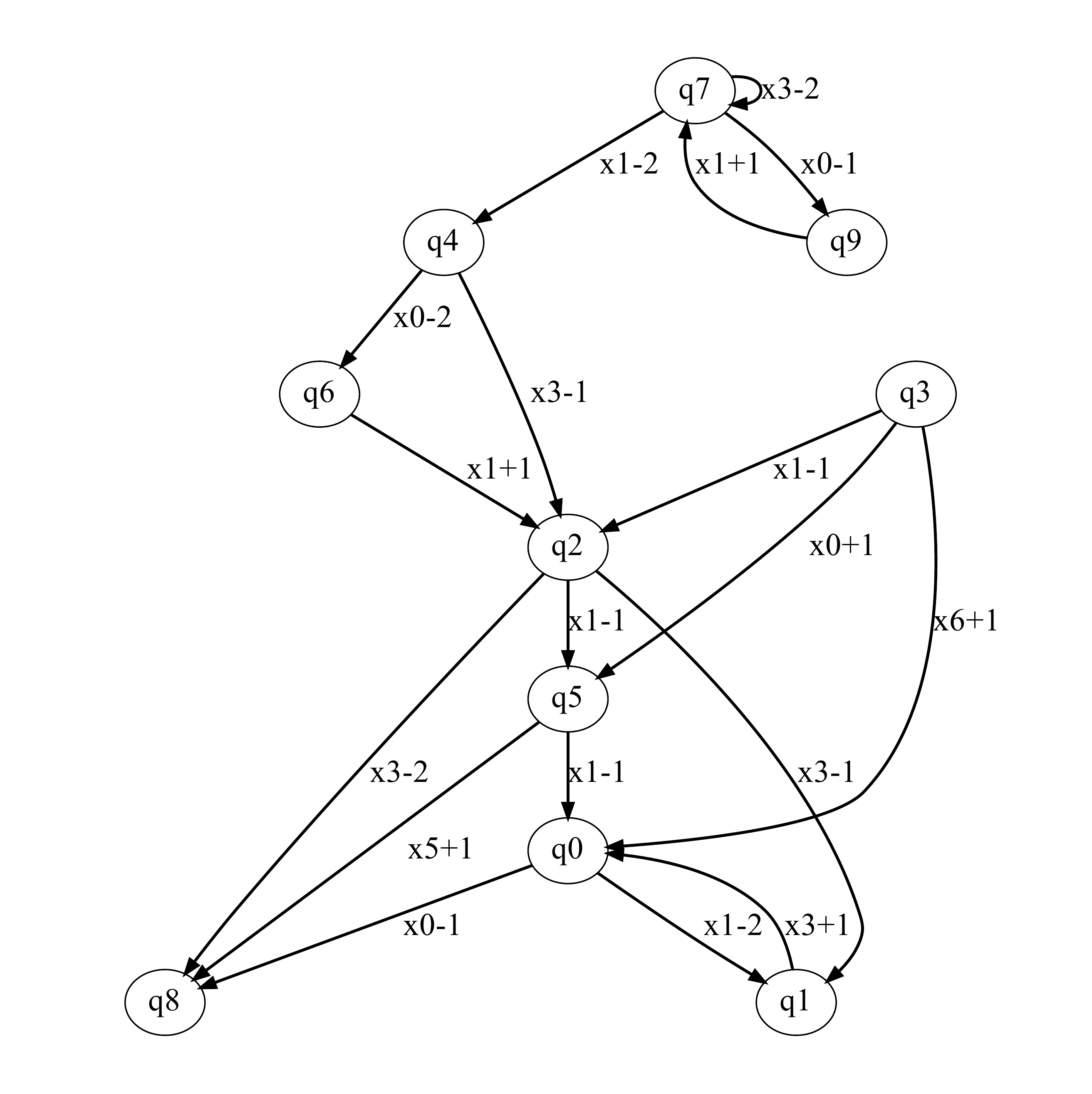}}
    \subcaptionbox{\small  $D_{\Phi_8^3}$\label{fig:phi8_det2}}
        {\includegraphics[width=2in]{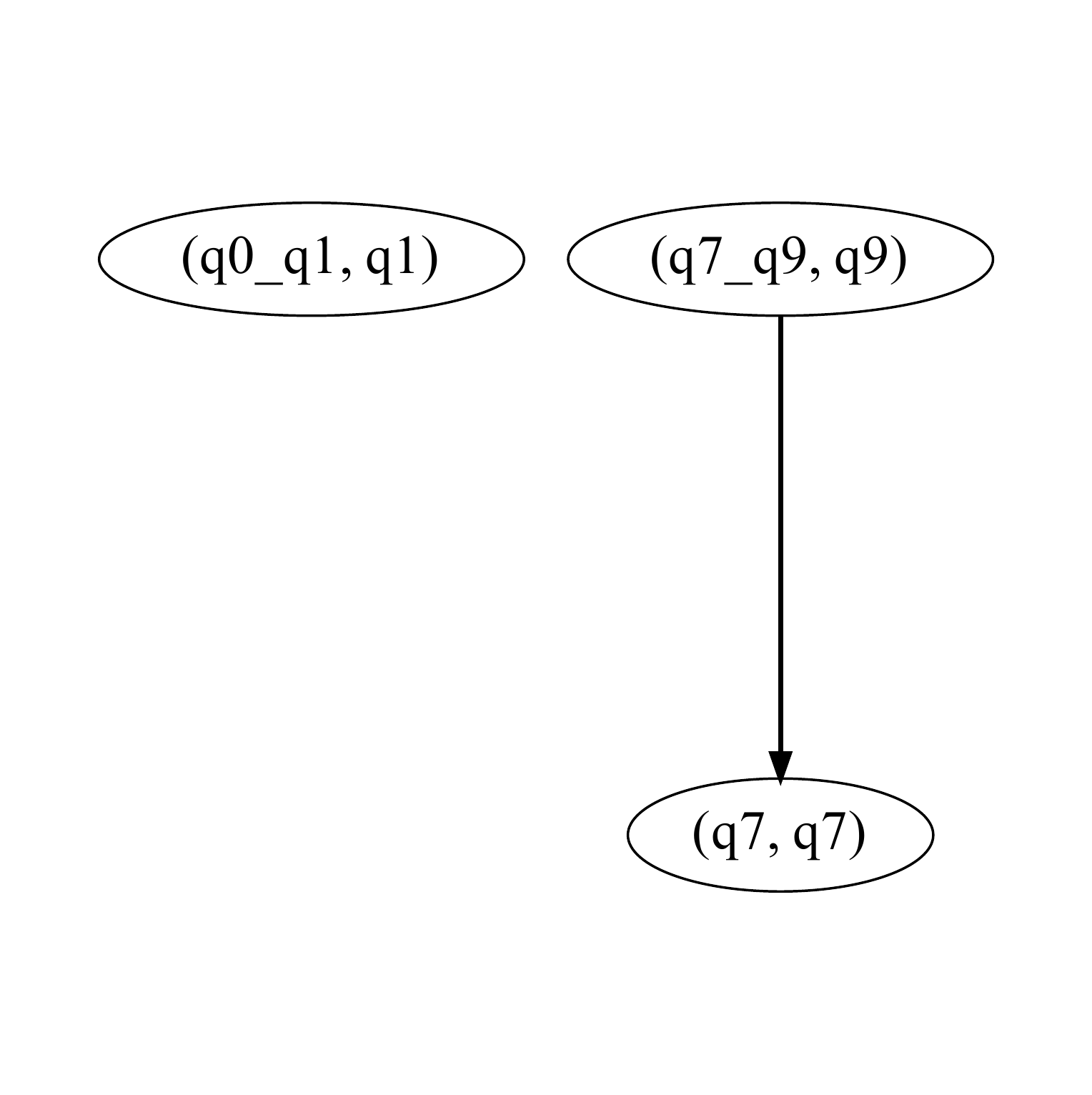}}
  \subcaptionbox{\small  $\Phi_8^4 = \Phi_8^3\setminus\set{x0-}$\label{fig:phi8_fmp3}}
        {\includegraphics[width=1.7in]{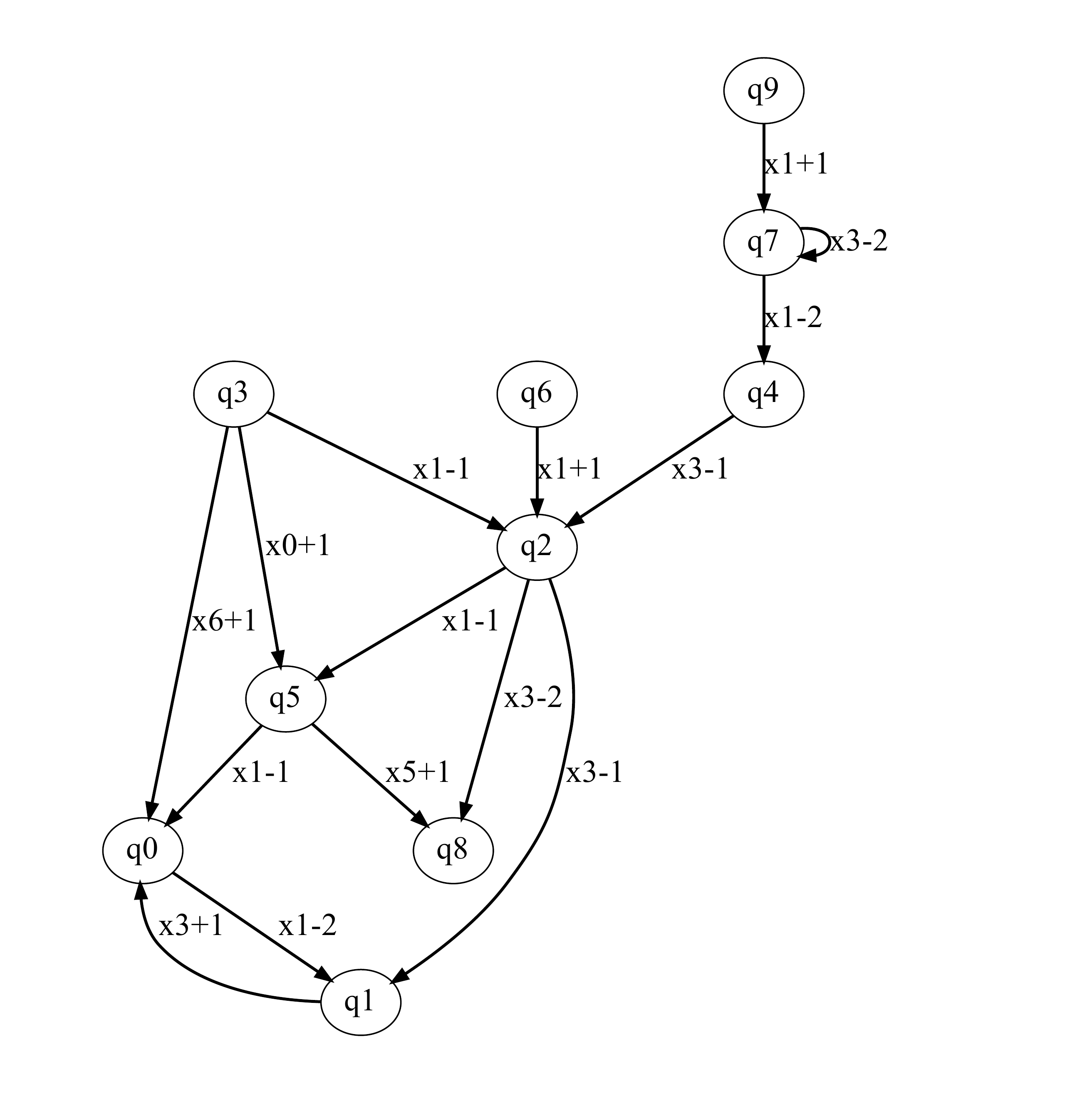}}
  \subcaptionbox{\small  $D_{\Phi_8^4}$\label{fig:phi8_det4}}
        {\includegraphics[width=1.5in]{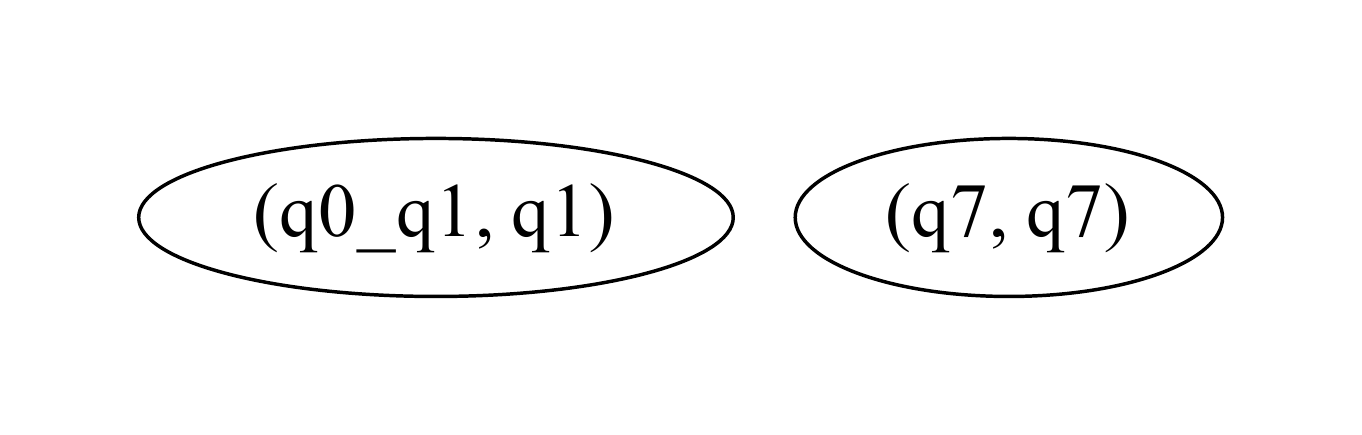}}
  \subcaptionbox{\small  $\Phi_8^5 = \Phi_8^4\setminus\set{x1-}$\label{fig:phi8_fmp5}}
        {\includegraphics[width=1.5in]{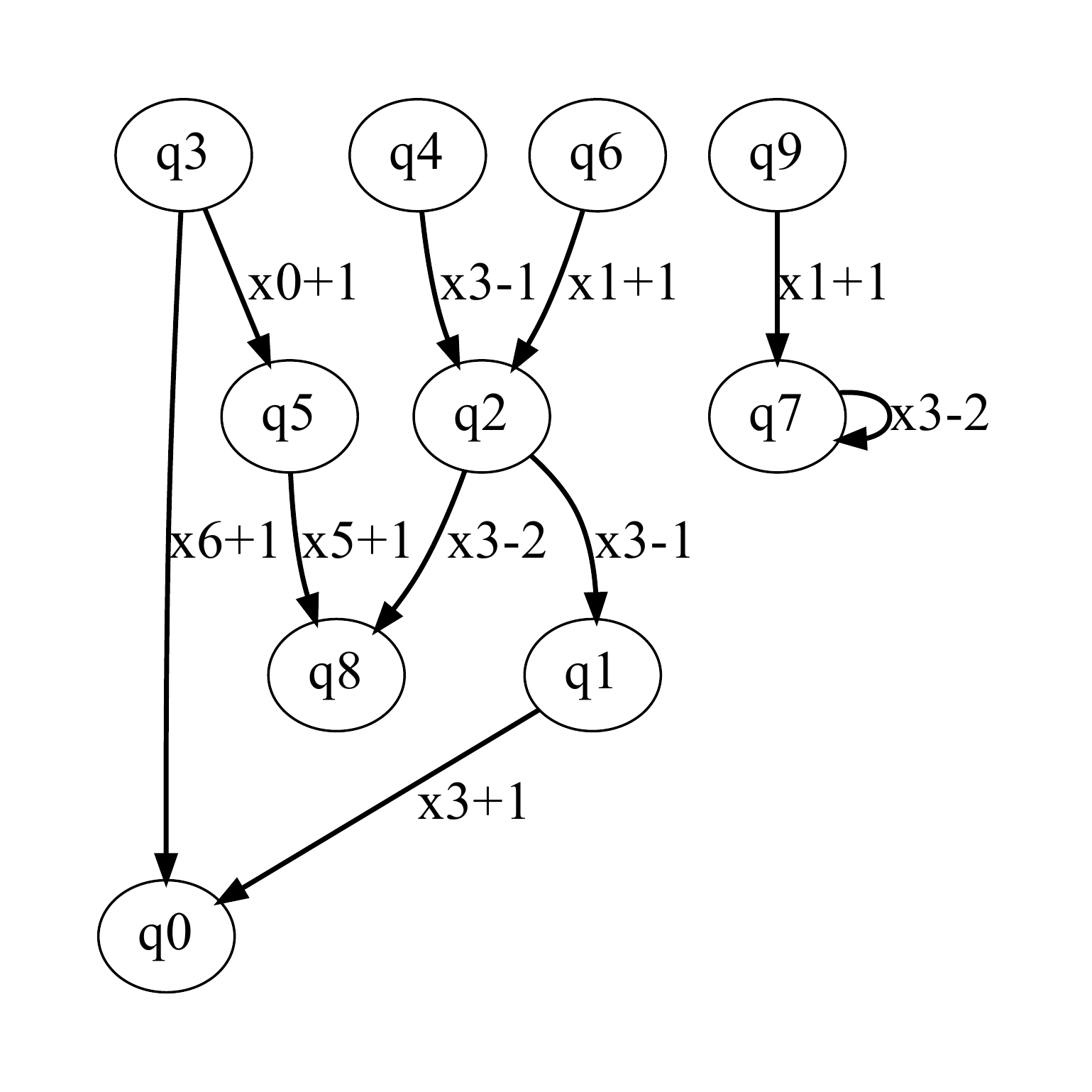}}
  \subcaptionbox{\small  $D_{\Phi_8^5}$\label{fig:phi8_det5}}
        {\includegraphics[width=0.7in]{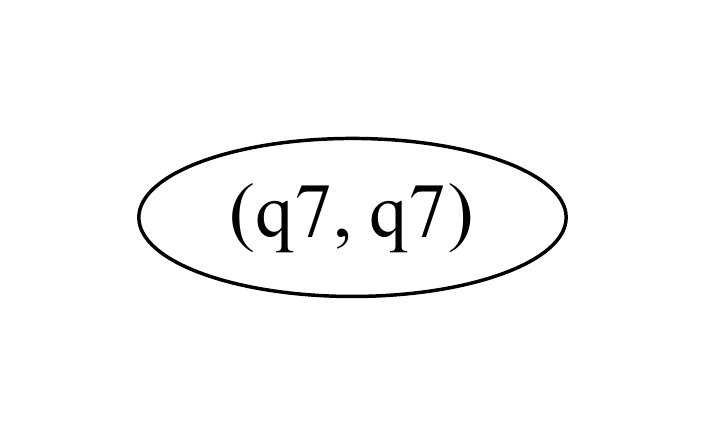}}
   \caption{\small (continued) Steps in \algo's execution on $\Phi_8$. Total runtime: 1.2s. }\label{fig:algo_phi8_2}
\end{figure}

\clearpage

\bibliography{planning}
\bibliographystyle{theapa}

\end{document}